\documentclass{article}

\usepackage{PRIMEarxiv}

\usepackage[utf8]{inputenc} 
\usepackage[T1]{fontenc}    
\usepackage{hyperref}       
\usepackage{url}            
\usepackage{booktabs}       
\usepackage{amsfonts}       
\usepackage{nicefrac}       
\usepackage{microtype}      
\usepackage{fancyhdr}       
\usepackage{graphicx}       
\usepackage{cite}
\usepackage{amsmath,amssymb,amsfonts}
\usepackage{algorithmic}
\usepackage{textcomp}
\usepackage{xcolor}
\usepackage{subcaption}
\usepackage{soul}
\usepackage{tikz}
\usepackage{forest}
\usetikzlibrary{trees,positioning,shapes,shadows,arrows.meta}

\graphicspath{{media/}}     

\title{A Survey of Transformer-based Language Models with Focus on
Efficiency}

\author{
  Wazib~Ansar \\
  A. K. Choudhury School of IT \\
  University of Calcutta, 
  Kolkata, India\\
  \texttt{waakcs\_rs@caluniv.ac.in} \\
   \And
  Saptarsi~Goswami \\
  Department of Computer Science \\
  Bangabasi Morning College \\
  Kolkata, India\\
  \texttt{sgakc@caluniv.ac.in} \\
  \And
  Amlan~Chakrabarti \\
  A. K. Choudhury School of IT \\
  University of Calcutta \\
  Kolkata, India\\
  \texttt{acakcs@caluniv.ac.in} \\
}

\begin{document}
\maketitle

\begin{abstract}
The emergence of Transformer-based Large Language Models (LLMs) has substantially augmented the capabilities of Natural Language Processing (NLP), thereby intensifying the demand for computational resources. Therefore, enhancing efficiency based on factors like computational requirements, energy consumption, carbon footprint and financial cost has become a vital area of research. This motivates us to conduct a survey on Transformer-based LLMs in NLP from the perspective of efficiency. In this survey of 312 articles, the efficiency-improvement endeavors have been systematically discussed targeting various aspects such as data curation, model design, model downsizing, and dynamic inferencing. This has been augmented with efficiency considerations in model adaptation strategies like pre-training, fine-tuning, prompt-engineering and Retrieval-Augmented Generation (RAG). Furthermore, a statistical analysis followed by an in-depth evaluation of the efficiency and efficacy of more than 30 renowned NLP models has been performed on 13 evaluation benchmarks. This paper offers valuable insights for researchers, professionals as well as scholars, and explores the trend of research toward sustainable practices in NLP.

\end{abstract}

\keywords{Large Language Model (LLM) \and LLM Adaptation \and LLM Evaluation \and LLM Scaling \and Model Optimization \and Natural Language Processing (NLP)}

\section{Introduction}
\label{intro}
There has been a remarkable transformation in the field of Natural Language Processing (NLP) over the past decade. The driving force behind this has been the Transformers architecture \cite{vaswani2017attention} and its incorporation in Language Models (LM) like Bidirectional Encoder Representations from Transformers (BERT) \cite{kenton2019bert}, XLNet \cite{yang2019xlnet}, Bidirectional and Auto-Regressive Transformers (BART) \cite{lewis2020bart}, Generative-Pre-trained Transformer (GPT) \cite{radford2018improving} (along with its successors i.e. GPT-2 \cite{radford2019language}, GPT-3 \cite{brown2020language}, GPT-4 \cite{achiam2023gpt} and GPT-5\footnote{\url{https://openai.com/index/introducing-gpt-5/}}), LLaMA \cite{touvron2023llama} (its subsequent versions LLaMA 2 \cite{touvron2023llama2}, LLaMA 3 \cite{dubey2024llama} and LLaMa 4\footnote{\url{https://developer.meta.com/ai/models/llama-4/}}), Gemini 2.5 \cite{comanici2025gemini}, Gemini 3.1 Pro\footnote{\url{https://ai.google.dev/gemini-api/docs/models/gemini-3.1-pro-preview}}, Gemma 3 \cite{team2025gemma}, o3 \cite{el2025competitive}, Phi-4 \cite{abdin2024phi}, Mistral 7B \cite{jiang2023mistral7b}, Mixtral 8x7B \cite{jiang2024mixtral}, and the DeepSeek family (DeepSeek-v2 \cite{liu2024deepseek}, DeepSeek-v3 \cite{liu2024deepseek3}, DeepSeek-r1 \cite{guo2025deepseek}). These advancements have empowered NLP models to perform complex linguistic tasks relating to understanding natural language and even generating responses as a human would provide \cite{chowdhary2020natural}.

These research accomplishments have been facilitated by the availability of voluminous data, sophisticated modeling approaches, and high-end computing resources. As the model complexity rises, computational requirements surge exponentially \cite{sengupta2025position}. With the apparent deceleration of Moore's Law\footnote{\url{https://www.nature.com/news/the-chips-are-down-for-moore-s-law-1.19338}}, increasing the performance of algorithms comes at the cost of straining the computing resources with faltering efficiency. This leads to high energy requirements translating into a hike in carbon emissions \cite{halgamuge2026emissions, strubell2019energy, patterson2021carbon}. Therefore, the need of the hour is to think out of the box and devise sustainable methodologies that can keep the performance growth rate steady while being efficient enough to be deployed in resource-constrained environments \cite{schwartz2020green}.

The term "efficiency" can be broadly defined as the trade-off between the performance and the associated cost factors. The goal of efficient modeling is to achieve Pareto improvement by reducing the training as well as inference cost of a model to achieve a benchmark level of performance \cite{durlich2023concept}. The cost factors include the number of Floating-point Operations (FLOPs), inference time, throughput, model size, number of model parameters, energy consumption, carbon emissions, and pricing. The efforts to enhance efficiency can cover various aspects of modeling i.e. data curation, model design, model downsizing and dynamic inferencing. Efficiency can also be enhanced during Large Language Model (LLM) adaptation stages like pre-training, fine-tuning, prompt-engineering, and Retrieval-Augmented-Generation (RAG). Thus, achieving efficiency improvement in NLP models is a multifaceted task full of nuances.

To summarize the developments towards enhancing efficiency of LLMs, various surveys have been conducted before. Bannour and Ligozat \cite{bannour2021evaluating} performed a systematic review of the carbon footprint of NLP models. They studied the accuracy as well as applicability of tools for assessing the energy consumption and carbon emissions of contemporary NLP models. However, the study was confined to assessing the environmental impact of NLP models for a single task of named entity recognition. Khadivi and Sato \cite{khadivi2023bibliometric} performed a bibliometric analysis of papers published between 2002 to 2021 based on factors like growth-rate, doubling-time, and collaboration among authors. Treviso et al. \cite{treviso2023efficient} performed a literature review of efficient approaches in NLP focusing on data processing, model design, and hardware utilization. While the paper deals with theoretical narratives, a comparative analysis of results is missing from the viewpoint of efficiency. Xu and McAuley \cite{xu2023survey} conducted a review of model compression and acceleration including metrics for efficiency evaluation. Their study was limited to pre-trained models and did not account for data or parameter efficiency. Tay et al. \cite{catania2023conversational} presented a taxonomy of efficient Transformer models in NLP in the form of a literature review. However, it majorly focuses on model design and lacks coverage of efficiency considerations during other phases like data curation, or LLM adaptation. Wan et al. \cite{wan2024efficient} presented a systematic review of efficient LLMs spanning model-centric, data-centric, and framework-centric methods. While the review offers breadth and structure, its limitations lie in the lack of empirical benchmarking across methods and limited discussion on scalability and environmental impact of LLMs. Zhou et al. \cite{zhou2024survey} surveyed methods to reduce computational and memory costs during inference in LLMs. However, it provides limited coverage of data curation, pre‑training, and fine‑tuning strategies, which are critical for end‑to‑end efficiency. Moradi et al. \cite{moradi2025critical} performed a critical review of various LLM adaptation strategies. In their study, they suggested future research directions like enhancing efficiency, interpretability, and data curation quality of LLMs. Overall, existing surveys suffer from two major drawbacks: Firstly, there is a dearth of a holistic coverage spanning all phases of model design and adaptation. Secondly, they lack statistical analysis of the research landscape and extensive evaluation of performance-efficiency trade-offs of LLMs. In particular, they fall short of performance comparison across established benchmarks together with estimation of cost-effectiveness, and the environmental impact based on diverse metrics.

To address these shortcomings, we augment the body of knowledge with a survey of the efficiency considerations of Transformer-based models in NLP. Firstly, presents a primer on NLP, the evolution of Transformers, components of the Transformer architecture, genesis of LLMs, the laws of scaling, and the techniques for estimation of energy usage and carbon footprint of LLMs. Secondly, it discusses the efficient modeling considerations for model optimization and LLM adaptation. The model optimization strategies cover data curation, model design, model downsizing and dynamic inferencing. Whereas, efficient LLM adaptation strategies span across pre-training, fine-tuning, prompt-engineering and RAG. Thirdly, it lays forth the evaluation benchmarks and metrics to assess the efficiency, cost and environmental impact of LLMs. Finally, a statistical analysis of the current LLM landscape, performance comparisons across multiple benchmarks, and efficiency evaluations using diverse metrics has been conducted. This paper targets researchers, professionals as well as scholars interested in NLP particularly LLMs, and wish to design efficient and lean models with reduced computational requirements. The key contributions of this paper are as follows:

\begin{enumerate}
    \item We conduct a survey of 312 articles on the efficiency considerations of Transformer-based LMs. 

    \item We present a comprehensive foundational primer detailing the evolution of Transformers, the genesis of LLMs, scaling laws, and techniques for estimating the energy usage and carbon footprint of LLMs.

    \item We conduct a structured analysis of efficiency considerations covering both model optimization (data curation, model design, model downsizing and dynamic inferencing) and LLM adaptation (pre-training, fine-tuning, prompt-engineering, and RAG).

    \item We perform extensive statistical and empirical evaluation through detailed analysis of the current LLM landscape, performance comparisons on multiple benchmarks, together with efficiency estimation based on various measures.

    \item We assess the current trend towards efficient modeling in LLMs and present a road map for Pareto-optimality.
\end{enumerate}

The remainder of this paper has been organized in the following manner: Section \ref{sm} puts forth the methodology adopted for this survey. Section \ref{fc} enunciates the foundational concepts associated with Transformers in NLP, the laws of scaling LLMs along with estimation of energy consumption and carbon footprint. Section \ref{emc} showcases the endeavours towards efficient modeling. Section \ref{ev} discusses the benchmarks for LLM evaluation and efficiency measures. Section \ref{rr} presents the results of statistical analysis, performance evaluation, and efficiency estimation of LLMs with associated discussion. Finally, in Section \ref{concl}, the conclusions are drawn and the future scope is ushered.

\begin{figure*}[!t]
\centering
\includegraphics[width=0.6\linewidth]{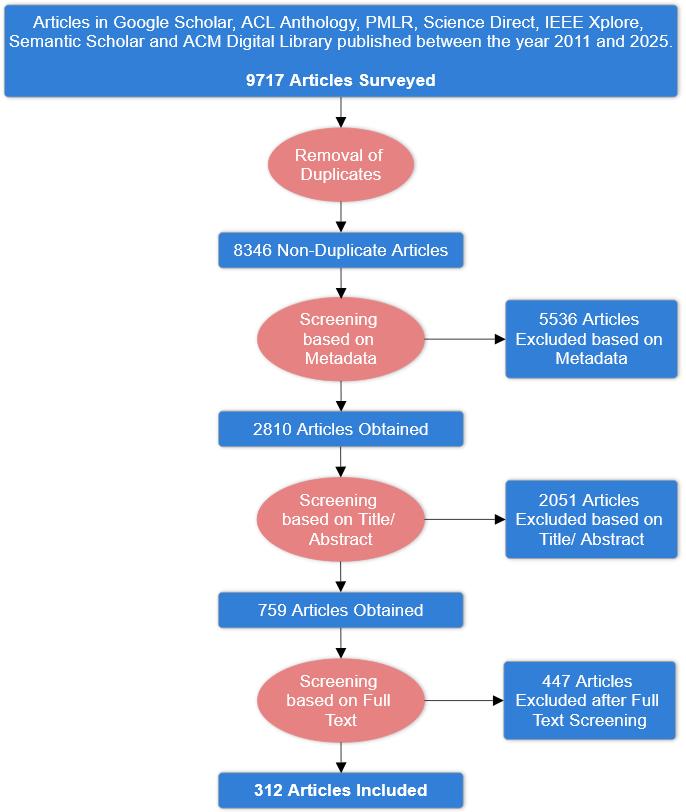}
\caption{The survey process flow diagram}
\label{fig18}       
\end{figure*}

\section{Survey Process}
\label{sm}

In this section, the survey process comprising the selection criteria for the articles along with the research questions have been discussed.

\subsection{Selection Criteria}
\label{sc}

The survey comprises original as well as review articles written in the English language published in digital libraries like Google Scholar\footnote{\url{https://scholar.google.co.in}}, ACL Anthology\footnote{\url{https://aclanthology.org/}}, PMLR\footnote{\url{https://proceedings.mlr.press/}}, ACM Digital Library\footnote{\url{https://dl.acm.org}}, IEEE Xplore, Semantic Scholar\footnote{\url{https://www.semanticscholar.org}} and Science Direct\footnote{\url{https://www.sciencedirect.com}}. The articles were retrieved using keywords like "NLP", "LLM", "Transformers", "embedding", "pre-training", "fine-tuning", "prompt-engineering", "RAG", alongside keywords like "sustainability", "carbon footprint", "energy", "emissions", "cost", "pricing", and "efficiency". Articles published in the last 15 years, i.e. between 2011 to 2025 were included with primary focus on articles published since 2017 (the year when the term "Transformer" was introduced). Recently published papers were given higher priority and special attention was paid to include the pioneering works. While selecting articles on the same topic or having similar methodology, the CORE ranking of the conference or the journal's impact factor were taken into account together with the number of citations. The duplicate articles were removed if it was observed that the writers had written similar works or if pre-print versions were present for the published articles. At first, 9,717 articles were retrieved. Out of which, 1371 duplicates were removed. After processing article metadata like "title", "abstract", "reference", "author list", "year of publication", "name of journal or conference" and "number of citations", 5,536 articles were filtered out through statistical examination based on the above-mentioned criteria. From the remaining 2,810 articles, 759 articles were short-listed after going through the title and abstract. Finally, the full-text screening of the short-listed articles was performed leading to 312 articles to be included in this survey. For a systematic presentation, the review of articles has been structured into coherent sections according to their significance. The survey process has been illustrated in Figure \ref{fig18}.

\subsection{Research Questions}
\label{rq}

The following Research Questions (RQs) have been formulated for this survey and attempts have been made to answer them in this paper:

\begin{itemize}
    \item[\textbf{RQ1:}] What are the components of Transformer architecture, how does the variants of attention mechanism affect computational complexity and how did Transformers become the backbone architecture of modern LLMs?

    \item[\textbf{RQ2:}] How have LLMs scaled in the recent past, what are the scaling laws and how to estimate the carbon footprint? 

    \item[\textbf{RQ3:}] What efficiency considerations are there concerning data curation, model design, model downsizing and dynamic inferencing?

    \item[\textbf{RQ4:}] What efficiency considerations are there targeting various stages of model adaptation like pre-training, fine-tuning, prompt-engineering and RAG?

    \item[\textbf{RQ5:}] Which are the benchmarks for evaluation of modern LLMs and what are the measures to estimate the efficiency of LLMs?

    \item[\textbf{RQ6:}] What is the research trend and to what extent efficiency is being considered in modern LLMs while improving model efficacy?
    
\end{itemize}

The answers to RQ1, RQ3, RQ3, RQ4, RQ5 and RQ6 have been provided in Section \ref{trfr}, Section \ref{sol}, Section \ref{mo}, Section \ref{mas}, Section \ref{ev}, and Section \ref{rr} respectively.

\section{Foundational Concepts}
\label{fc}

\subsection{Overview of NLP}
\label{nlp}
\begin{figure*}[!t]
    \centering
    \includegraphics[width=0.81\linewidth]{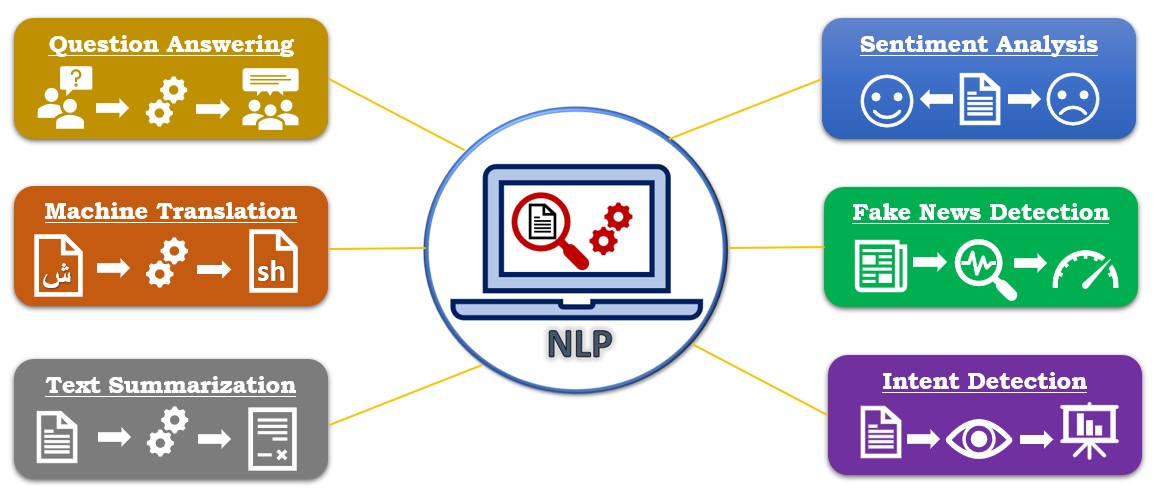}
    \caption{Applications of NLP}
    \label{fig5}       
    \end{figure*}

NLP is a research domain concerned with providing computing devices the ability to comprehend and process input text in natural language understood by human beings. Using NLP, one can extract meaningful information from unstructured text and even synthesize outputs in natural language \cite{chowdhary2020natural}. Two complementary facets of NLP are Natural Language Understanding (NLU) and Natural Language Generation (NLG) as illustrated in Figure \ref{fig5}. NLU enables computers to understand and derive meaning from natural language. By bridging the gap between unstructured text data and representations that are understood by machines, NLU enables machines to comprehend and process natural language input. Instances include sentiment analysis \cite{liu2020sentiment}, intent detection \cite{weld2022survey}, and fake news detection \cite{ansar2021combating}. On the other hand, NLG enables computers to produce natural language from structured data or other unstructured text inputs. The objective of NLG is to communicate information in a way that is comprehensible to human beings and appropriate as per the given context. Instances include question answering \cite{lewis2018generative}, machine translation \cite{cho2014properties} and text summarization \cite{nallapati2016sequence}.

Over the years, NLP has evolved significantly driven by groundbreaking innovations and advances in computational capabilities. The advent of Deep Learning (DL) models enabled LMs to learn complex linguistic features and representations directly from raw text, eliminating the need for manual feature engineering used in earlier methods\footnote{As this survey focuses on Transformer-based LMs, earlier works on symbolic Artificial Intelligence (AI) and Traditional Machine Learning (TML) have been excluded to maintain specificity.}. This led to the development of models like Recurrent Neural Networks (RNN) \cite{irsoy2014opinion} and CNN \cite{sharma2020sentimental} that could understand the context better than Traditional Machine Learning (TML). While CNN performs better at capturing local spatial relationships \cite{sharma2020sentimental}, RNN is more suitable for global temporal dependencies \cite{irsoy2014opinion}. However, RNNs failed to capture dependencies beyond a certain length due to inherent “vanishing-gradient” and “exploding-gradient” issues \cite{ribeiro2020beyond}. Long Short Term Memory (LSTM) based NLP models \cite{jelodar2020deep} ameliorated this issue with a gated mechanism having the gradients pruned leading to improved accuracy. Building on these foundations, pre-trained LMs like Embeddings from Language-Models (ELMo) \cite{peters-etal-2018-deep} and Universal Language Model Fine-tuning (ULMFit) \cite{howard2018universal} leveraged transfer learning to adapt general-purpose representations to specific downstream tasks with remarkable efficiency. The transformative leap occurred with the Transformer-based models discussed in Section \ref{trfr}. The Transformers eliminated the sequential bottlenecks of RNNs and enabled LMs to capture global dependencies across lengthy sequences. Its utility was further amplified with large-scale pre-trained models or LLMs setting new benchmarks across a wide range of NLP applications.

\subsection{RQ1: Transformers in NLP}
\label{trfr}
This section discusses the evolution of Transformers from encoder–decoder architectures. It elaborates the concept of attention mechanism along with other fundamental components, and explores their integration into LLMs.

\subsubsection{Encoder-Decoder Architectures}
\label{eda}

A series of developments paved the way for the Transformers. Earlier works on NLG tasks like machine translation applied sequence-to-sequence models comprising two RNN blocks namely, the encoder and the decoder \cite{cho2014properties, sutskever2014sequence}. Given an input sequence $X=(x_1, x_2, ... , x_n)$, the RNN-based \textit{encoder} derives a hidden representation $H=(h_1, h_2, ... , h_n)$. Subsequently, a few other non-linear functions can also be applied to obtain the final $H$. For $t^{th}$ time-step, $h_t$ is calculated from $x_t$ and $h_{t-1}$ as shown in Equation (\ref{eq41}). 

\begin{equation}
\label{eq41}
h_t = RNN(x_t, h_{t-1})
\end{equation}

The \textit{decoder} predicts one output token at each time-step on the basis of the previously predicted tokens ${y_1, y_2, ..., y_{t-1}}$ and $H$ as a joint probability distribution shown in Equation (\ref{eq42}).

\begin{equation}
\label{eq42}
y_t = p(y_t | {y_1, y_2, ..., y_{t-1}}, H)
\end{equation}

However, the above approach leads to loss of information with growing length of input due to compression of information into a fixed-length vector. To ameliorate this issue, Bahdanau et al. \cite{bahdanau2015neural} deployed a soft-search mechanism for identifying the significant tokens from the input sequence for the prediction of the output at a given time-step. For this, they introduced the term context vector $c_t$ derived from $H$ which weighs the significance of the token hidden states as shown in Equation (\ref{eq53}).

\begin{equation}
\label{eq53}
c_t = \sum_{i=1}^{n} \alpha_{ti} \cdot h_i
\end{equation}

\begin{table*}[h!]
  \begin{center}
    \caption{Commonly Used Compatibility Functions}
    \label{table:table2}
     \rotatebox{0}{ \resizebox{0.9\textwidth}{!}{    
 \begin{tabular}{@{}llll@{}}
 \hline
 Type & Representation & Complexity & Reference\\ 
 \hline
 Similarity & $f_c^{s}(Q_{s}, K_{s})=sim(Q_{s}, K_{s})$  & $\mathcal{O}(n_{q}n_{k}d_{k})$ & Graves et al. \cite{graves2014neural} \\
  \hline
 Multiplicative or Dot & $f_c^{d}(Q_{s}, K_{s})=Q_{s}^{T} \cdot K_{s}$ & $\mathcal{O}(n_{q}n_{k}d_{k})$ & Luong et al. \cite{luong2015effective}  \\
  \hline 
 Scaled Multiplicative & $f_c^{m}(Q_{s}, K_{s})=\frac{Q_{s}^{T} \cdot K_{s}}{\sqrt{D}}$ & $\mathcal{O}(n_{q}n_{k}d_{k})$ & Vaswani et al. \cite{vaswani2017attention} \\
 \hline   
 Bilinear & $f_c^{b}(Q_{s}, K_{s})=Q_{s}^{T} \cdot W \cdot K_{s}$ & $\mathcal{O}(n_{q}n_{k}d_{k}+n_{k}d_{k}^{2})$ & Luong et al. \cite{luong2015effective} \\
  \hline 
 Concat & $f_c^{c}(Q_{s}, K_{s})=V_{a}^{T} \cdot \phi(W[K_{s};Q_{s}] + b)$ & $\mathcal{O}(n_{q}n_{k}d_{k})$ & Luong et al. \cite{luong2015effective} \\
  \hline   
 Additive & $f_c^{a}(Q_{s}, K_{s})=V_{a}^{T}g(W_1 K_{s} + W_2 Q_{s} + b)$ & $\mathcal{O}(n_{q}n_{k}d_{k})$ & Bahdanau et al. \cite{bahdanau2015neural}  \\ 
 \hline
 \multicolumn{4}{p{15cm}}{Note: $n_{q}, n_{k}, d_{k}$ denotes the length of $Q_{s}$, length of $K_{s}$, and its dimensionality respectively. $W, W_1, W_2$ and $b$ are learnable parameters.}
\end{tabular}}}
\end{center}
\end{table*}

where $\alpha_{ti}$ is a distribution (often a softmax) function as follows:

\begin{equation}
\label{eq43}
\alpha_{ti} = \frac{exp(e_{ti})}{\sum_{j=1}^{n} exp(e_{tj})}
\end{equation}

given that, 

\begin{equation}
\label{eq44}
exp(e_{ti}) = f_a(s_{t-1}, h_{i})
\end{equation}

Here, $exp(e_{ti})$ evaluates the alignment between the output at position $t$ and the input tokens around position $i$. The $f_a(*)$ function takes the previous hidden state $s_{t-1}$ of the RNN decoder and $i^{th}$ time-step hidden representation $h_{i}$. Finally, the decoder applies a non-linear function $f_d(*)$ to generate the output $y_t$ for time-step $t$ as follows:

\begin{equation}
\label{eq45}
y_t = f_d(y_{t-1}, s_{t}, c_{t})
\end{equation}

\subsubsection{Attention Mechanism}
\label{am}

The above developments led to the foundation of the attention mechanism, an indispensable component of modern Transformer architecture. To compute attention, the input is transformed into an embedded sequence $Z \in \mathbb{R}^{L} \times \mathbb{R}^{D}$ comprising token and positional embeddings where $L$ is the sequence length and $D$ is embedding dimension. Then, key $K_{s}$, query $Q_{s}$, and value $V_{s}$ are calculated through linear transformations on the sequence $Z$ as follows:

\begin{equation}
\label{eq46}
Q_{s},K_{s},V_{s}=ZW^{q},ZW^{k},ZW^{v}
\end{equation}

where $W^{q} \in \mathbb{R}^{D \times d_{q}}, W^{k} \in \mathbb{R}^{D \times d_{k}}$ and $W^{v} \in \mathbb{R}^{D \times d_{v}} $ denote the weight matrices corresponding to $K_{s}$, $Q_{s}$ and $V_{s}$. The key $K$ represents the input features. These features might be at character-level, word-level, document-level, or a combination of multiple features. $Q_{s}$ is the vector whose relationship with $K$ is computed during attention computation. This is accomplished through a compatibility function $f_c(*)$ as follows:

\begin{equation}
\label{eq47}
e_a = f_c(Q_{s}, K_{s})
\end{equation}

One might notice the similarity between Equation (\ref{eq47}) and the alignment function in Equation (\ref{eq44}) wherein the alignment between the previous decoded token and the hidden states is computed.

Table \ref{table:table2} presents the variants of the compatibility function $f_c(*)$. The \textit{similarity attention} $f_c^{s}(*)$ \cite{graves2014neural} encompasses various functions (for instance cosine similarity) to measure the likeness between $Q_{s}$ and $K_{s}$. This is a foundational approach, as many other methods can be seen as its specific instances. The \textit{multiplicative or dot-product attention} $f_c^{d}(*)$ \cite{luong2015effective} is based on the idea to compute a simple dot product between $Q_{s}$ and $K_{s}$ vectors. This simplicity makes it fast and scalable, but it can be sensitive to the scale of the input vectors. A major drawback is that if the vectors have a large variance, the dot product can become very large, pushing the softmax function into regions with near-zero gradients, which can hinder training. The \textit{scaled multiplicative attention} $f_c^{m}(*)$ \cite{vaswani2017attention}addresses the issue of large dot-product values in multiplicative attention. By dividing the dot product of $Q_{s}$ and $K_{s}$ vectors by the square root of the dimension of the key vectors $\sqrt{D}$, it stabilizes the gradients and prevents the softmax function from saturating. This simple scaling factor is a crucial component of the standard Transformer model and is widely adopted due to its effectiveness in improving training stability and performance. \textit{Bilinear attention} $f_c^{b}(*)$ \cite{luong2015effective} introduces a weight matrix $W$ that is learned during training. This allows the model to learn a more complex, non-linear relationship between $Q_{s}$ and $K_{s}$ vectors, potentially capturing more intricate interactions. While it offers greater expressive power, the inclusion of $W$ increases the number of learnable parameters, making it computationally expensive and prone to overfitting on smaller datasets. Luong et al. \cite{luong2015effective} also proposed the \textit{Concat attention} that concatenates $Q_{s}$ and $K_{s}$ vectors and then passes the result through a feed-forward neural network to calculate the attention score. It utilizes additional learned weight matrix $W$, context vector $V_a$ and bias $b$. This aids in learning more complex, non-linear relationships between $Q_{s}$ and $K_{s}$. \textit{Additive or Bahdanau attention} $f_c^{a}(*)$ \cite{bahdanau2015neural} applies a feed-forward neural network with a single hidden layer to compute the attention scores based on learnable weight matrices $W_1$, $W_2$. Historically, this method was popular in sequence-to-sequence models before the rise of the Transformer architecture. It can capture more complex, non-linear dependencies between $Q_{s}$ and $K_{s}$. However, it is also more computationally intensive due to the additional parameters and the need for a non-linear activation function. 

Following this, the attention weights $a_w$ are obtained after being fed into a distribution function $f_{\delta}(*)$ to normalize the alignment scores and transform it into a probability distribution as follows:

\begin{equation}
\label{eq48}
a_w = f_\delta(e_a)
\end{equation}

The distribution function can have varied forms with softmax activation being the most widely used \cite{vaswani2017attention}. To obtain the attention-weighted representation of the input $Z'$, pairwise inner product between $V_{s}$ and $a_w$ is computed as follows:

\begin{equation}
\label{eq49}
Z' = a_w \cdot V_{s}
\end{equation}

$V_{s}$ represents the sequence vector upon which the attention weights are applied to determine the significant tokens. Finally, the attention-based context vector $C_a$ is obtained as the element-wise sum of $Z'$ such that elements with higher attention weights have more significance compared to lower attention weights as shown in Equation (\ref{eq50}).

\begin{equation}
\label{eq50}
C_a = \sum z'_{j} \textit{, } \forall z'_{j} \in Z'
\end{equation}

\begin{figure}
    \centering
    \includegraphics[width=0.6\linewidth]{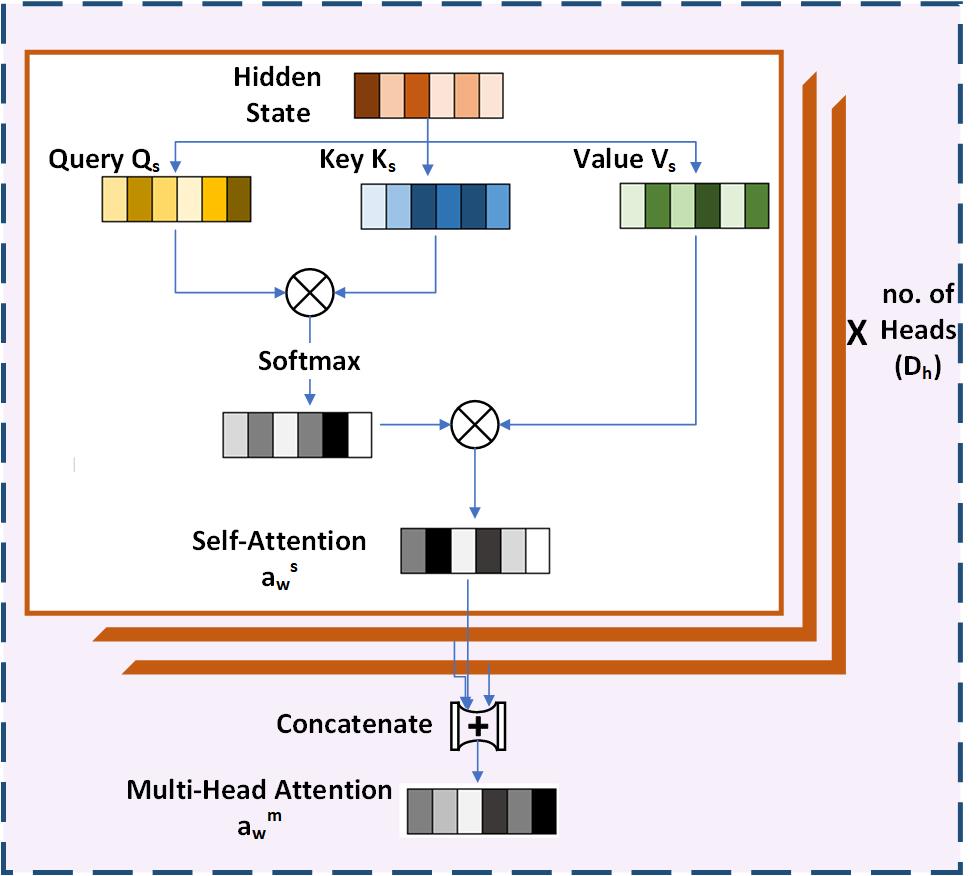}
    \caption{Illustration of the Multi-Head Attention}
    \label{fig10}       
    \end{figure}

Often, there is only one input sequence, and attention is computed solely on it. It gave rise to \textit{self-attention} or \textit{intra-attention}, a concept refined in many later works \cite{lin2017structured, kim2016structured}. This is accomplished by having the same vector for both $K_{s}$ and $Q_{s}$. In this manner, it helps to capture the relevance of a particular token in a sequence concerning other tokens in it. Furthermore, \textit{Multi-Head Attention (MHA)} (as illustrated in Figure \ref{fig10}) was devised to accommodate parallel computation of attention at diverse positions. It concatenates $a_w$ computations from $D_h$ attention heads and projects them through $W^{o} \in \mathbb{R}^{D} \times \mathbb{R}^{D}$ as shown in Equation (\ref{eq40}).

\begin{equation}
\label{eq40}
MHA_{w}=Concatenate(a_{w}[i])W^{o}, \forall{i \in D_h}
\end{equation}

Subsequent developments led to more efficient versions of MHA such as Multi-Query Attention (MQA) \cite{shazeer2019fast} which significantly reduces memory and computational overhead. While MHA uses separate weight matrices to create a unique set of keys and values for each attention head, MQA shares a single set of keys and values across all heads. This means that instead of creating a matrix for keys and values for each head, the model only computes one key matrix $K_{m}$ and one value matrix $V_{m}$ as portrayed in Equation (\ref{eq18}). However, queries are generated independently for all $D_h$ attention heads. Finally, MQA is computed (Equation \ref{eq19}) in a similar manner as in MHA but with the modified $K_{m}$ and $V_{m}$ matrices.

\begin{equation}
\label{eq18}
K_{m}=\frac{1}{d_h}\sum_{h \in D_h}{K_{s}^{h}}, \quad V_{m}=\frac{1}{d_h}\sum_{h \in D_h}{V_{s}^{h}}
\end{equation}

\begin{equation}
\label{eq19}
MQA_{w}=Concatenate(a_{w}[i])W^{o}, \forall{i \in D_h}
\end{equation}

where

\begin{equation}
\label{eq20}
a_w = f_\delta(f_c(Q_{s}, K_{m}))
\end{equation}

Grouped-Query Attention (GQA) \cite{ainslie2023gqa} further strikes a balance between the high performance of MHA and the memory efficiency of MQA. GQA addresses the potential performance degradation of MQA by not sharing the keys and values across all heads, but rather by sharing them across $g$ groups of heads such that $g \in (1<g<D_h) \cap (D_h (\text{mod } g)=0)$ as shown in Equation (\ref{eq21}).

\begin{equation}
\label{eq21}
\begin{split}
GQA_{w}=Concatenate(f_\delta(f_c(Q_{s}[i], K_{m}[j])))W^{o}, \\ \forall{i \in D_h \text{and} \forall{j \in g}}
\end{split}
\end{equation}

with the keys and values computed for the $j^{th}$ group as follows, 

\begin{equation}
\label{eq22}
K_{g}[j]=\frac{g}{D_h}\sum_{h \in j}{K_{s}^{h}}, \quad V_{g}[j]=\frac{g}{d_h}\sum_{h \in j}{V_{s}^{h}}
\end{equation}

Comparing the time and space complexities of MHA, MQA and GQA, it can be inferred that all three share the same time complexity of $\mathcal{O}(n_{k}D_hd_{k})$  because the fundamental operation of matrix multiplication between the query and keys for all the heads remain the same. However, the space complexity varies due to grouping of keys and values. This results in $\mathcal{O}(n_{k}D_hd_{k})$, $\mathcal{O}(n_{k}d_{k})$ and $\mathcal{O}(n_{k}gd_{k})$ for MHA, MQA and GQA respectively.

\subsubsection{Transformer Architecture}
\label{ta}

\begin{figure}
    \centering
    \includegraphics[width=0.6\linewidth]{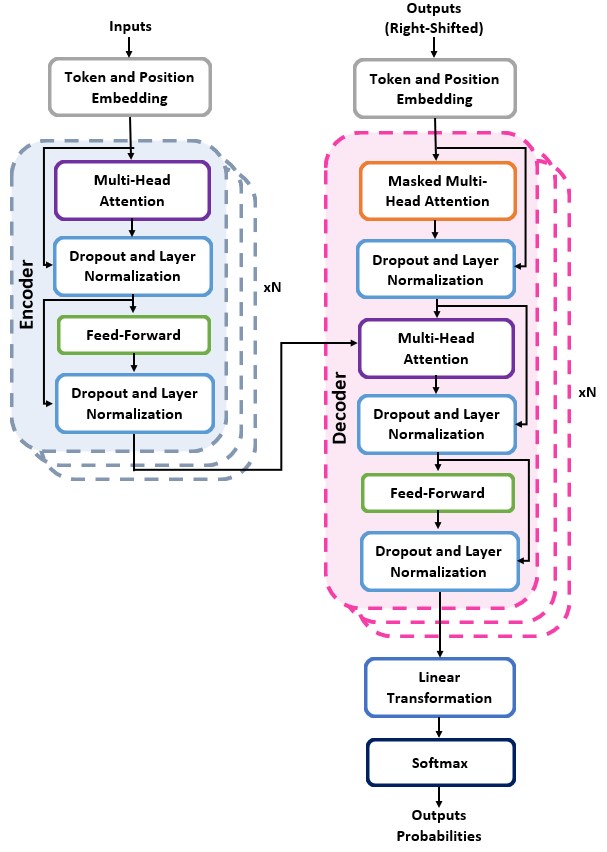}
    \caption{Illustration of the Transformer architecture (Inspired from Vaswani et al. \cite{vaswani2017attention})}
    \label{fig2}       
    \end{figure}

The Transformer architecture by Vaswani et al. \cite{vaswani2017attention} is a milestone achievement. It is a sequence-to-sequence model entirely based on a multi-head scaled multiplicative attention mechanism without any CNN or RNN units. It consists of a stack of encoder and decoder blocks as shown in Figure \ref{fig2}. The \textit{encoder block} processes each token in the input sequence and creates a contextual representation of it. The encoding process is an "understanding" phase in which the Transformer grasps the meaning of the input sequence. The \textit{decoder block} receives the encoder output, along with previously generated tokens, and produces the output sequence one token at a time. This is known as the "generation" phase. While the MHA block has been explained above, the remaining blocks have been described herein. 

The \textit{feed-forward block} consists of two linear layers with a non-linear activation function in between. It processes each token's representation independently and transforms the data from the attention layers, augmenting the LM's learning capability. The activation function can be Rectified Linear Unit (ReLU) (Equation \ref{eq23}) used in the first Transformer architecture \cite{vaswani2017attention}; Gaussian Error Linear Unit (GeLU) (Equation \ref{eq24}) found in BERT \cite{kenton2019bert}, GPT-2 \cite{radford2019language}, GPT-3 \cite{brown2020language},and Chinchilla \cite{hoffmann2022training}; Switched Gated Linear Unit (SwiGLU) (Equation \ref{eq25}) found in PaLM \cite{chowdhery2023palm}, LaMDA \cite{thoppilan2022lamda}, LLaMA \cite{touvron2023llama}); or Gated GELU (GeGLU) (Equation \ref{eq26}) found in T5 \cite{raffel2020exploring}). GeGLU combines gating with GeLU for better control over information flow. Whereas, SwiGLU is a GeLU variant that uses the Swish activation function to offer a smooth, non-monotonic function that can capture more complex patterns in data and improve the training stability in LLMs. 

\begin{equation}
\label{eq23}
ReLU(x)=max(0,x)
\end{equation}

\begin{equation}
\label{eq24}
GELU(x)=x \cdot \phi(x)
\end{equation}

\begin{equation}
\label{eq25}
SwiGLU(x)=(xU) \otimes (xW \cdot \frac{1}{1+e^{-xW}})
\end{equation}

\begin{equation}
\label{eq26}
GeGLU=(xU) \otimes GELU(xV)
\end{equation} 

where $U$ and $V$ are learnable weight matrices. $\phi(x)$ is the cumulative distribution function (CDF) of the standard normal distribution

The \textit{residual connections} serve as "shortcuts" that allow information to bypass a layer and be added to its output. This helps to prevent the vanishing-gradient problem, where gradients become too small to effectively update the model's weights during training.

Moreover, \textit{layer normalization} is applied upon the output of the intermediate blocks. It works by computing the mean and variance of the activations across the features of a single data point. This helps to stabilize the training process and make the model more robust to input data variations. While Layer Normalization is the standard in Transformers, modern LLMs have adopted other normalization techniques, primarily Root Mean Square Normalization (RMSNorm) (used in LLaMA 2 \cite{touvron2023llama2}, LLaMA 3 \cite{dubey2024llama} and Mistral 7B \cite{jiang2023mistral7b}). Instead of calculating both mean and variance of the activations as in layer normalization, it scales them only by their Root Mean Square (RMS). This reduces computational overhead, posing a significant advantage for training and inference with massive models containing billions of parameters. To further prevent vanishing or exploding gradients, LLaMA 2 uses pre-normalization, i.e. applies normalization before the MHA or feed-forward blocks contrary to conventional post-normalization.

Overall, the Transformer provides exceptional sequence representation and supports parallel training, unlike the LSTM-based methods. The scaling of Transformers based on the number of model parameters, the size of the training dataset, and the computational resources improved its capabilities following power-law relationships (explained in Section \ref{sol}). This led to the genesis of Transformer-based LLMs that has transformed the field of NLP. Such LLMs pre-trained on large datasets just need to be fine-tuned as per the application. One of the first foundational LLMs was OpenAI's Generative-Pre-trained Transformer (OpenAI GPT) based upon \textit{Transformer-decoder architecture} with unidirectional context parsing. Subsequently, Bidirectional Encoder Representations from Transformers (BERT) \cite{kenton2019bert} adopted bidirectional context-parsing deploying a \textit{Transformer-encoder architecture}. However, BERT has drawbacks like the exclusion of "Mask" tokens during fine-tuning, and parallel predictions without dependency consideration. These drawbacks were resolved by XLNet through "permutation language modeling" wherein the prediction tokens are permuted randomly \cite{yang2019xlnet}. The successors of OpenAI GPT i.e. GPT-2 \cite{radford2019language} and GPT-3 \cite{brown2020language} further enhance the performance and reusability with the concept of "In-Context Learning" (ICL). It allows the model to be conditioned with just a few instances or description of the application. Besides, LLMs have been devised utilizing the entire \textit{Transformer encoder-decoder architecture} such as T5 Transformer \cite{raffel2020exploring}, PEGASUS \cite{zhang2020pegasus} and BART \cite{lewis2020bart}. Along with these, several other LLMs have been systematically discussed in this paper.

\begin{table}[t!]
\centering
\caption{Parameter and compute estimates for a Transformer model}
\label{table:table5}
\begin{tabular}{lll}
\hline
\textbf{Operation} & \textbf{Parameters} & \textbf{FLOPs per Token} \\ \hline
Embedding & $(n_{\text{vocab}} + n_C) d_M$ & $4 d_M$ \\
Attention- QKV & $n_L d_M 3 d_A$ & $2 n_L d_M 3 d_A n_C$ \\
Attention- Mask & $-$ & $2 n_L n_C d_A$ \\
Attention- Project & $n_L d_A d_M$ & $2 n_L d_A d_M$ \\
Feedforward & $n_L 2 d_M d_F$ & $2 n_L 2 d_M d_F$ \\
De-embedding & $2 d_M n_{\text{vocab}}$ & $2 d_M n_{\text{vocab}}$ \\ \hline
\textbf{Total (Non-Embedding)} & $\boldsymbol{\bar{N} = 2 d_M n_L(2 d_A + d_F)}$ & $\boldsymbol{C_F \approx 2\bar{N} + 2 n_L n_C d_M}$ \\ \hline
 \multicolumn{3}{p{0.72\linewidth}}{Source: Based on a study by Kaplan et al. \cite{kaplan2020scaling}}\\
\end{tabular}
\end{table}

\subsection{RQ2: Scaling of LLMs}
\label{sol}
The advancement of LLMs has been propelled by the principle of "up-scaling" to improve performance. Notable studies have demonstrated that increasing model parameters, dataset size, and compute leads to gains in performance \cite{kaplan2020scaling, hoffmann2022training}. However, scaling incurs huge costs in model adaptation and inference due to the requirement of plenty of sophisticated computational resources (GPUs, TPUs, storage devices, network infrastructure, etc.) along with heightened energy consumption and carbon footprint \cite{sengupta2025position}. By the year 2030, data centers supporting LLMs is set to consume nearly 945 Tera Watt-hours of electricity, exceeding the total electricity consumption of Japan\footnote{\url{https://www.wsj.com/articles/google-wants-you-to-know-the-environmental-cost-of-quizzing-its-ai-143cfe19}}. In the same study, it was reported that the emissions from Google's AI data centers have increased by 51\% since 2019. Besides, recent studies have noted "diminishing returns" of scaling. This leads to a reduced gain in LLM performance after a certain point despite scaling by the same factor \cite{diaz2024scaling}. Morevover, the present LLMs have already used up almost the entire textual data available on the internet leaving very little scope for increasing the training data for LLMs. In this scenario, increasing compute without adding more high-quality, unique human-generated data provides suboptimal gains \cite{muennighoff2023scaling}. In the remainder of this section, the scaling laws, their implications as well as efforts towards down-scaling LLMs have been discussed.

\subsubsection{Transformer Parameters and Compute Calculation}
\label{tpf}

Kaplan et al. \cite{kaplan2020scaling} presented a relationship between the size and compute of a Transformer model in terms of number of parameters and FLOPs respectively. In their study, they defined several key hyperparameters: $n_C$, $n_L$, $d_M$, $d_F$, $d_A$, and $n_H$ denoting the number of input tokens, number of layers, residual stream dimensions, intermediate feed-forward layer dimensions, attention output dimensions and the number of attention heads respectively. Based on the multiply-accumulate operations in the network with the given hyperparameters, the total number of non-embedding parameters $\bar{N}$ and FLOPs for a single forward pass $C_F$ is estimated in Equations \ref{eq1} and \ref{eq2} respectively. Supporting it, Table \ref{table:table5} provides a break-up of the computation.

\begin{equation}
\label{eq1}
\bar{N} \approx 2 d_M n_L(2 d_A + d_F)
\end{equation}

\begin{equation}
\label{eq2}
C_F \approx 2\bar{N} + 2 n_L n_C d_M
\end{equation}

\subsubsection{Laws for Scaling}
\label{lws}

The insights from the power laws of scaling laid the empirical foundation for scaling strategies that now underpin the development of SOTA LLMs. Kaplan et al. \cite{kaplan2020scaling} utilized "power-law" to predict the test loss exclusively constrained on $\bar{N}$, size of data $S$ and optimal compute budget $c_{min}$ as shown in Equations (\ref{eq3}), (\ref{eq4}) and (\ref{eq5}) respectively.

    \begin{equation}
    \label{eq3}
    L(\bar{N}) = (\bar{N}_c/\bar{N})^{\alpha_{\bar{N}}}; \quad \alpha_{\bar{N}} \sim 0.076, \quad \bar{N}_c \sim 8.8 \times 10^{13}
    \end{equation}

    \begin{equation}
    \label{eq4}
    L(S) = (S_c/S)^{\alpha_S}; \quad \alpha_S \sim 0.095, \quad S_c \sim 5.4 \times 10^{13}
    \end{equation}

    \begin{equation}
    \label{eq5}
    \begin{split}
    L(C_{\text{min}}) = (C_{\text{c}}^{\text{min}} / C_{\text{min}})^{\alpha_C^{\text{min}}}; \quad \alpha_C^{\text{min}} \sim 0.050, \\ C_{\text{c}}^{\text{min}} \sim 3.1 \times 10^{8}
    \end{split}
    \end{equation}

Combining Equations \ref{eq3}, and \ref{eq4}, the loss based on both $\bar{N}$ and $S$ is computed as follows:

\begin{equation}
\label{eq7}
L(\bar{N},S) = \left[\left(\frac{{\bar{N}}_{c}}{\bar{N}}\right)^{\alpha_{\bar{N}}} + \frac{S_c}{S} \right]^{\alpha_S}
\end{equation}

Based on it, it was observed that the performance of LM improves proportionally with size, data, and compute, following a power-law relationship. Larger models not only achieve lower loss but also become more sample-efficient, requiring fewer data to reach the same performance as smaller ones. The study finds that for a fixed compute budget, it's more effective to train larger models for fewer steps than to fully train smaller ones—highlighting the benefits of under-training. Architectural tweaks like depth and width have relatively minor effects compared to scaling.

Rae et al. \cite{rae2021scaling} assessed scalability across a wide range of LLMs having parameters from a few millions to a 280B LLM- Gopher proposed by them. They highlighted that scaling LMs brings performance gains across most tasks but also exposes efficiency trade‑offs such as the balance between model size and dataset scale. Gopher being trained on fewer tokens relative to its size, made it compute‑inefficient compared to smaller models trained on more data. This revealed that scaling laws hold, but optimal scaling requires aligning parameters, data, and compute budgets rather than maximizing size alone.

In a subsequent study by Hoffmann et al. \cite{hoffmann2022training} it was stated that to achieve compute-optimal training, model size and training data size should grow in tandem, i.e. doubling the number of parameters should be matched by doubling the training tokens. They observed that LLMs like GPT-3 were under-trained with respect to their size. This was further demonstrated by training a 70B parameter model (Chinchilla) on more data which outperformed an LLM having 280B parameters. Their claims were supported by scaling laws named as "Chinchilla Law" to compute the loss $L_{c}(N,S)$ with $N$ and $S$ denoting the number of model parameters and size of training data as shown in Equation (\ref{eq6}).

\begin{equation}
\label{eq6}
L^{c}(N,S) \triangleq E + \frac{A}{N^\alpha} + \frac{B}{S^\beta}
\end{equation}

The parameters $A, B, E, \alpha$, and $\beta$ are estimated by minimizing the Huber loss as follows: 

\begin{equation}
\label{eq8}
\min_{A,B,E,\alpha,\beta} \sum_{\text{Runs } i} \text{Huber}\left( \log L^{c}(N_i, S_i) - \log L^c_i \right)
\end{equation}

After this, the optimal compute $C$ allocation to $N$ and $S$ is done as follows:

\begin{equation}
\label{eq9}
N_{\text{opt}}(C) = G(C/6)^a, \quad S_{\text{opt}}(C) = G^{-1}(C/6)^b 
\end{equation}

\begin{equation}
\label{eq35}
\text{where }
G = \left( \frac{\alpha A}{\beta B} \right)^{\frac{1}{\alpha+\beta}}, \quad a = \frac{\beta}{\alpha+\beta}, \quad b = \frac{\alpha}{\alpha+\beta} \nonumber
\end{equation}

It shows that $N$ and $S$ need to be scaled in almost equal ratio for optimal compute utilization given that $\alpha \approx \beta$.

Traditional scaling laws assume unlimited data, but real-world scenarios often face data scarcity. For this, Muennighoff et al. \cite{muennighoff2023scaling} explored how to train LLMs effectively when the amount of available training data is limited. They commented that repeating data for up to four training epochs can curb loss similar to having unique data, but returns diminish sharply beyond that due to overfitting. They propose a revised version of the Chinchilla Law (Equation \ref{eq6} that accounts for these diminishing returns by incorporating additional parameters such as unique data budget $S_{B}$, unique tokens count $U_{S}=min\{S_{B}, S\}$, and repetition count $R_{S}=\frac{S}{U_{S}}-1$ as shown in Equation (\ref{eq10}).

\begin{equation}
\label{eq10}
L^{m}(N,S) \triangleq E + \frac{A}{N'^\alpha} + \frac{B}{S'^\beta}
\end{equation}

Based on optimization through an exponential decay function, the effective values of model parameters $N'$ and data size $S'$ is computed as follows:

\begin{equation}
\label{eq11}
N' = U_N + U_N R_N^*(1 - e^{-\frac{R_N}{R_N^*}}), \quad S' = U_S + U_S R_S^*(1 - e^{-\frac{R_S}{R_S^*}})
\end{equation}

where $U_N=min\{N_{opt}, N\}$ is the optimum computational budget for distinct tokens and $R_{N}=\frac{N}{U_{N}}-1$ counts the repetitions. The half-life of repeated data and excess parameters is approximately represented by the constants $R_N^*, R_S^*$ learned during optimization. This leads to more resource-aware approaches to scaling, especially in low-data regimes.

Diaz and Madaio \cite{diaz2024scaling} examined the validity of the scaling laws rooted in "bigger-is-better" ideology. They criticized them as not being universal truths but rather shaped by narrow evaluation metrics that often reflect the values of dominant groups. As datasets grow to include more diverse populations, these metrics may fail to capture what matters to different communities, leading to models that overlook or even harm marginalized voices. The authors caution that scaling up can amplify biases and toxic outputs, and they advocate for a shift toward pluralistic, context-sensitive evaluations and smaller, community-aligned models that prioritize ethical and inclusive design over sheer size.

Villalobos et al. \cite{villalobos2024position} discovered that as LLMs scale in size and compute, the need for clean, diverse, and high-quality human-generated textual data increases exponentially. But the internet’s ability to supply such massive amount of data might soon be exhausted. They estimated the total usable public text data to be around $300\times 10^{12}$, tokens and if current rate of scaling continues, this reserve could be exhausted between the year 2026 and 2032. The paper models different scenarios for dataset growth and projects it as a cumulative distibution function as follows:

\begin{equation}
\label{eq12}
F_{S(x)} = \frac{1}{2}\left(F_{S_{H}(x)} + F_{S_{C}(x)}\right)
\end{equation}
where $S_H(x)$ and $S_C(x)$ denote the historical and computational projection of data sizes calculated as follows:

\begin{equation}
\label{eq27}
S_{H}(x) = G_{D}^{x-x_{0}}D(x_{0}), \quad S_{C}(x) = \sqrt{\frac{20}{6}.C(x)}
\end{equation}

where $G_{D}, x, x_{0}, D(\cdot)$ and $C(x)$ denote the growth-factor per year, the given year, the base year, a lognormal distribution and the probabilistic estimate of the maximum compute for a training session. The authors highlight that without new sources of data or more efficient training methods, scaling LLMs could hit a hard ceiling. This raises concerns about the future of LLM scaling and underscores the need for innovations in data efficiency, synthetic data generation, and alternatives like code or multimodal inputs.

\subsubsection{Estimating Energy Usage and Carbon Footprint}
\label{ecp}
Several studies aimed at estimating energy usage and carbon footprint have been reviewed herein. Strubell et al. \cite{strubell2019energy} calculated the power consumption, carbon emissions, along with the monetary cost associated with training a set of NLP models. They found that training a NLP model can cost as much as a trans-Atlantic flight. Anthony et al. \cite{wolff2020carbontracker} developed Carbontracker, a tool that monitors CPU, GPU, and DRAM statistics to calculate power consumption and estimate the resulting carbon emissions. Jiang et al. \cite{jiang2024preventing} examined the energy consumption and carbon emissions for the overall life-cycle of intelligent chatbots powered by LLMs. They identified eight life-cycle stages, i.e. research and development, hardware production, logistics, data curation and maintenance, LLM adaptation, provision of end-user services, recycling and disposal of waste.  The paper emphasizes that the electricity demand for training and upkeep of these models is immense, and the associated carbon emissions are often overlooked in the race for innovation. To address this, they propose three major pathways: comprehensive and continuous assessment covering all life-cycle stages, providing incentives for studying the environmental impacts along with choosing greener solutions, and fostering global cooperation for standardized benchmarks. LLMCO2 \cite{fu2025llmco2} presents a GNN–based model to estimate the carbon emissions during LLM inference. It was seen that inference stage emissions often surpass those during the training stage due to widespread adoption and huge scale of user interaction. The traditional estimation methods were noted to falter due to disregarding the dynamic nature of inference workloads. LLMCO2 addresses this by modeling the prefill as well as decode phases of inference, hardware configurations and incorporating realistic distributions of request. This resulted in approximately 67\% better accuracy than prior approaches. 

Lannelongue et al. \cite{wolff2020carbontracker} proposed a methodological framework to calculate energy consumption based on factors such as runtime, hardware type (CPU/GPU/TPU), memory usage, and power-usage effectiveness (PUE) of data center. The total energy consumption $E_t$ (in kWh) is estimated as: 

 \begin{equation}
\label{eq34}
E_t= T_{r} \times (n_{c} \times P_{c} \times U_{c} \times + n_{m} \times P_{m}) \times PUE \times 0.001
\end{equation}

where $T_{r}$, $n_c$, $n_m$, $u_c$, $P_c$ and $P_m$ denote the run time (in hours), number of computing cores,  available memory (in GB), core utilization rate ($u_c \in [0,1]$), power consumption (in W) of a single core, power consumption (in W) of the memory. Here, Power-Usage Effectiveness ($PUE$) is specific to the deployment scenario and is computed as follows: 

 \begin{equation}
\label{eq32}
PUE= \frac{P_t}{P_e}
\end{equation}

where $P_t$ and $P_e$ denote the total power delivered to the data centre, and the power consumed by the equipment. Based on this, the carbon emission $C_e$ is determined by the Equation (\ref{eq33}).

 \begin{equation}
\label{eq33}
C_e= E_t \times CI
\end{equation}

Here, carbon intensity $CI$ accounts for the amount of carbon emission per kWh of electricity generated. 

LLMCarbon \cite{faiz2024llmcarbon} introduces a comprehensive framework for estimating LLM operational emissions based on runtime energy consumption and embodied emissions from chip fabrication, transportation, and end-of-life disposal. It accounts for architectural parameters, hardware efficiency, data center energy profiles, and semiconductor manufacturing impacts. The total carbon emission $CE$ is computed as a sum of operational as well as embodied emissions $CE_{\text{op}}$ and $CE\text{emb}$ respectively as shown in Equation (\ref{eq16}). 

\begin{equation}
\label{eq16}
CE = CE_{\text{op}} + CE\text{emb}
\end{equation}

such that, 

\begin{equation}
\label{eq17}
CE_{\text{op}}=\sum_{i \in hardware}{(P_i \cdot eff_i \cdot n_i \cdot T_i)} \cdot \text{PUE} \cdot \text{CI}
\end{equation}

\begin{equation}
\label{eq36}
CE_{\text{emb}}=\sum_{i \in hardware}{\frac{T_i \cdot A_i \cdot CPA_i}{LS_i}}
\end{equation}

where $P_i$, $eff_i$, $n_i$, $T_i$, $A_i$, $CPA_i$ and $LS_i$ denote the peak power, hardware efficiency, count, processing time, chip area, $\text{CO}_{2}$ emission per unit area and lifespan for $i^{th}$ hardware. This facilitates developers and cloud providers to make environmentally informed decisions before committing to resource-intensive LM development.

Jegham et al. \cite{jegham2025hungry} laid a comprehensive framework for evaluating the environmental impact of LLM inference across 30 commercially deployed SOTA LLMs. By combining publicly available performance data of APIs, environmental factors specific to regions, and statistically-inferred configurations of hardware, they estimated energy use per query $E_{\text{per\_query}}$, water consumption $W_{c}$, and carbon emissions $C_{e}$ as depicted in Equations \ref{eq13}, \ref{eq14} and \ref{eq15} respectively.

\begin{equation}
\label{eq13}
\begin{split}
    E_{\text{per\_query}} (\text{kWh}) = \left( \underbrace{\frac{\frac{L_{o}}{\text{TPS}} + T_{l}}{3600}}_{Inference Time} \right) \\
    \cdot \left( \underbrace{P_{\text{GPU}} \times U_{\text{GPU}}}_{\text{GPU Power (kW)}} + \underbrace{P_{\text{non-GPU}} \times U_{\text{non-GPU}}}_{\text{Non-GPU Power (kW)}} \right) \cdot \text{PUE}
\end{split}
\end{equation}

where $L_{o}$ is the length of output, TPS is the tokens per second, and $T_{l}$ is the latency. $P_{\text{GPU}}$ and $U_{\text{GPU}}$ is the maximum as well as aggregate GPU power requirement while $P_{\text{non-GPU}}$ and $U_{\text{non-GPU}}$ represent the maximum and aggregate non-GPU component power requirement.

\begin{equation}
\label{eq14}
W_{c}\text{(L)} = \underbrace{\frac{E_{\text{per\_query}}}{\text{PUE}} \cdot \text{WUE}_{\text{site}}}_{\text{On-Site Cooling}} +  \underbrace{E_{\text{per\_query}} \cdot \text{WUE}_{\text{source}}}_{\text{Off-Site Electricity}}
\end{equation}

\begin{equation}
\label{eq15}
C_{e}\text{(kgCO}_2\text{e)} = E_{\text{query}} \cdot \text{CIF}
\end{equation}

where $\text{WUE}$ and $\text{CIF}$ denotes water usage and carbon emissions per kWh of energy utilized respectively. Their findings reveal that despite efficient individual queries, the global scale of usage, i.e. nearly 700M queries per day translates into annual electricity consumption equivalent to 35,000 U.S. homes, freshwater evaporation matching the drinking needs of 1.2M people, and carbon emissions requiring a forest the size of Chicago to offset.

\begin{figure*}
    \centering

\tikzset{
    basic/.style  = {draw, text width=2cm, align=center, fill=blue!20, font=\sffamily, rectangle},
    root/.style   = {basic, rounded corners=2pt, thin, align=center, fill=green!30},
    onode/.style = {basic, thin, rounded corners=2pt, align=center, fill=green!90!yellow!10,text width=2cm,},
    tnode/.style = {basic, thin, align=left, fill=pink!60, text width=4.5cm, align=center},
    xnode/.style = {basic, thin, rounded corners=2pt, align=center,  fill=pink!10!blue!80!red!10, text width=2cm,},
    wnode/.style = {basic, thin, align=left, fill=pink!20, text width=3cm},
    edge from parent/.style={draw=black, edge from parent fork right}

}
\scalebox{0.75}{%
\begin{forest} for tree={
    grow=east,
    growth parent anchor=west,
    parent anchor=east,
    child anchor=west,
    edge path={\noexpand\path[\forestoption{edge},->, >={latex}] 
         (!u.parent anchor) -- +(6pt,0pt) |-  (.child anchor) 
         \forestoption{edge label};}
}
[Model Optimization, basic,  l sep=6mm,
	[Data Curation, xnode,  l sep=6mm,
		[Data De-Duplication, tnode]
		[Data Subset Selection (DSS), tnode]
		[Active Learning (AL), tnode]
		[Curriculum Learning (CuL), tnode]
		[Dimensionality Reduction, tnode]]
	[Model Design, xnode,  l sep=6mm,
		[Chunking, tnode]
		[Mixture-of-Experts, tnode]
		[Low-Rank Approximation, tnode]
		[Clustering, tnode]
		[Parameter Sharing, tnode]]
	[Model Downsizing, xnode,  l sep=6mm,
		[Pruning, tnode, l sep=6mm,
            [Structured Pruning, wnode]
			[Unstructured Pruning, wnode]
			[Contextual Pruning, wnode]]
		[Knowledge Distillation (KD), tnode, l sep=6mm,
            [White-box KD, wnode]
			[Black-Box KD, wnode]]
		[Quantization, tnode, l sep=6mm,
            [Post-Training Quantization (PTQ), wnode,l sep=6mm,
				[Weight-only PTQ, onode]
				[Weight-Activation PTQ, onode]]
			[Quantization-Aware Training (QAT), wnode]]]
	[Dynamic Inferencing, xnode,  l sep=6mm,
		[Early Exit (EE), tnode]
		[Token Pruning (TP), tnode, l sep=6mm,
            [Value-Based Scoring, wnode]
			[Predictive Modeling, wnode]
			[Token Parallelism (TPA), wnode]]]]     
\end{forest}
}
    \caption{Taxonomy of model optimization strategies}
    \label{fig:lit_surv2}
\end{figure*}

\section{Efficient Modeling Considerations}
\label{emc}

As can be inferred from the preceding Section, the exorbitant requirement of computational resources and training data by LLMs ushers in the need for efficient modeling practices. The remainder of this section discusses the efficient modeling considerations grouped under model optimization and LLM adaptation.

\subsection{RQ3: Model Optimization}
\label{mo}
To enhance the efficiency of NLP models, numerous optimization strategies have been devised to target various stages of model development as highlighted in Figure \ref{fig:lit_surv2}. In this section, commentary on such techniques grouped under data curation, model design, model downsizing, and dynamic inferencing have been presented.

\subsubsection{Data Curation}
\label{dc}
Data curation plays a vital role in the efficiency of an LLM. A dataset with reduced sequence lengths or less number of training samples minimizes the model complexity and reduces the training effort significantly \cite{hoffmann2022empirical}. The popular approaches for efficient data curation have been discussed herein-below, while Figure \ref{fig:lit_surv4} presents a taxonomy of the developments.

\begin{figure*}
    \centering

\tikzset{
    basic/.style  = {draw, text width=2cm, align=center, fill=pink!10!blue!80!red!10, font=\sffamily, rectangle},
    root/.style   = {basic, rounded corners=2pt, thin, align=center, fill=green!30},
    onode/.style = {basic, thin, rounded corners=2pt, align=center, fill=green!60,text width=2cm,},
    tnode/.style = {basic, thin, align=left, fill=pink!0, text width=9.3cm, align=left},
    xnode/.style = {basic, thin, rounded corners=2pt, align=center,  fill=pink!60, text width=3.6cm,},
    wnode/.style = {basic, thin, align=left, fill=pink!10, text width=6cm},
    edge from parent/.style={draw=black, edge from parent fork right}

}
\scalebox{0.75}{%
\begin{forest} for tree={
    grow=east,
    growth parent anchor=west,
    parent anchor=east,
    child anchor=west,
    edge path={\noexpand\path[\forestoption{edge},->, >={latex}] 
         (!u.parent anchor) -- +(6pt,0pt) |-  (.child anchor) 
         \forestoption{edge label};}
}
	[Data Curation	, basic,  l sep=6mm,
		[Data De-Duplication, xnode, l sep=6mm,
            [CCNet \cite{wenzek2020ccnet}; 
 	 	 	Generalized Entropy \cite{mitchell2022measuring}; 
 	 	 	D4 \cite{tirumala2023d4}; 
 	 	 	LeSum \cite{ansar2026lesum}; 
 	 	 	IRS Monitoring \cite{yin2024entropy}; 
 	 	 	Prompt Engineering and Sequential Optimization\cite{guang2024data}; 
 	 	 	Calibrating Lightweight Models \cite{cunha2025noise} , tnode]] 
		[Data Subset Selection (DSS), xnode, l sep=6mm,
            [Ask-LLM \cite{sachdeva2024train}; 
 	 	 	SelectLLM \cite{parkar2024selectllm}; 
 	 	 	SubLIME \cite{saranathan2025sublime}; 
 	 	 	LAMDAS \cite{wu2025lamdas}; 
 	 	 	I3S \cite{zhang2025i3s}, tnode]] 	
		[Active Learning (AL), xnode, l sep=6mm,
            [Learning from Unsupervised Pre-Training \cite{yuan2020cold}; 
 	 	 	Learning from Contrastive Examples \cite{margatina2021active}; 
 	 	 	ICL for AL \cite{margatina2023active}; 
 	 	 	Collaborating LLMs with Human Annotators \cite{rouzegar2024enhancing}; 
 	 	 	NoiseAL \cite{yuan2024hide}; 
 	 	 	AnchorAL \cite{lesci2024anchoral} , tnode] ] 	
		[Curriculum Learning (CuL), xnode, l sep=6mm,
            [Curri-DPO \cite{pattnaik2024enhancing}; 
 	 	 	Phi-4 \cite{abdin2024phi} ; 
 	 	 	Qwen-3 \cite{yang2025qwen3} ; 
 	 	 	Xu et al. \cite{xu2024context} ; 
 	 	 	EASE \cite{wu2024prompt} ; 
 	 	 	Ordering Strategies in ICL \cite{guo2024makes}, tnode]] 
		[Dimensionality Reduction, xnode, l sep=6mm,
            [TexIm \cite{ansar2023texim}; 
 	 	 	TexIm FAST \cite{ansar2025texim}; 
 	 	 	PCA in Multimodal LLMs \cite{martin2024larger}; 
 	 	 	Embedding Compression \cite{ignacio2025unet}, tnode]]
			]
\end{forest}
}

    \caption{Taxonomy of the data curation strategies}
    \label{fig:lit_surv4}
\end{figure*}

\begin{itemize}
    \item \textbf{Data De-Duplication:} It enhances the efficiency of an LLM and improves its performance compared to training on the entire corpus \cite{lee2022deduplicating}. In the case of pre-trained LMs, such filtering can be applied both during the pre-training \cite{zhang2022opt} as well as the fine-tuning stages \cite{mishra2020we}. CCNet \cite{wenzek2020ccnet} introduces a language identification and quality filtering mechanism using perplexity scores from LMs, enabling the extraction of cleaner and more diverse data across multiple languages. It offers a scalable solution for filtering and de-duplicating massive web-crawled corpora. Mitchell et al. \cite{mitchell2022measuring} utilized generalized entropy indices to assess the distributional richness of samples, arguing that high-entropy data tends to offer more learning potential. Document De-Duplication and Diversification (D4) \cite{tirumala2023d4} identifies duplicate data samples that fall within a small epsilon-radius ball of one another in the embedding space, effectively capturing semantic similarity through spatial closeness. It leverages the intuition that redundant samples tend to cluster tightly in high-dimensional representations, offering a straightforward yet powerful mechanism for filtering excessive data. Moreover, this can be extended to decomposing the individual text sequences into smaller sub-sequences and evaluating redundancy among them through a graph-based salience estimation to retain only the essential information discarding the irrelevant portions \cite{ansar2026lesum}. Yin et al. \cite{yin2024entropy} maintained an “Information Redundancy State (IRS),” quantifying the incremental information gain by each sample. Through this, they filtered low-value samples that contribute minimally to model learning. Guang et al. \cite{guang2024data} investigated the use of prompt-engineering and sequential optimization to identify and eliminate redundant training samples. By iteratively refining prompts and analyzing model responses, they developed a dynamic filtering mechanism that adapts to the evolving learning state of the model, to streamline training without compromising performance. Cunha et al. \cite{cunha2025noise} introduced a lightweight and scalable framework based on the idea that if a simple model, such as logistic regression can accurately predict certain samples, those samples may not contribute significant new information to the training process. By calibrating these classifiers and analyzing their confidence levels, a fast and interpretable way to assess the informational value of individual data points was developed. Although filtering eliminates biases inherent in the dataset, their application is restricted to cases with abundant data as the performance reduces with insufficient data \cite{le2020adversarial}.

    \item \textbf{Data Subset Selection (DSS)}: It aims to identify the most informative and task-relevant samples that contribute meaningfully to LLM learning process. Sachdeva et al. \cite{sachdeva2024train} presented two innovative DSS strategies: Ask-LLM and Density Sampling. Ask-LLM uses instruction-tuned LLMs to evaluate and rank training samples based on their quality in a zero-shot setting, enabling the selection of high-impact data without manual heuristics. Density Sampling complements this by ensuring diverse and representative coverage of the data distribution, helping LLMs generalize better with fewer examples. Together, these methods outperform traditional full-data training, achieving faster convergence and higher accuracy while discarding up to 90\% of the original dataset. SelectLLM \cite{parkar2024selectllm} ranks unlabeled instructions based on their estimated utility. Without access to ground-truth labels, it relies on the model’s internal representations and reasoning capabilities. It demonstrates that LLMs can predict instruction importance with high fidelity, enabling more efficient use of annotation budgets. SubLIME \cite{saranathan2025sublime} selects compact, high-quality subsets of evaluation data that preserve the relative performance rankings of LLMs. It predicts how well a small subset can approximate the full evaluation rankings using rank correlation metrics like Kendall’s tau. It trains a meta-model to estimate this correlation based on features of the subset and the models being evaluated. This proves to be effective for rapid benchmarking and iterative development of LLMs, where full-scale evaluation is impractical. LAMDAS \cite{wu2025lamdas} prompts instruction-tuned LLMs with task descriptions and candidate samples, then interprets their responses to determine domain alignment. This implicit classification approach enables efficient and scalable data selection without labeled examples or fine-tuning.

    Apart from selecting samples altogether, DSS can be more fine-grained by operating on individual features or embedding components. Importance Sampling Subspace Selection (I3S) \cite{zhang2025i3s} combines importance sampling with low-rank optimization. The core idea is that not all data dimensions contribute equally to learning, and by identifying a subspace of high-impact dimensions influencing the LLM’s gradient updates through importance sampling, pre-training can be made more efficient.

    \item \textbf{Active Learning (AL):} While duplicate removal applies to archived datasets, active learning comes into play while collecting data. It aims to reduce the training data while retaining model performance by labeling the most informative samples and selecting them for training \cite{ren2021survey}. For the identification of informative samples, various approaches have been adopted. Yuan et al. \cite{yuan2020cold} addressed the challenge of how to initiate AL effectively when labeled data is scarce or nonexistent. Traditional AL relies on an initial labeled seed set to begin querying, but they proposed a cold-start strategy using self-supervised LM. By leveraging the representations learned from unsupervised pre-training, their method identifies informative examples for labeling without requiring any initial annotations. Margatina et al. \cite{margatina2021active} introduced an AL strategy focusing on contrastive examples, i.e. instances that are semantically similar but belong to different classes. Their approach aims to improve model generalization by challenging the decision boundaries more effectively than traditional uncertainty or diversity sampling. By leveraging sentence embeddings and contrastive sampling, it identifies pairs of examples that are close in representation space yet have different labels, encouraging the model to learn fine-grained distinctions. It demonstrated better performance on classification tasks, especially in low-resource setting. Margatina et al. \cite{margatina2023active} further reformulated AL in the context of LLMs by treating ICL as a form of pool-based AL, where the goal is to select the most informative examples from a pool to maximize performance on unseen data. Instead of relying on traditional uncertainty or diversity sampling, they demonstrated that choosing examples based on semantic similarity to the target input leveraging static embeddings significantly boosts performance across NLU tasks. Rouzegar and Makrehchi \cite{rouzegar2024enhancing} proposed a hybrid framework based on collaborating LLMs with human annotators in an AL loop to improve text classification performance while reducing annotation costs. It leverages uncertainty sampling to identify the most informative examples for annotation, and then intelligently balances cost and reliability by routing low-uncertainty samples to LLMs and reserving high-uncertainty ones for human experts. NoiseAL \cite{yuan2024hide} tackles the challenge of learning from noisy labels in real-world datasets through a collaborative strategy that combines SLMs and LLMs within an AL loop. Initially, two SMs form a co-prediction network to estimate label noise and partition the data into subsets based on confidence. A dynamic thresholding mechanism helps identify clean versus noisy samples. These subsets are then selectively routed to LLMs, which act as annotators to rectify noisy labels. Finally, different optimization objectives are applied to each subset depending on its noise level. It reduces reliance on human annotation, improves label quality through LLM assistance, and follows adaptive learning tailored to noise severity. AnchorAL \cite{lesci2024anchoral} introduces an anchor-based sampling strategy to overcome the scalability and class imbalance challenges in traditional AL. Instead of querying the entire unlabeled pool, it selects a small, dynamic subpool by identifying class-specific labeled examples called anchors followed by retrieving their most similar unlabeled counterparts. It reduces computational overhead while improving the discovery of minority class instances, leading to more balanced and effective training sets. It speeds up execution and improves performance, making it suitable for application on real-world datasets. Thus, AL boosts up performance in low-resource settings. However, its efficacy relies on handling cold-starts, uncertainty estimation, noise mitigation and its integration with ICL.

    \item \textbf{Curriculum Learning (CuL):} It hinges upon the ordering of samples in the dataset to improve utilization. The ordering approach deploys heuristics capturing the complexity of sequences and determines a pace to progressively move from simpler sequences to complex sequences \cite{press2021shortformer}. Curri-DPO \cite{pattnaik2024enhancing} applies CuL to Direct Preference Optimization (DPO) by organizing multiple preference pairs per prompt from "easy" to "hard" based on response quality and preference strength. The curriculum is built using ranked responses, where the model learns incrementally from the most preferred to the least preferred outputs. This structured progression allows the model to first handle clear, unambiguous preferences before tackling more nuanced or borderline cases. This reduces the cognitive load during training and enhances generalization with gains up to 7.5\% compared to standard DPO. Phi-4 \cite{abdin2024phi} implements CuL through strategic sequencing of high-quality synthetic data. Rather than relying solely on traditional distillation from a teacher model, it organizes synthetic examples such that training begins with simpler, foundational tasks and gradually progresses to more complex reasoning challenges. This structured progression enables the model to build robust capabilities incrementally, leading to superior performance in STEM and reasoning benchmarks despite its relatively modest 14B parameter size. However, validation of the synthetic data and designing an optimal curriculum that balances task complexity with learning progression is a challenge. Qwen-3 \cite{yang2025qwen3} extends CuL beyond training into inference, offering a dynamic and adaptive approach to reasoning. The model family, spanning from 0.6B to 235B parameters, incorporates a dual-mode system— thinking mode for complex, multi-step tasks and default mode for simpler, fast-response queries. This design mirrors CuL by allocating cognitive effort based on task difficulty. The introduction of a thinking budget further refines this process, enabling intelligent management of computational resources prioritizing deeper reasoning when required. However, managing mode-switching logic and tuning the thinking budget introduces overhead during inference. Xu et al. \cite{xu2024context} arranged in-context examples in ICL following CuL based on label distributions in the prompt. By leveraging label distribution statistics as an indicator to difficulty, they ensured that the model is exposed to a balanced and pedagogically meaningful sequence of examples in the prompt. Efficient Ordering-Aware Automated Selection of Exemplars (EASE) \cite{wu2024prompt} not only selects the most informative examples for prompting but also arranges them in an order of increasing complexity or relevance for CuL. By modeling both selection and ordering jointly, EASE ensures that the model is exposed to a structured sequence of examples that facilitates better generalization and task adaptation in few-shot learning scenarios. Guo et al. \cite{guo2024makes} investigated the effect of sequencing examples in ICL across multiple tasks and LLMs. They discovered that certain ordering strategies such as grouping by label similarity or arranging examples from simple to complex can significantly enhance prediction accuracy. For this, they proposed DEmO, that adaptively selects performant example orders without requiring additional in-domain data. It filters candidate orders based on label fairness ensuring a balanced representation. Then it chooses the most influential order for each test instance using a content-free metric.

    \begin{figure*}
    \centering

\tikzset{
    basic/.style  = {draw, text width=2cm, align=center, fill=pink!10!blue!80!red!10, font=\sffamily, rectangle},
    root/.style   = {basic, rounded corners=2pt, thin, align=center, fill=green!30},
    onode/.style = {basic, thin, rounded corners=2pt, align=center, fill=green!60,text width=2cm,},
    tnode/.style = {basic, thin, align=left, fill=pink!0, text width=8.1cm, align=left},
    xnode/.style = {basic, thin, rounded corners=2pt, align=center,  fill=pink!60, text width=3.6cm,},
    wnode/.style = {basic, thin, align=left, fill=pink!10, text width=6cm},
    edge from parent/.style={draw=black, edge from parent fork right}

}
\scalebox{0.75}{%
\begin{forest} for tree={
    grow=east,
    growth parent anchor=west,
    parent anchor=east,
    child anchor=west,
    edge path={\noexpand\path[\forestoption{edge},->, >={latex}] 
         (!u.parent anchor) -- +(6pt,0pt) |-  (.child anchor) 
         \forestoption{edge label};}
}
[Model Design, basic,  l sep=6mm,
        [Chunking, xnode, l sep=6mm,
            [Transformer-XL \cite{dai2019transformer}; 
            $\infty$-former \cite{martins2022former}; 
            Mistral 7B  \cite{jiang2023mistral7b}; 
            StreamingLLM \cite{xiaoefficient}; 
            ABSA BERT \cite{ansar2021efficient}; 
            MemWalker \cite{chen2023walking}; 
            RAPTOR \cite{sarthi2024raptor}; 
            Sparse Transformer \cite{child2019generating}; 
            Longformer \cite{beltagy2020longformer}; 
            ETC \cite{ainslie2020etc}; 
            BigBird \cite{zaheer2020big}; 
            Memory Compressed Transformer \cite{liu2018generating}; 
            BlockBERT \cite{qiu2020blockwise}; 
            FlashAttention \cite{dao2022flashattention}; 
            Dynamic Sparse Flash Attention \cite{pagliardini2023fast}, tnode]]
[Mixture-of-Experts, xnode, l sep=6mm,
        [GShard \cite{lepikhingshard}; 
        Switch Transformer \cite{fedus2022switch}; 
        BASE \cite{lewis2021base}; 
        Expert Choice \cite{zhou2022mixture}; 
        GLaM \cite{du2022glam}; 
        FasterMoE \cite{he2022fastermoe}; 
        PanGu-$\Sigma$ \cite{ren2023pangu}; 
        Mixtral 8x7B \cite{jiang2024mixtral}; 
        LLaMA 2 \cite{touvron2023llama2}; 
        FLAN-MoE \cite{shen2024mixture}; 
        DeepSeekMoE \cite{dai2024deepseekmoe}, tnode]]
[Low-Rank Approximation, xnode, l sep=6mm,
        [Activation-aware SVD \cite{yuan2023asvd}; 
        Fisher-weighted SVD \cite{hsulanguage}; 
        SVD-LLM \cite{wangsvd}; 
        Linformer \cite{wang2020linformer}; 
        Performers \cite{choromanski2020masked}; 
        Sumformer \cite{alberti2023sumformer}; 
        LoSparse \cite{li2023losparse}; 
        DSFormer \cite{chand2023dsformer}; 
        DRONE \cite{chen2021drone}; 
        LPLR factorization \cite{saha2023matrix}; 
        ZeroQuant-V2 \cite{yao2023zeroquant}; 
        LoRD \cite{kaushal2024lord}; 
        ALBERT \cite{lan2019albert}; 
        SAFE \cite{reid2021subformer}; 
        TensorGPT \cite{xu2023tensorgpt}, tnode]]
[Clustering, xnode, l sep=6mm,
        [Reformer \cite{kitaev2019reformer}; 
        Routing Transformer \cite{roy2021efficient}; 
        Clustered Attention \cite{vyas2020fast}; 
        ClusterFormer \cite{wang2022clusterformer}; 
        HyperAttention \cite{han2024hyperattention}, tnode]]
[Parameter Sharing, xnode, l sep=6mm,
        [Tied Transformer \cite{xia2019tied}; 
        Recurrent Stacking \cite{dabre2019recurrent}; 
        ALBERT \cite{lan2019albert}; 
        Universal Transformer \cite{dehghaniuniversal}; 
        Pre-trained Checkpoints \cite{rothe2020leveraging}; 
        Perceiver \cite{jaegle2021perceiver}; 
        Subformer \cite{reid2021subformer}; 
        Basis Sharing \cite{wang2025basis}, tnode]]
]
\end{forest}
}

    \caption{Taxonomy of the model design strategies}
    \label{fig:lit_surv5}
\end{figure*}

    \item \textbf{Dimensionality Reduction:} To ameliorate the curse of dimensionality, a few studies have attempted to enhance efficiency without compromising the informativeness of embeddings. TexIm \cite{ansar2023texim} encodes text into a latent space that can be effectively mapped to image features with significantly lower memory footprint. It leverages BERT’s contextual embedding to transform token representations into low-dimensional RGB image pixels applying dimensionality reduction techniques like Principal Component Analysis (PCA). It contributes to the growing field of vision-language integration by showing how Transformer-based LMs can enhance visual understanding. Its subsequent version, TexIm FAST \cite{ansar2025texim} utilized a Transformer-based variational autoencoder for creating uniform-dimensional image representations of text sequences. These representations can then be used as input to vision-based LLMs. It further reduced the memory consumption by 66.67\% compared to the initial TexIm demonstrating exceptional ability in NLU tasks like semantic similarity evaluation and sentiment analysis. Martín-Fernández et al. \cite{martin2024larger} applied pretrained multimodal LLMs to extract rich semantic features from text and image inputs applying PCA. Their experiments show that using larger encoders with smaller, label-efficient regressors yields competitive performance across social perception benchmarks. Ignacio et al. \cite{ignacio2025unet} formulated a transformer encapsulated in the UNet architecture. The encoder block in UNet, applies dimensionality reduction for efficient representation of the input to be fed to the transformer. Although this approach enabled utilization of LLMs on constrained resources, the representational capacity degraded with compression of the input representations. These works contribute to a growing body of research on representation efficiency, suggesting that dimensionality reduction is not just a computational convenience but a strategic necessity in certain LLM applications.
    
\end{itemize}

\subsubsection{Model Design}
\label{md}
To achieve desirable efficiency, the model design should be such that it delivers optimal results with minimum complexity. The contributions towards efficient model design as shown in Figure \ref{fig:lit_surv5} can be grouped under the following categories:

\begin{itemize}

    \item \textbf{Chunking:} It deals with slicing a sequence into several blocks, processing each block individually, and connecting the representations of these blocks through some mechanism. Transformer-XL \cite{dai2019transformer} introduces a segment-level recurrence mechanism that allows hidden states from previous segments to be reused as memory for the current segment. This helps to efficiently compute attention for long sequences by breaking them down into multiple blocks. $\infty$-former \cite{martins2022former} maintains a compressed memory state that is updated recurrently to process streaming data or long contexts without quadratic spike in memory. It deploys kernel-based attention and a state-space formulation to ensure efficient and differentiable memory updates. Mistral 7B  \cite{jiang2023mistral7b} implements Sliding Window Attention (SWA) to restrict each token to attend only to its neighboring tokens within a predefined window. This dramatically reduces computational and memory costs while preserving local contextual information. It is effective in scenarios where long-range dependencies are less critical or can be captured through stacked layers. StreamingLLM \cite{xiaoefficient} tackles two major challenges in streaming applications: the high memory cost of caching Key-Value (KV) states during decoding, and the inability of standard LLMs to generalize beyond their training sequence length. It hinges on the attention sink phenomenon, where initial tokens attract strong attention scores regardless of their semantic importance. By retaining the KV states of these initial tokens and introducing a placeholder sink token during pre-training, it achieves up to 22.2$\times$ speedup over sliding window recomputation baselines.

    Chunking can be applied directly on the input for filtering and segmentation. This reduces the input memory requirements and allows efficient computation. ABSA BERT \cite{ansar2021efficient} breaks down each sequence based on significant phrases contained in it while filtering out irrelevant chunks of tokens before being fed into the BERT model. MemWalker \cite{chen2023walking} learns to navigate selectively through relevant parts of a long document, akin to how humans skim and focus on key sections. It combines a retrieval-based memory module with a navigation controller that dynamically chooses which memory slots to attend to, enabling efficient handling of inputs far beyond typical Transformer context limits. Recursive Abstractive Processing for Tree-Organized Retrieval (RAPTOR) \cite{sarthi2024raptor} constructs a hierarchical tree of recursive summaries by embedding, clustering, and summarizing text chunks from the bottom up. While inferencing, it retrieves from this tree, allowing efficient reasoning over both fine-grained and high-level content.

    \begin{figure*}[t!]
    \centering
    \begin{subfigure}[h]{0.18\textwidth}
        \centering
        \includegraphics[width=\linewidth]{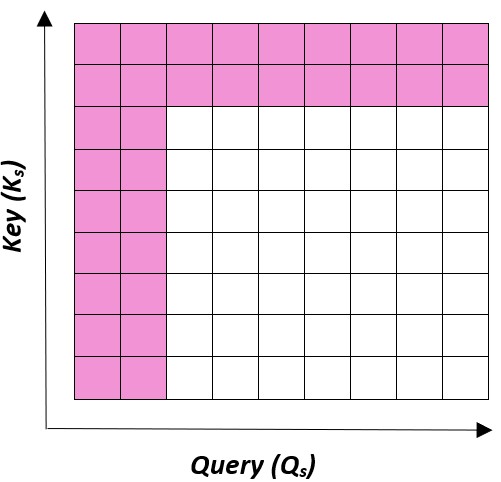}
        \caption{Global}
        \label{fig17a}       
    \end{subfigure}%
    ~ 
    \begin{subfigure}[h]{0.18\textwidth}
        \centering
        \includegraphics[width=\linewidth]{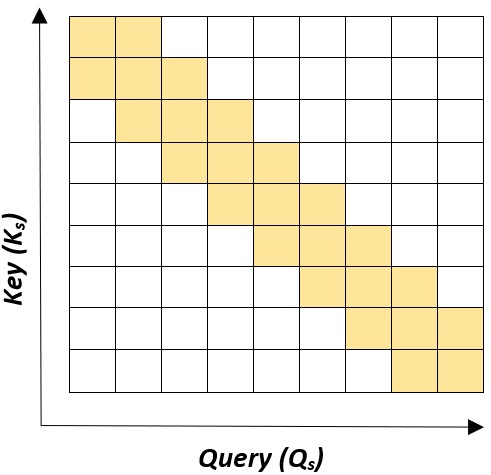}
        \caption{Band}
        \label{fig17b}       
    \end{subfigure}
    ~
        \begin{subfigure}[h]{0.18\textwidth}
        \centering
        \includegraphics[width=\linewidth]{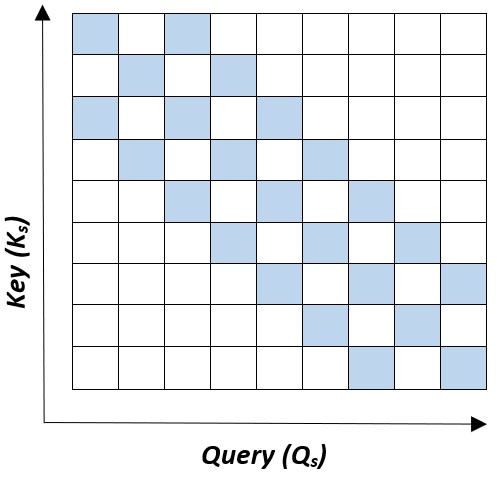}
        \caption{Dilated}
        \label{fig17c}       
    \end{subfigure}%
    ~ 
    \begin{subfigure}[h]{0.18\textwidth}
        \centering
        \includegraphics[width=\linewidth]{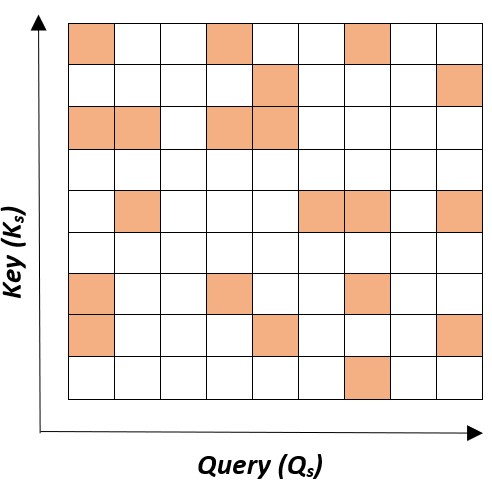}
        \caption{Random}
        \label{fig17d}       
    \end{subfigure}
     ~ 
    \begin{subfigure}[h]{0.18\textwidth}
        \centering
        \includegraphics[width=\linewidth]{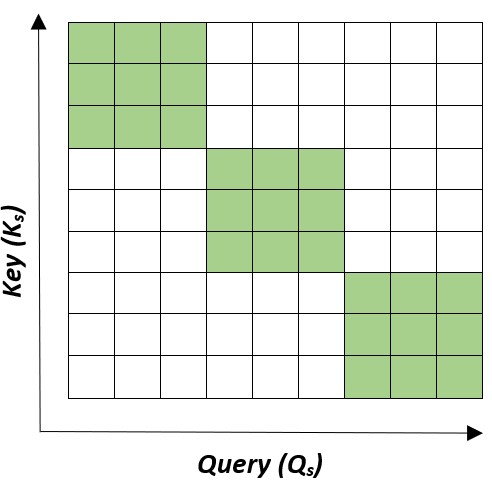}
        \caption{Block}
        \label{fig17e}       
    \end{subfigure}
    
    \caption{Common types of sparse attention patterns}
    \label{fig17}
    
\end{figure*}

    \item \textbf{Sparse Attention:} A few contributions attempt sparsification of the attention matrix to reduce the complexity. This implies limiting the count of keys to be attended by queries based either on certain pre-defined patterns or input-conditioned connections. Some common patterns might be global attention, band attention, dilated attention, random attention, and block attention as illustrated in Figure \ref{fig17}. It exploits the inherent sparsity in the attention matrix in real-life applications. Sparse Transformer \cite{child2019generating} factorizes the attention matrix to attain sparse patterns where connectivity is established between a pre-defined set of tokens. This reduces the complexity of attention to $O(L\sqrt{L})$ for a sequence length $L$. Longformer \cite{beltagy2020longformer} employs attention at fixed intervals in a strided fashion. It adopts a blend of band attention, dilated attention, and global attention to achieve nearly $O(L)$ complexity. Extended Transformer Construction (ETC) \cite{ainslie2020etc} follows a similar approach agglomerating global attention and local band attention with relative positional encoding. Additionally, it employs masking through Contrastive Predictive Coding as a pre-training objective. BigBird \cite{zaheer2020big} builds upon ETC by applying random patterns of sparse-attention. It can handle sequences eight times the length and achieve linear complexity compared to the conventional attention mechanism. Memory Compressed Transformer \cite{liu2018generating} reduces the number of query-key pairs applying strided convolution. BlockBERT \cite{qiu2020blockwise} proposes an efficient version of BERT by incorporating block-wise patterns in the attention matrix for sparsity. FlashAttention \cite{dao2022flashattention} uses a tiling strategy that fits intermediate computations into fast GPU memory (SRAM), drastically reducing memory overhead and avoiding slow reads/writes to DRAM. This design enables training and inference on longer sequences up to 16K tokens with significant speedups achieving linear time complexity. Dynamic Sparse Flash Attention \cite{pagliardini2023fast} combines the speed of FlashAttention with dynamic sparsity to handle long sequences efficiently. It dynamically selects a sparse set of relevant keys for each query using a learned scoring function, allowing the model to focus computation on the most informative parts of the input lowering memory and compute costs. 

    \item \textbf{Mixture-of-Experts (MoE):} The concept of sparsification for efficient computation has been taken forward with the notion of Mixture-of-Experts (MoE). It divides the task into sub-tasks, assigning each to a specialized model or "expert" for that sub-task. It activates only a few experts during inference to reduce computational load activating only a small subset out of a huge set of LLM parameters. Notable MoE-based LLMs include GShard \cite{lepikhingshard} which uses sparsely-gated MoE layers within a Transformer architecture, allowing only a subset of experts to activate per input. It demonstrated that models could scale beyond 600 billion parameters using automatic sharding, with minimal changes to existing codebases. Switch Transformer \cite{fedus2022switch} deploys a switch routing mechanism that assigns each token to a single expert, simplifying computation and enabling models with up to one trillion parameters and 2,048 experts. Balanced ASsignment of Experts (BASE) layer \cite{lewis2021base} approaches token-to-expert allocation as a linear assignment problem, ensuring balanced expert usage and achieving lower validation perplexity than Switch Transformer. Expert Choice \cite{zhou2022mixture} flipped the routing paradigm by allowing experts to select top-k tokens, improving flexibility and outperforming dense models like T5 on GLUE and SuperGLUE benchmarks. Models like GLaM \cite{du2022glam} demonstrate that it helps to attain high accuracy along with efficient use of resources. FasterMoE \cite{he2022fastermoe} tackles the load imbalance in MoE models through fine-grained concurrent scheduling for distributed computing. PanGu-$\Sigma$ \cite{ren2023pangu} transitioned from a dense Transformer to a sparse MoE using random routed experts, expert computation and storage separation. It is trained on over 329 billion tokens and excels in Chinese zero-shot tasks compared to ERNIE 3.0 Titan \cite{wang2021ernie}. Mixtral 8x7B \cite{jiang2024mixtral} with 46.7 billion total parameters, activates only two of its eight 7B experts per token. This results in faster inference and superior performance over LLaMA 2 \cite{touvron2023llama2} having 70B parameters. FLAN-MoE \cite{shen2024mixture} demonstrated that MoE models benefit more from instruction tuning than dense ones, combining both techniques to improve zero-shot and few-shot capabilities without increasing memory or compute demands, outperforming FLAN-T5 in generalization tasks. DeepSeekMoE \cite{dai2024deepseekmoe} includes fine-grained expert segmentation and shared expert isolation to enhance specialization and reduce redundancy. With models ranging from 2B to 145B parameters, it matches or exceeds the performance of GShard 2.9B, and LLaMA2 with over 55\% less resource requirements. LLaMA 4 Maverick\footnote{\url{https://ai.meta.com/blog/llama-4-multimodal-intelligence/}} is multimodal model with a total of 400B parameters out of which 17B are active, leveraging 128 routed experts and a shared expert to optimize inference efficiency. It outperforms GPT-4o and Gemini 2.0 Flash across key benchmarks and matches DeepSeek v3 in reasoning and coding tasks, despite using less than half the active parameters.

    \item \textbf{Low-Rank Approximation (LRA):} A low-rank matrix $A \in  \mathbb{R}^{d_{m} \times d_{n}}$ is a matrix whose rank $r$ (the number of linearly independent rows or columns) is significantly smaller than its dimensions, i.e. $r \ll \min(d_{m}, d_{n})$. In LLMs, weight matrices are often low-rank, meaning they contain more parameters than necessary for effective representation. To reduce this redundancy, LRA has emerged as a powerful technique by reducing a matrix $A \in  \mathbb{R}^{d_{m} \times d_{n}}$ with two smaller matrices $B \in  \mathbb{R}^{d_{m} \times r}$ and $C \in  \mathbb{R}^{r \times d_{n}}$ significantly reducing memory and computational costs. The most widely used LRA method is Singular Value Decomposition (SVD) that decomposes a matrix into $A = U\Sigma V^\top$ where $U$ and $V$ are orthogonal matrices and $\Sigma$ contains the singular values. Truncating smaller singular values allows the matrix to be approximated with significantly fewer parameters, reducing space complexity from $O(d_{m}d_{n})$ to $O(d_{m}r+rd_{m})$. More advanced versions of SVD have emerged. Activation-aware SVD (ASVD) \cite{yuan2023asvd} scales weight matrices based on activation distributions and searches for optimal truncation ranks, improving decomposition accuracy without retraining. Fisher-weighted SVD (FWSVD) \cite{hsulanguage} uses Fisher information to prioritize important weights, though it requires task-specific gradient computation. SVD-LLM \cite{wangsvd} addresses the limitations of both by mapping singular values directly to compression loss through truncation-aware data whitening and layer-wise updates, achieving superior performance and speed. 
    
    Linformer \cite{wang2020linformer} performs LRA of the self-attention matrix to enhance efficiency. Similarly, the application of kernels for the approximation of self-attention has gained popularity as it reduces the effort required to compute self-attention for the entire sequence matrix. A prominent example of this is Performers \cite{choromanski2020masked}. Sumformer \cite{alberti2023sumformer} bridges the gap between Linformer and Performer. It introduces a theoretically grounded Transformer architecture that achieves universal approximation for permutation-equivariant sequence-to-sequence functions. The authors demonstrate that a single attention layer is sufficient for universal approximation, challenging the conventional wisdom that deep stacking is necessary. LoSparse \cite{li2023losparse} combines LRA with pruning, retaining expressive neurons while eliminating redundant ones. This mitigates the risk of over-pruning and maintains model coherence. DSFormer \cite{chand2023dsformer} decomposes weight matrices into a semi-structured sparse matrix and a small dense matrix, balancing sparsity with learnability. DRONE \cite{chen2021drone} adopts a data-aware approach that minimizes output approximation error rather than weight reconstruction error, leading to better task-specific performance. Hybrid methods such as Low Precision and Low Rank (LPLR) factorization \cite{saha2023matrix} and ZeroQuant-V2 \cite{yao2023zeroquant} integrate quantization with LRA to further reduce model size and improve inference speed. LoRD \cite{kaushal2024lord} demonstrates the scalability of LRA by compressing a 16B parameter LLM to 13.2B with minimal performance degradation. 

    The high-dimensional LLM embeddings containing redundant parameters, have also been targeted for LRA. ALBERT \cite{lan2019albert} applies matrix factorization to reduce embedding size. Self-Attentive Factorized Embeddings (SAFE) \cite{reid2021subformer} adds a lightweight self-attention layer atop a linear projection to enhance expressiveness in compressed embeddings. TensorGPT \cite{xu2023tensorgpt} utilizes Tensor-Train Decomposition (TTD) to represent token embeddings as Matrix Product States (MPS), enabling distributed and efficient computation.

    \item \textbf{Clustering:} It refers to grouping related elements, features in a sequence, or even attention heads to achieve efficient computation of attention. It attempts to learn patterns in the data by capturing relevant tokens and clustering them together into buckets. Based on the similarity metric applied for clustering, various models have been devised. Reformer \cite{kitaev2019reformer} utilizes a hashing-based similarity measure while the Routing Transformer \cite{roy2021efficient} deploys a K-means clustering algorithm. Clustered Attention \cite{vyas2020fast} reduces the number of attention computations by grouping queries into clusters and computes attention only for the centroids of these clusters. To refine approximation, it identifies the most relevant keys for each query and computes exact dot products selectively. This results in a model with linear complexity relative to sequence length for a fixed number of clusters. Unlike methods that rely on fixed patterns or separate clustering processes, ClusterFormer \cite{wang2022clusterformer} outperforms Reformer and Routing Transformer by jointly training the clustering task with the target task, benefitting from shared optimization. It computes attention within each cluster of tokens independently, significantly improving computational efficiency. HyperAttention \cite{han2024hyperattention} achieves linear time complexity by combining two key techniques: Locality Sensitive Hashing (LSH) to detect heavy elements in the attention matrix, and Uniform Column Sampling (UCS) to approximate the heavy as well as light components. This helps to bypass the quadratic bottleneck of traditional attention while maintaining high-quality approximations. Its modular design enables seamless integration with fast low-level implementations like FlashAttention and reduces inference time especially for very long sequences.

    \item \textbf{Parameter Sharing:} The complexity of a model is proportional to the number of parameters. Hence, reducing the number of parameters can be beneficial for model efficiency. Tied Transformer \cite{xia2019tied} tied the weights, i.e. applied parameter sharing between the Transformer encoder and decoder blocks, achieving results comparable to the standard Transformer. It demonstrated that strategic weight sharing can maintain model expressiveness while improving computational and memory efficiency. Dabre and Fujita \cite{dabre2019recurrent} proposed recurrent stacking for building compact models with fewer parameters by reusing the same set of weights across multiple layers. ALBERT \cite{lan2019albert} applies matrix decomposition upon the embedding layer along with cross-layer parameter sharing between the Transformer layers, reducing the number of unique parameters and memory usage. Universal Transformer \cite{dehghaniuniversal} shares parameters across all layers incorporating recurrent computation. It enables dynamic depth through a halting mechanism that determines how many steps each input token requires augmenting efficiency through a more compact structure. Rothe, Narayan, and Severyn \cite{rothe2020leveraging} initialized both the encoder and decoder with shared parameters derived from a pre-trained LM, significantly reducing the memory footprint. It demonstrated the utility of parameter sharing and transfer learning in LLMs towards enhancing efficiency while maintaining competitive performance. Rather than explicitly tying weights across layers, Perceiver \cite{jaegle2021perceiver} achieves parameter efficiency by decoupling input size from model size and reusing latent representations. It uses a fixed-size latent array that interacts with the input via cross-attention. It reuses attention and/or feedforward blocks across layers, mimicking parameter sharing. This aids in reducing total unique parameters while maintaining depth and expressiveness. Subformer \cite{reid2021subformer} adopted “sandwich-style” sharing, tying parameters across the middle layers while keeping the initial and final layers independent. This configuration strikes a balance between improving parameter efficiency and preserving performance. Takase and Kiyono \cite{takase2023lessons} conducted a study on parameter sharing strategies across layers in Transformer-based LMs, providing practical insights to how different parameter sharing schemes affect LM performance and efficiency. Rather than uniformly applying a single layer’s weights to all layers, they explore more flexible configurations where a smaller set of layers $\lambda'$ is reused across a larger architecture having $\lambda$ layers such that $\lambda'< \lambda$. Basis Sharing \cite{wang2025basis} uses a basis decomposition strategy that decomposes the matrices into shared basis components and layer-specific coefficients applying SVD. This enables multiple layers to reuse a compact set of basis vectors, significantly reducing the number of parameters while preserving model performance.

\end{itemize}

    \begin{figure*}
    \centering

\tikzset{
    basic/.style  = {draw, text width=2cm, align=center, fill=pink!10!blue!80!red!10, font=\sffamily, rectangle},
    root/.style   = {basic, rounded corners=2pt, thin, align=center, fill=green!30},
    onode/.style = {basic, thin, rounded corners=2pt, align=center, fill=green!90!yellow!10,text width=1.5cm,},
    tnode/.style = {basic, thin, align=left, fill=pink!20, text width=2.1cm, align=center},
    xnode/.style = {basic, thin, rounded corners=2pt, align=center,  fill=pink!60, text width=2.1cm,},
    wnode/.style = {basic, thin, align=left, fill=pink!0, text width=5.4cm},
    edge from parent/.style={draw=black, edge from parent fork right}

}
\scalebox{0.75}{%
\begin{forest} for tree={
    grow=east,
    growth parent anchor=west,
    parent anchor=east,
    child anchor=west,
    edge path={\noexpand\path[\forestoption{edge},->, >={latex}] 
         (!u.parent anchor) -- +(6pt,0pt) |-  (.child anchor) 
         \forestoption{edge label};}
}
[Model Downsizing, basic,  l sep=6mm,
 	[Pruning, xnode, l sep=6mm,
        [Structured Pruning, tnode, l sep=6mm,
            [LLM-Pruner \cite{ma2023llm}; 
 	 	 	Sheared LLaMA \cite{xia2024sheared}; 
 	 	 	LoSparse \cite{li2023losparse}; 
 	 	 	ZipLM \cite{kurtic2023ziplm}, wnode]]
		[Unstructured Pruning, tnode, l sep=6mm,
            [SparseGPT \cite{frantar2023sparsegpt}; 
 	 	 	Wanda \cite{sun2023a}; 
 	 	 	Hessian SparseGPT \cite{shao2024one}; 
 	 	 	DynaTran \cite{tuli2023acceltran}; 
 	 	 	SparseLLM \cite{bai2024gradient}, wnode]] 	 	 		
		[Contextual Pruning, tnode, l sep=6mm,
            [Deja Vu \cite{liu2023deja}; 
 	 	 	Mini-GPTs \cite{valicenti2023mini}, wnode]]]	 	 		
	[Knowledge Distillation (KD), xnode, l sep=6mm,
        [White-box KD, tnode, l sep=6mm,
            [Baby LLaMA \cite{timiryasov2023baby}; 
 	 	 	MiniLLM \cite{gu2024minillm}; 
 	 	 	KPTD \cite{padmanabhan2023propagating}; 
 	 	 	TED \cite{liang2023less}; 
 	 	 	MiniMA \cite{zhang2023lifting}; 
 	 	 	GKD \cite{agarwal2024policy}, wnode]]	 	 		
		[Black-Box KD, tnode, l sep=6mm,
            [Multitask-ICT \cite{huang2022context}; 
 	 	 	MCKD \cite{zhao2024multistage}; 
 	 	 	Distilling Step-by-Step \cite{hsieh2023distilling}; 
 	 	 	SCoTD \cite{li2023symbolic}; 
 	 	 	Fine-tune-CoT \cite{ho2023large}; 
 	 	 	Socratic CoT \cite{shridhar2023distilling}; 
 	 	 	DISCO \cite{chen2023disco}; 
 	 	 	LaMini-LM \cite{wu2024lamini}; 
 	 	 	Lion \cite{jiang2023lion}, wnode]]]			
	[Quantization, xnode, l sep=6mm,
        [Post-Training Quantization (PTQ), tnode, l sep=6mm,
            [Weight-only PTQ, onode, l sep=6mm, 
 	 	 	 	[GPTQ \cite{frantar2022gptq}; 
 	 	 	 	AWQ \cite{lin2024awq}; 
 	 	 	 	OWQ \cite{lee2024owq}; 
 	 	 	 	SpQR \cite{dettmers2024spqr}; 
 	 	 	 	SqueezeLLM \cite{kim2024squeezellm}; 
 	 	 	 	FineQuant \cite{kim2023finequant} ; 
 	 	 	 	QuantEase \cite{behdin2023quantease} ; 
 	 	 	 	AffineQuant \cite{ma2024affinequant}, wnode]]		
 	 	 	[Weight-Activation PTQ, onode, l sep=6mm,
 	 	 	 	[ZeroQuant \cite{yao2022zeroquant}; 
 	 	 	 	ZeroQuant-V2 \cite{yao2023zeroquant}; 
 	 	 	 	ZeroQuant-FP \cite{wu2023zeroquant}; 
 	 	 	 	FlexGen \cite{sheng2023flexgen}; 
 	 	 	 	SmoothQuant \cite{xiao2023smoothquant}; 
 	 	 	 	RPTQ \cite{yuan2023rptq}; 
 	 	 	 	OliVe \cite{guo2023olive}; 
 	 	 	 	OmniQuant \cite{shao2024omniquant}; 
 	 	 	 	QLLM \cite{liu2024qllm}; 
 	 	 	 	Atom \cite{zhao2024atom}, wnode]]]
		[Quantization-Aware Training (QAT), tnode, l sep=6mm,
            [QuantGPT \cite{tao2022compression}; 
 	 	 	LLM-QAT \cite{liu2024llm}; 
 	 	 	BitNet \cite{wang2023bitnet}; 
 	 	 	BEExformer \cite{ansar2024beexformer}, wnode]]]
    ]        
\end{forest}
}
    \caption{Taxonomy of model downsizing strategies}
    \label{fig:lit_surv6}
\end{figure*}

\subsubsection{Model Downsizing}
\label{mc}
After model designing, the model can be further downsized to reduce the computational requirements. The model downsizing techniques as illustrated in Figure \ref{fig:lit_surv6} have been discussed herein-below.

\begin{itemize}
    \item \textbf{Pruning:} It deals with removing the undesired parameters from the neural network \cite{liang2021pruning}. This brings down the computational complexity, memory footprint as well as the bandwidth leading to a rise in model efficiency. It also serves as a regularizer and prevents overfitting. Pruning can be applied during pre-training \cite{louizos2018learning}, fine-tuning \cite{sanh2020movement}, or at the time of inference \cite{fan2019reducing}. Pruning can be further categorized as follows: 

    \begin{enumerate}
        \item \textbf{Structured Pruning:} It prunes a section of the neural network such as entire hidden layers or attention heads \cite{sajjad2023effect}. LLM-Pruner \cite{ma2023llm} introduces a task-agnostic framework that identifies and removes non-essential interconnected structures using gradient and Hessian-based importance estimation, enabling efficient pruning with minimal data and recovery via LoRA. Sheared LLaMA \cite{xia2024sheared} enhances LLM-Pruner with targeted structured pruning, which reshapes models by removing layers, heads, and hidden dimensions; accompanied by dynamic batch loading, which adapts training data composition based on domain-specific loss. It achieves aggressive compression—reducing LLaMA2-7B to 1.3B parameters with strong performance. LoSparse \cite{li2023losparse} proposes a hybrid low-rank and sparse decomposition of weight matrices, balancing expressiveness and diversity during pruning. ZipLM \cite{kurtic2023ziplm} takes a hardware-aware approach, iteratively shrinking weight matrices to meet predefined speedup targets across tasks and environments. Overall, structured pruning has emerged as a powerful strategy for compressing LLMs while preserving performance and deployment efficiency.

        \item \textbf{Unstructured Pruning:} It offers fine-grained control by removing individual weights based on certain thresholds. SparseGPT \cite{frantar2023sparsegpt} frames pruning as a sparse regression problem, using Hessian inversion to prune up to 60\% of parameters without retraining. To simplify the process, Wanda \cite{sun2023a} introduces a lightweight criterion based on the product of weight magnitude and input activation norm, avoiding second-order computations while maintaining competitive performance. Shao et al. \cite{shao2024one} further refine SparseGPT by Hessian sensitivity-aware mixed-sparsity pruning, which flexibly assigns sparsity across layers to minimize pruning-induced error. Beyond weight pruning, DynaTran \cite{tuli2023acceltran} proposes runtime activation pruning for improved inference throughput, implemented via the ASIC-based AccelTran architecture with hardware-aware optimizations like matrix tiling. Expanding the scope, SparseLLM \cite{bai2024gradient} introduces a global pruning framework that transcends layer-wise constraints, using a novel optimization strategy to retain accuracy while reducing resource demands—ideal for large-scale, resource-constrained deployments. Overall, unstructured pruning enables high sparsity with minimal performance loss. However, its irregular sparsity patterns require specialized hardware accelerators. 

        \item \textbf{Contextual Pruning:} It has emerged as a promising alternative to traditional pruning methods, aiming to reduce inference cost without re-training or compromising model integrity. Deja Vu \cite{liu2023deja} dynamically selects small, input-dependent subsets of attention heads and parameters that closely approximate the LM’s output. This enables precise predictions while maintaining learning capabilities. Also, its hardware-optimized asynchronous execution significantly boosts inference speed. Complementing this, Mini-GPTs \cite{valicenti2023mini} introduce domain-specific pruning by identifying and retaining only contextually relevant components of LLMs. They preserve essential functionalities while drastically reducing model size and computational requirements, making them ideal for specialized tasks across diverse domains. Together, these approaches highlight the potential of contextual pruning to deliver efficient, scalable, and task-adaptive LLMs sans the overhead of re-training.
    \end{enumerate}

    \item \textbf{Knowledge Distillation (KD):} It is based on condensing the knowledge from a highly complex model termed as "teacher model" to a less complex model termed as "student model" based on a custom loss function \cite{hinton2015distilling}. The predictions from the teacher model are converted into a probabilistic distribution having soft labels and fed into the student model. The soft labels accelerate the learning process of the student model due to more variance imbibed in it. A temperature T is applied to obtain the soft labels from the logits of the teacher model. The higher the value of T, the softer the probability distribution. Thereafter, the loss between the soft labels and the hard targets is computed and minimized. Although KD enhances efficiency, it hinders the overall model performance and generalizability while hyperparameter tuning of the student model increases computational cost \cite{stanton2021does}. The KD approaches can be classified as White-Box and Black-Box approaches as follows:  

    \begin{enumerate}
        \item \textbf{White-box KD:} It has emerged as a powerful paradigm for compressing and enhancing LLMs by directly leveraging the internal parameters, logits, or intermediate representations of the teacher model. Baby LLaMA \cite{timiryasov2023baby} distills an ensemble of GPT-2 and smaller LLaMA models into a compact 58M-parameter student model that not only retains but exceeds the performance of its teachers, showcasing the strength of ensemble-based distillation. MiniLLM \cite{gu2024minillm} rethinks the conventional KD objective of Kullback-Leibler (KL) divergence by introducing a reverse KL divergence formulation optimized via policy gradients. This proves to be effective for open-ended text generation tasks with complex output spaces. Knowledge Propagation Through Distillation (KPTD) \cite{padmanabhan2023propagating} takes a semantic route, transferring knowledge from entity definitions by prompting the teacher to generate explanatory text and aligning the student’s parameter distribution accordingly, enabling deeper conceptual understanding. Task-aware layEr-wise Distillation (TED) \cite{liang2023less} innovates with layer-specific filters that extract and align internal states between teacher and student, yielding consistent gains in both continual pre-training and downstream fine-tuning. Token-Scaled Logit Distillation (TSLD) \cite{kim2023token} shifts focus to token-level granularity, reforming intermediate representations to improve quantization-aware training (QAT), a crucial step for deploying LLMs in resource-constrained environments. MiniMA \cite{zhang2023lifting} addresses the often-overlooked capacity-performance tradeoff by formalizing it into a distillation principle and introducing a 3B model that redefines the Pareto frontier for compute-efficient accuracy. Generalized KD (GKD) \cite{agarwal2024policy} tackles the challenge of distribution mismatch by sampling student outputs during training, making it a flexible enhancement that can be layered onto other distillation strategies. Together, these works reflect a rich and rapidly evolving landscape of white-box KD, each contributing unique mechanisms to push the boundaries of model compression, generalization, and task-specific adaptation.

        \item \textbf{Black-Box KD:} It is a strategy for training smaller LM using only the output predictions of large teacher models, without access to their internal parameters or architecture. This approach is particularly valuable for distilling generalization and emergent abilities such as ICL, CoT reasoning, and Instruction Following (IF). For ICL, methods like Multitask-ICT \cite{huang2022context} and Multistage Collaborative Knowledge Distillation (MCKD) \cite{zhao2024multistage} transfer few-shot learning capabilities by combining in-context objectives with language modeling, often through multi-stage distillation. CoT reasoning is distilled using techniques such as Distilling Step-by-Step \cite{hsieh2023distilling}, Symbolic Chain-of-Thought Distillation (SCoTD) \cite{li2023symbolic}, Fine-tune-CoT \cite{ho2023large}, and Socratic CoT \cite{shridhar2023distilling}, which extract rationales and intermediate steps from teacher models to guide student learning. These methods often outperform traditional fine-tuning, even with less data, by enriching training with diverse reasoning paths. IF ability is addressed through models like DIStilled COunterfactual Data (DISCO) \cite{chen2023disco}, LaMini-LM \cite{wu2024lamini}, and Lion \cite{jiang2023lion}, which generate and filter complex instructions or counterfactual data to improve student adherence to task directives. Lion’s adversarial cycle and GPT-4 \cite{achiam2023gpt} generated datasets further enhance instruction-following performance. From these efforts, black-box KD demonstrates that even without internal access, teacher models like GPT-3 and GPT-4 can effectively guide smaller models toward high-level reasoning and task execution, often rivaling or surpassing much larger counterparts.
    \end{enumerate}

    \item \textbf{Quantization:} It is the process of representing model parameters in memory with reduced precision. This brings down the memory requirements of the model enabling efficient computation. Instances include converting values in 32-bit floating-point notation 'FP32' to reduced precision floating-point (FP) or integer (Int) notations like FP16, FP8, Int8, Int4, and binary (1-bit) \cite{rokh2023comprehensive, dettmers2024spqr}. The efficiency gains through quantization has made it quintessential for LM deployment on edge devices \cite{rahman2023quantized}. The two primary quantization strategies are as follows: 
    \begin{enumerate}
        \item \textbf{Post-Training Quantization (PTQ):} It denotes the technique in which a pre-trained LM is quantized after the training is complete \cite{shen2024efficient, shomron2021post, yao2022zeroquant}. Its primary advantage is its simplicity and speed, as it doesn't require any retraining or fine-tuning of the model. The sub-types of PTQ are as follows:

    \begin{enumerate}
        \item \textbf{Static PTQ:} Both the weights and activations of the LM are quantized to a lower bit-width. This is performed offline, and the quantized values are used during inference. The LM is run on a small, representative calibration dataset to collect a distribution of the activation values at each layer. This information is then used to determine the optimal quantization range for each weight and activation, ensuring minimal quantization errors. Once the quantization ranges are set, the entire model is quantized and frozen for inference \cite{shen2024efficient}.

        \item \textbf{Dynamic PTQ:} It quantizes only the weights of the LM offline. Whereas, the activations are quantized on-the-fly during inference. \cite{shomron2021post} While this requires a slightly more runtime, it proves to be beneficial with additional flexibility for models with highly dynamic activation ranges, such as those with variable sequence lengths.
        
    \end{enumerate}

    Besides, PTQ can be classified on the basis of whether both weights and activations are quantized as follows:

    \begin{enumerate}
        \item \textbf{Weight-only PTQ:} In this, only the weights of LM are quantized while the activations are used as is. Weight-only quantization for LLMs has seen a rapid evolution. GPTQ \cite{frantar2022gptq} marks an early milestone by refining the traditional OBQ \cite{frantar2022optimal} algorithm adopting a fixed left-to-right quantization order per row, reducing the need for frequent Hessian updates and lowering computational cost. Building on this, AWQ \cite{lin2024awq} introduced activation-aware quantization by keeping salient weights tied to high activation magnitudes in higher precision. OWQ \cite{lee2024owq} tackled activation outliers using mixed-precision strategies, while SpQR \cite{dettmers2024spqr} and SqueezeLLM \cite{kim2024squeezellm} focused on isolating and preserving weight outliers through selective high-precision storage and non-uniform quantization. FineQuant \cite{kim2023finequant} proposed a heuristic-based column-wise granularity scheme, and QuantEase \cite{behdin2023quantease} refined GPTQ’s compensation process using Coordinate Descent. AffineQuant \cite{ma2024affinequant} expanded the optimization space by introducing affine transformations, further minimizing quantization error. Collectively,  these methods reflect a diverse set of innovations aimed at balancing computational efficiency, and performance.

        \item \textbf{Weight-Activation PTQ:} It involves compressing LLMs by quantizing both weights and activations. ZeroQuant \cite{yao2022zeroquant} introduced fine-grained quantization with kernel fusion and layer-wise KD. It was later extended in ZeroQuant-V2 \cite{yao2023zeroquant} with Low-Rank Compensation to reduce quantization error while ZeroQuant-FP \cite{wu2023zeroquant} explored floating-point formats (FP4/FP8), showing superior activation quantization over integer types. FlexGen \cite{sheng2023flexgen} focused on memory efficiency by quantizing weights and KV cache into Int4 for large-batch inference. SmoothQuant \cite{xiao2023smoothquant} tackled activation outliers via reparameterization, scaling weight channels while compressing activation ranges. Reorder-based PTQ (RPTQ) \cite{yuan2023rptq} addressed activation channel variability by clustering similar distributions for localized quantization. OliVe \cite{guo2023olive} sacrificed nearby normal values to better represent outliers. OmniQuant \cite{shao2024omniquant} moved beyond empirical heuristics by optimizing clipping boundaries and scaling factors. QLLM \cite{liu2024qllm} introduced channel reassembly and learnable low-rank parameters, outperforming SmoothQuant on LLaMA models. Atom \cite{zhao2024atom} combined mixed-precision and dynamic quantization for activations and KV cache, enhancing throughput without compromising accuracy. Together, these methods showcase an array of innovations targeting activation sensitivity, memory efficiency, and precision control in LLM quantization.
    \end{enumerate}

        \item \textbf{Quantization-Aware Training (QAT):} It is a more sophisticated approach with the quantization process being incorporated directly into LM training. This enables the LM to be more robust to loss of information during quantization to very low precision, often leading to significantly better performance than PTQ. During the forward pass, the LM weights and activations are simulated to be quantized via a rounding function. However, the actual, full-precision floating-point values are maintained. This is due to the fact that the rounding function is not differentiable. This would lead to zero gradient preventing the LM from learning. The Straight-Through Estimator (STE) is a trick that bypasses this problem. During backpropagation, it passes the gradients straight through to the full-precision \textit{latent weights}. This effectively allows the full-precision weights to be updated based on the loss, as if the quantization had not occurred in the forward pass. After training is completed, the latent weights are quantized to produce the final, low-precision weights of the quantized LM \cite{hubara2016binarized, chen2025efficientqat, liu2020bi}.

        Among QAT endeavors, QuantGPT \cite{tao2022compression} combines contrastive and logit distillation during auto-regressive pre-training, achieving compression rates about 14.4$\times$ for GPT-2 and 13.4$\times$ for BART, while maintaining performance parity with full-precision models. LLM-QAT \cite{liu2024llm} uses synthetic data generated by LLMs themselves for KD, allowing quantization independent of original training data. It extends QAT to weights, activations, and the key-value cache, improving throughput and long-context handling, especially in low-bit regimes.  
        
        Binarized Transformer models have pushed the boundaries of QAT. BinaryBERT applies weight binarization with ternary-weight splitting and KD to mitigate performance loss. However, it struggles with activation binarization due to precision degradation. BiBERT advances toward full binarization by maximizing information entropy and using direction-matching distillation based on similarity matrices to address directional mismatches. Binarized Transformer (BiT) adopts a multi-step distillation pipeline, gradually reducing precision and introducing elastic binary activations with trainable parameters for smoother optimization. BitNet \cite{wang2023bitnet} introduces 1-bit quantization for weights and activations in a Transformer architecture, while preserving high-precision optimizer states and gradients. Besides significant gains in memory and energy efficiency, it demonstrates competitive performance compared to FP16 baselines. BEExformer \cite{ansar2024beexformer} takes a more architectural approach, binarizing both weights and activations in a gating-based Transformer using a second-order differentiable approximation of the sign function. This enables effective gradient updates and significant reduction in model size while maintaining inference efficiency. Collectively, these methods showcase integrated training strategies that optimize both model size, runtime efficiency and performance.
    \end{enumerate}

\end{itemize}

    \begin{figure*}
    \centering

\tikzset{
    basic/.style  = {draw, text width=2cm, align=center, fill=pink!10!blue!80!red!10, font=\sffamily, rectangle},
    root/.style   = {basic, rounded corners=2pt, thin, align=center, fill=green!30},
    onode/.style = {basic, thin, rounded corners=2pt, align=center, fill=green!90!yellow!10,text width=1.5cm,},
    tnode/.style = {basic, thin, align=left, fill=pink!20, text width=3cm, align=center},
    xnode/.style = {basic, thin, rounded corners=2pt, align=center,  fill=pink!60, text width=2.1cm,},
    wnode/.style = {basic, thin, align=left, fill=pink!0, text width=6cm},
    edge from parent/.style={draw=black, edge from parent fork right}

}
\scalebox{0.75}{%
\begin{forest} for tree={
    grow=east,
    growth parent anchor=west,
    parent anchor=east,
    child anchor=west,
    edge path={\noexpand\path[\forestoption{edge},->, >={latex}] 
         (!u.parent anchor) -- +(6pt,0pt) |-  (.child anchor) 
         \forestoption{edge label};}
}
[Dynamic Inferencing, basic,  l sep=6mm,
 	[Early Exit (EE),  xnode, l sep=6mm,
        [DeeBERT \cite{xin2020deebert} ; 
 	 	 RightTool \cite{schwartz2020right}; 
 	 	 PABEE \cite{zhou2020bert}; 
 	 	 Global Past-Future EE \cite{liao2021global}; 
 	 	 HASHEE \cite{sun2022simple} ; 
 	 	 PCEE-BERT \cite{zhang2022pcee}; 
 	 	 SkipBERT \cite{wang2022skipbert}; 
 	 	 CALM \cite{schuster2022confident}; 
 	 	 ElasticBERT \cite{liu2022towards} ; 
 	 	 BE3R \cite{mangrulkar2022be3r}; 
 	 	 MuE \cite{tang2023you}; 
 	 	 SkipDecode \cite{del2023skipdecode}; 
 	 	 Short-Cutting Transformer \cite{din2024jump}; 
 	 	 BEExformer \cite{ansar2024beexformer}, wnode]]
	[Token Pruning (TP),  xnode, l sep=6mm,
        [Value-Based Scoring, tnode, l sep=6mm,
            [SpAtten \cite{wang2021spatten}
			LTP \cite{kim2022learned}
			ToP \cite{li2023constraint}, wnode]]
 	 	 [Predictive Modeling, tnode, l sep=6mm,
            [TR-BERT \cite{ye2021tr}
			Transkimmer \cite{guan2022transkimmer} 
			PuMer \cite{cao2023pumer}
			Infor-Coef \cite{tan2023infor}
			SMART-TRIM \cite{wang2024smarttrim}
			LLMLingua-2 \cite{pan2024llmlingua}
			CCM \cite{kim2024compressed}
			GRIFFIN \cite{dongprompt}
			LazyLLM \cite{fulazyllm}, wnode]]]
    [Token Parallelism (TPA), xnode, l sep=6mm,
        [Speculative Computation \cite{burton2012speculative}
        Speculative Decoding \cite{leviathan2023fast}
        Speculative Sampling \cite{chen2023accelerating}
        Staged Speculative Decoding \cite{spectoraccelerating}, wnode]]
] 
\end{forest}
}
    \caption{Taxonomy of dynamic inferencing strategies}
    \label{fig:lit_surv7}
\end{figure*}

\subsubsection{Dynamic Inferencing}
In addition to static methods that reduce the number of parameters in LLMs for faster inference, dynamic inferencing techniques offer input-specific efficiency improvements without altering model size. These flexible and context-aware acceleration methods fall into three main categories as depicted in Figure \ref{fig:lit_surv7} and described as follows:

\begin{itemize}
    \item \textbf{Early Exit (EE):} It involves strategies to speed up inference by halting computation at intermediate layers once sufficient confidence is reached, effectively making the model shallower for certain inputs that require less processing. Various models have explored different criteria for EE. Dynamic Early Exiting BERT (DeeBERT) \cite{xin2020deebert} uses entropy thresholds, RightTool \cite{schwartz2020right} relies on softmax scores, and Patience-based Early Exit (PABEE) \cite{zhou2020bert} exits when predictions stabilize across layers. Global Past-Future EE \cite{liao2021global} enriches these modules with linguistic cues from surrounding layers. Hash-based Early Exiting (HASHEE) \cite{sun2022simple} uses a hashing-based approach relying on non-parametric decisions based on sample similarity. Patient and Confident Early Exiting BERT (PCEE-BERT) \cite{zhang2022pcee} combines confidence scores with a patience counter, while SkipBERT \cite{wang2022skipbert} skips shallow layers when pre-computed chunks are detected. CALM \cite{schuster2022confident} exits when the confidence gap between the top predictions exceeds a threshold. Instead of recomputing or storing every intermediate state, it copies forward the hidden states from earlier layers, trimming the Key Value (KV) cache to keep memory usage in check. ElasticBERT \cite{liu2022towards} integrates dynamic exits with Masked Language Modeling (MLM) and Sentence-Order Prediction (SOP) losses. BE3R \cite{mangrulkar2022be3r} routes inputs to expert models based on sentence complexity. MuE \cite{tang2023you} adapts EE to multimodal LLMs by evaluating layer-wise input similarity, addressing challenges in encoder-decoder dependencies. SkipDecode \cite{del2023skipdecode} introduces batch-level exit points and monotonic exit strategies to optimize caching and batch inference. Short-Cutting Transformer \cite{din2024jump} bypasses intermediate layers using linear transformations. BEExformer \cite{ansar2024beexformer} relies on fractional changes in entropy of the logits between the Transformer units for EE. This eliminates the need for absolute thresholds. Combined with a soft-routing loss calculation technique, it enhances the inferential ability mitigating the problem of "overthinking" in deep networks. Collectively, these methods demonstrate that EE can significantly accelerate LLM inference by customizing computation depth to input complexity, while maintaining performance through carefully designed exit criteria and architectural adaptations.

    \item \textbf{Token Pruning (TP):} It selectively skips processing of less important tokens in higher layers, shortening the effective input and reducing computational load. It considers token relevance for allocating computational resources. Techniques under TP fall into two categories as follows:
    
    \begin{enumerate}
        \item \textbf{Value-Based Scoring:} It relies on value-based scoring mechanisms such as attention weights to identify and eliminate less important tokens. SpAtten \cite{wang2021spatten} exemplifies this approach by ranking tokens according to their importance scores and retaining only the top-k most relevant ones. Learned Token Pruning (LTP) \cite{kim2022learned} enhances a variant of BERT by applying a learnable threshold at each layer, allowing for adaptive pruning based on token significance. Token Pruning ToP \cite{li2023constraint} addresses the challenge of inaccurate token ranking in self-attention by distilling reliable importance scores from the final layer of an unpruned model into earlier layers of a pruned version, improving token selection accuracy throughout the network.

        \item \textbf{Predictive Modeling:} It introduces prediction modules before Transformer layers to estimate token importance more accurately. Token Reduction BERT (TR-BERT) \cite{ye2021tr} uses Reinforcement Learning (RL) to reduce the context by removing irrelevant tokens. Transkimmer \cite{guan2022transkimmer} adds MLP layers but may increase latency. Pruning and Merging (PuMer) \cite{cao2023pumer} combines text-informed pruning with modality-aware merging for vision LMs. Infor-Coef \cite{tan2023infor} applies dynamic downsampling guided by information bottleneck loss. SMART-TRIM \cite{wang2024smarttrim} integrates task-specific pruning without extra training. LLMLingua-2 \cite{pan2024llmlingua} compresses prompts via token classification, achieving significant latency reductions. Compressed Context Memory (CCM) \cite{kim2024compressed} leverages LoRA for context compression but faces memory challenges over time. GRIFFIN \cite{dongprompt} offers training-free pruning by identifying neuron activation patterns, maintaining performance with reduced parameters. LazyLLM \cite{fulazyllm} improves long-context inference by selectively computing key-value pairs, accelerating generation without compromising accuracy.
    \end{enumerate}

    \item \textbf{Token Parallelism (TPA):} It differs from the traditional autoregressive token-by-token generation by enabling simultaneous prediction of multiple tokens using specialized algorithms, significantly boosting throughput. Burton, Warren \cite{burton2012speculative} introduced speculative computation techniques that enable parallel token generation without sacrificing output fidelity. Speculative Decoding (SD) \cite{leviathan2023fast} uses smaller approximation models to generate preliminary token sequences that are then validated and expanded by the full-sized model, preserving the original distribution. Similarly, Speculative Sampling (SpS) \cite{chen2023accelerating} leveraged a modified rejection sampling strategy to produce multiple tokens per call, achieving 2$\times$ to 2.5$\times$ speedups without altering the model architecture. Building on these, Staged Speculative Decoding (SSD) \cite{spectoraccelerating} organizes speculative batches in a tree structure and adds an extra decoding stage to optimize performance in low-resource, on-device settings. It demonstrated a 3.16$\times$ reduction in decoding latency for a 762M parameter GPT-2-L model, maintaining output quality while significantly enhancing efficiency.
\end{itemize}

\begin{figure*}
    \centering

\tikzset{
    basic/.style  = {draw, text width=2cm, align=center, fill=blue!20, font=\sffamily, rectangle},
    root/.style   = {basic, rounded corners=2pt, thin, align=center, fill=green!30},
    onode/.style = {basic, thin, rounded corners=2pt, align=center, fill=green!90!yellow!10,text width=2cm,},
    tnode/.style = {basic, thin, align=left, fill=pink!60, text width=6cm, align=center},
    xnode/.style = {basic, thin, rounded corners=2pt, align=center,  fill=pink!10!blue!80!red!10, text width=3.6cm,},
    wnode/.style = {basic, thin, align=left, fill=pink!20, text width=3cm},
    edge from parent/.style={draw=black, edge from parent fork right}

}
\scalebox{0.75}{%
\begin{forest} for tree={
    grow=east,
    growth parent anchor=west,
    parent anchor=east,
    child anchor=west,
    edge path={\noexpand\path[\forestoption{edge},->, >={latex}] 
         (!u.parent anchor) -- +(6pt,0pt) |-  (.child anchor) 
         \forestoption{edge label};}
}
[LLM Adaptation, basic,  l sep=6mm,
	[Pre-Training, xnode,  l sep=6mm,
		[Causal Language Modeling (CLM), tnode]
		[Masked Language Modeling (MLM), tnode]
		[Replaced Token Detection (RTD), tnode]
		[Permutation Language Modeling (PLM), tnode]
		[Next Sentence Prediction (NSP), tnode]
		[Knowledge Enhanced Modeling (KEM), tnode]
		[Retrieval-Augmented Modeling (RAM), tnode]
		[Prompt Pre-Training (PPT), tnode]]
	[Fine-Tuning, xnode,  l sep=6mm,
		[Gradual Unfreezing, tnode]
		[Parameter Subset Tuning (PST), tnode]
		[Adapter Tuning, tnode]
		[Prefix Tuning, tnode]
		[Instruction Tuning, tnode]
		[Low-Rank Adaptation (LoRA), tnode]]
	[Prompt-Engineering, xnode,  l sep=6mm,
		[Prompt Pruning, tnode]
		[Prompt Summary, tnode]
		[Soft Prompt Compression, tnode]
		[Intermediate Reasoning Optimization, tnode]]	
	[Retrieval-Augmented Generation (RAG), xnode, l sep=6mm,
		[Querying and Retrieval, tnode]
		[Post-Retrieval Restructuring, tnode]]]
\end{forest}
}
    \caption{Taxonomy of efficiency considerations for LLM adaptation}
    \label{fig:lit_surv3}
\end{figure*}

\subsection{RQ4: LLM Adaptation Strategies}
\label{mas}

In this section, the various strategies to adapt an LLM to a given task via pre-training, fine-tuning, prompt-engineering, and RAG have been discussed along with the efficiency considerations. Figure \ref{fig:lit_surv3} presents a taxonomy of the approaches.

\begin{figure*}
    \centering

\tikzset{
    basic/.style  = {draw, text width=2cm, align=center, fill=pink!10!blue!80!red!10, font=\sffamily, rectangle},
    root/.style   = {basic, rounded corners=2pt, thin, align=center, fill=green!30},
    onode/.style = {basic, thin, rounded corners=2pt, align=center, fill=green!90!yellow!10,text width=1.5cm,},
    tnode/.style = {basic, thin, align=left, fill=pink!20, text width=2.1cm, align=center},
    xnode/.style = {basic, thin, rounded corners=2pt, align=center,  fill=pink!60, text width=6cm,},
    wnode/.style = {basic, thin, align=left, fill=pink!0, text width=6cm},
    edge from parent/.style={draw=black, edge from parent fork right}

}
\scalebox{0.75}{%
\begin{forest} for tree={
    grow=east,
    growth parent anchor=west,
    parent anchor=east,
    child anchor=west,
    edge path={\noexpand\path[\forestoption{edge},->, >={latex}] 
         (!u.parent anchor) -- +(6pt,0pt) |-  (.child anchor) 
         \forestoption{edge label};}
}
[Pre-Training, basic,  l sep=6mm,
 	[Causal Language Modeling (CLM),  xnode, l sep=6mm,
        [GPT \cite{radford2018improving}, wnode]]
	[Masked Language Modeling (MLM),  xnode, l sep=6mm,
        [BERT \cite{kenton2019bert}
		T5 \cite{raffel2020exploring}
		PEGASUS \cite{zhang2020pegasus}
		BART \cite{lewis2020bart}, wnode]]
	[Replaced Token Detection (RTD),  xnode, l sep=6mm,
        [ELECTRA \cite{clark2020electra}, wnode]]
	[Permutation Language Modeling (PLM),  xnode, l sep=6mm,
        [XLNet \cite{yang2019xlnet}, wnode]]
	[Next Sentence Prediction (NSP),  xnode, l sep=6mm,
        [BERT \cite{kenton2019bert}
		ALBERT \cite{lan2019albert}, wnode]]
	[Knowledge Enhanced Modeling (KEM),  xnode, l sep=6mm,
        [WKLM \cite{xiong2020pretrained}
		KEPLER \cite{wang2021kepler} 
		CoLAKE \cite{sun2020colake}
		ERNIE 3.0 \cite{sun2021ernie}, wnode]]
	[Retrieval-Augmented Modeling (RAM),  xnode, l sep=6mm,
        [REALM \cite{guu2020retrieval}, wnode]]
	[Prompt Pre-Training (PPT),  xnode, l sep=6mm,
        [Galactica \cite{taylor2022galactica}, wnode]]]       
\end{forest}
}
    \caption{Taxonomy of the pre-training strategies for LLMs}
    \label{fig:lit_surv8}
\end{figure*}

\subsubsection{Pre-Training}
\label{pt}
Creating an LLM does not only revolve around devising a complex architecture with millions of parameters. Rather, LMs need to be trained on datasets proportionate to the model size to deliver optimum performance \cite{hoffmann2022training}, i.e. large models need large datasets. But, high-quality annotated datasets are scarcely available for training a model in a supervised fashion. This is due to annotation being a timely and expensive task with annotators requiring expertise in the given language as well as domain knowledge. However, there exists plenty of unannotated textual content that can be utilized to make LLMs learn vital representations through unsupervised or self-supervised learning. Previously training or \textit{pre-training} is a model initialization strategy through discerning linguistic intricacies in a self-supervised fashion. It enhances the performance at downstream tasks with faster convergence even on limited data. The concept of pre-training gained traction with the surge in deep convolutional models following the ImageNet\footnote{\url{https://image-net.org/challenges/LSVRC/}} challenge in the early 2010s. In context to NLP, Collobert et al. \cite{collobert2011natural} first generated pre-trained word embeddings from large unannotated corpora. Subsequently, other pre-trained word embeddings like Global Vectors for Word Representation (GloVe) \cite{pennington2014glove} and Word2Vec \cite{mikolov2013distributed} were devised. Concerning pre-trained LM, Dai and Le \cite{dai2015semi} became the torchbearer followed by other models like ELMo \cite{peters-etal-2018-deep}, ULMFit \cite{howard2018universal}, GPT \cite{radford2018improving} and BERT \cite{kenton2019bert}. Since then, a plethora of LLMs have been developed with an upward trend in associated research. The strategies for pre-training LLMs as illustrated in Figure \ref{fig:lit_surv8} have been discussed as follows:

\begin{itemize}
    \item \textbf{Causal Language Modeling (CLM): } It predicts the next token in a sequence maximizing the likelihood of the conditional probability distribution over all the unique tokens based on the context. CLM works in a unidirectional manner, i.e. left-to-right manner. This implies that the context only includes the tokens to its left. CLM is more suited for NLG applications. A prominent example of an LLM using CLM is GPT \cite{radford2018improving}. For a given sequence $X=(x_1, ..., x_2, x_n)$, the loss function of CLM is computed as follows:

\begin{equation}
\label{eq37}
\mathcal{L}_{CLM} = -\sum_{t=1}^{T}{log p(x_{t}|X_{<t})}
\end{equation}

\item \textbf{Masked Language Modeling (MLM):} To ameliorate the limitation of CLM to attend only to tokens leftwards, MLM was devised where the context was constructed in a bidirectional fashion, i.e. allowing it to infer from tokens present in both right as well as left direction. This makes MLM the apt choice for NLU applications. MLM usually works by masking out some random percentage of tokens in the sequence and then predicting those tokens based on the context. One of the famous LLMs utilizing MLM is BERT \cite{kenton2019bert}. For a given sequence $X=(x_1, ..., x_2, x_n)$, the loss function for MLM is computed as follows:

\begin{equation}
\label{eq38}
\mathcal{L}_{MLM} = -\sum_{x' \in m(X)}{log p(x'|X_{\texttt{\textbackslash}m(X)})}
\end{equation}

where $m(X)$, $X_{\texttt{\textbackslash} m(X)}$ denote the masked tokens, and the remaining tokens in the sequence $X$ respectively. 

Vanilla MLM deals with replacing single tokens which can reduce their effectiveness at sequence-to-sequence NLG tasks. Instead of predicting a single token, predicting a span of tokens corresponding to a mask can reduce computations involved as can be seen in \textbf{Text-to-Text Transfer Transformer (T5)} \cite{raffel2020exploring}. In T5, the pre-training objective drops 15\% tokens through random sampling from the input. After that consecutive spans of dropped tokens are identified and replaced with special tokens called "sentinel" having a unique token ID representing each span. The target sequence consists of all the masked spans of tokens, separated by the same sentinel tokens used in the input, with an additional end of sequence sentinel token.  

Even entire sentences can be masked as in Pre-training with Extracted Gap-sentences for Abstractive Summarization (PEGASUS) \cite{zhang2020pegasus}. Here, masks representing the dropped sentences are concatenated to form a pseudo-summary which forms the target output for the decoder while the unmasked sentences are fed as input to the decoder. Bidirectional and Auto-Regressive Transformers (BART) \cite{lewis2020bart} applies noise to corrupt the input text and then performs denoising by reconstructing the span of text. A set of corruption strategies are adopted like masking individual tokens, randomly deleting tokens without substituting them with masks so that the model decides the location as well for the missing input, replacing spans of tokens with a single mask contrary to each span represented by a unique ID, shuffling entire sentences and rotating entire sequence in such a manner that it starts with a certain randomly chosen token. This allows pre-training on shorter sequences with equivalent efficacy contributing towards enhanced efficiency. A limitation of MLM is that the masked tokens are restricted to pre-training and are not available at the fine-tuning phase leading to a discrepancy.

\item \textbf{Replaced Token Detection (RTD):} It is a more efficient pre-training objective compared to MLM, requiring fewer samples to deliver comparable or even better results. RTD uses a light-weight generator network to provide replacements for certain tokens in the sequence. The training objective is to determine if each token is original or has been replaced. ELECTRA \cite{clark2020electra} applies RTD as pre-training objective to achieve performance comparable to RoBERTa and XLNet requiring only a quarter of their computational requirements. For a given sequence $X=(x_1, ..., x_2, x_n)$ with $x^{\text{corrupt}}$ denoting the replaced token, the loss function for RTD is computed as follows:

\begin{equation}
\label{eq39}
\begin{split}
\mathcal{L}_{\text{RTD}}(\mathbf{x}, \theta_D) = \mathbb{E}((\sum_{i=1}^{n} -(x_i^{\text{replaced}} = x_i)\log D(x_i^{\text{replaced}}, i) \\ - (x_i^{\text{replaced}} \neq x_i)\log(1 - D(x_i^{\text{replaced}}, i)))
\end{split}
\end{equation}

\item \textbf{Permutation Language Modeling (PLM):} To mitigate the drawback of MLM related to the unavailability of the mask token during fine-tuning, PLM was proposed \cite{yang2019xlnet}. PLM generates a random permutation of the input sequence wherein a permutation defines the order of token predictions\footnote{not to be confused with the order of tokens in the sequence}. Then, the model tries to predict some of the tokens selected as the target considering its position and the remaining tokens. To achieve faster convergence, the endmost tokens are often predicted. A popular LLM utilizing PLM is XLNet \cite{yang2019xlnet}. Given an input sequence $X$ with $X^p$ being its random permutation sequence, the equation for the loss function of PLM is as follows:

\begin{equation}
\label{eq51}
\mathcal{L}_{PLM} = -\sum_{t=1}^{T}{log p(x^{p}_{t}|X^{p}_{<t})}
\end{equation}

\item \textbf{Next Sentence Prediction (NSP):} It has been devised to capture inter-sentence relationships useful in tasks like Natural Language Inference (NLI) \cite{yu2024natural}. It takes two input sentences and aims to determine whether the second sentence is a continuation of the first sequence. It has been used as a pre-training objective in LLMs like BERT.

\item \textbf{Contrastive Learning (CL):} It aims to capture linguistic contextual information by distinguishing between valid and invalid samples employing similarity evaluation. Next-Sentence Prediction (NSP) utilized in BERT \cite{kenton2019bert} is an example of CL. Here, the objective is to identify whether a pair of sentences are next to each other given a set of contiguous and non-contiguous sentences. Although NSP focuses on the topic as well as coherence prediction, a few works have questioned its effectiveness and even demonstrated performance drop due to NSP \cite{liu2019roberta}. To resolve this issue, Sentence-Order Prediction (SOP) was proposed to predict the order of sentences instead of predicting whether a given sentence is next to another sentence. For instance, ALBERT \cite{lan2019albert} achieves superior performance by modeling the inter-sentence coherence through SOP. The loss functions for both SOP and NSP aim to determine the constructiveness of two sentences $X$ and $Y$ as follows:

\begin{equation}
\label{eq52}
\mathcal{L}_{NSP/SOP} = -log p(k|X,Y) \textit{, } \forall p \in {0,1}
\end{equation}

\item \textbf{Knowledge Enhanced Modeling (KEM):} It aims to overcome the lack of real-world knowledge limitation of LLMs by integrating external knowledge sources, during the pre-training phase. Weakly-supervised Knowledge-pretrained Language Model (WKLM) \cite{xiong2020pretrained} substituted entity references in the original texts with names of other entities belonging to the same category by looking up their type from Wikidata\footnote{\url{https://www.wikidata.org/}}. Then, the model is trained to identify which mention is accurate versus those selected at random. Knowledge Embedding and Pre-trained Language Representation (KEPLER) \cite{wang2021kepler} enhanced model training by combining knowledge embeddings with MLM, aiming to unify world knowledge and language understanding within a shared semantic space. Contextualized Language and Knowledge Embedding (CoLAKE) \cite{sun2020colake} merged linguistic and factual information into a word-knowledge graph, learning joint contextual representations through an extended MLM approach. Another way to incorporate world knowledge is by adding extra annotations to large-scale datasets. ERNIE 3.0 \cite{sun2021ernie} uses Universal Knowledge-Text Prediction (UKTP). It requires both unstructured text and knowledge graphs. The model is given a pair consisting of a knowledge graph triple and a corresponding sentence. It is then trained to predict either a masked relation in the triple or a masked word in the sentence. This process enables the model to bridge logical relationships from the knowledge graph with the linguistic information in the text, thereby augmenting knowledge retention with reasoning abilities.

\item \textbf{Retrieval-Augmented Modeling (RAM):} It emphasizes obtaining relevant information from an external knowledge sources, rather than trying to memorize all facts within its own parameters. For instance, Retrieval-Augmented Language Model (REALM) \cite{guu2020retrieval} augments an LLM with a separate neural retriever that searches a knowledge corpus like Wikipedia\footnote{\url{https://www.wikipedia.org/}} for relevant documents to aid in predicting masked tokens. Both the retriever and the LM—is trained end-to-end using a modified MLM objective. This process encourages the retriever to find the most useful information for the LM to make accurate predictions, resulting in a more knowledgeable and interpretable model.

\item \textbf{Prompt Pre-Training (PPT):} It involves augmentation of pre-training data with task prompts to reduce the total amount of tokens required for training and even enable zero-shot abilities for certain tasks. One such LLM utilizing PPT is Galactica \cite{taylor2022galactica} which embeds task-specific prompts directly into the training data, enabling strong performance on scientific tasks like summarization and question answering without extra fine-tuning. It improves efficiency at smaller scales and enhances both generality and task-specific capabilities outperforming LLMs like PaLM \cite{chowdhery2023palm}, GPT-3 \cite{brown2020language} and BLOOM \cite{workshop2022bloom} on certain tasks.

\end{itemize}

\subsubsection{Fine-Tuning}
\label{ft}
\begin{figure}
    \centering
    \includegraphics[width=0.6\linewidth]{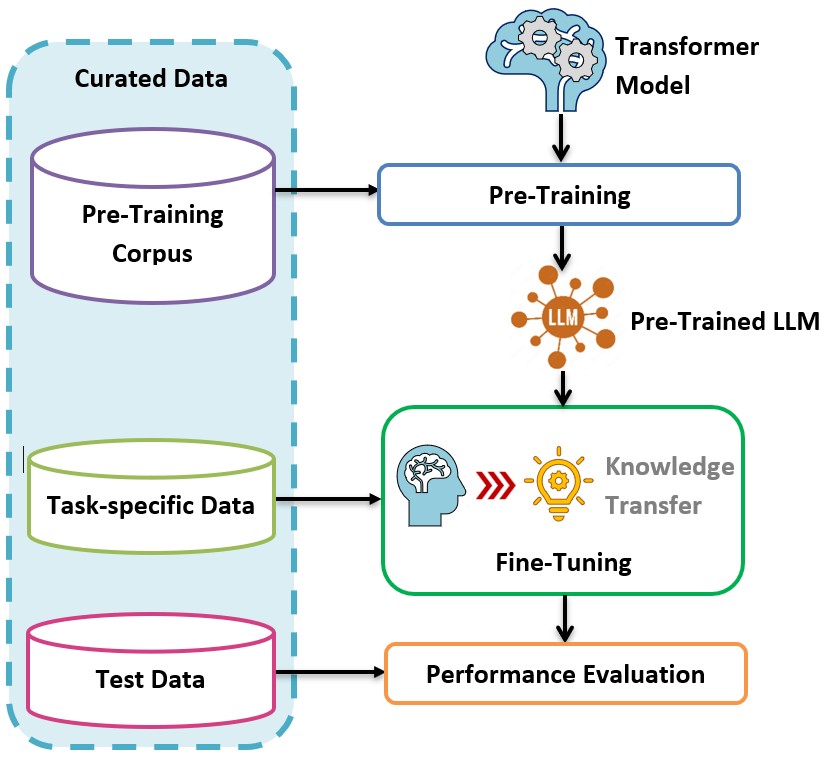}
    \caption{Illustration of fine-tuning process}
    \label{fig22}       
    \end{figure}

To make a pre-trained LM excel at a domain-specific task, additional \textit{fine-tuning} is required upon a smaller set of annotated samples specific to the downstream task as illustrated in Figure \ref{fig22}. It involves unfreezing a few layers of the LLM while retaining the weights of the other layers calculated during pre-training. Usually, the output layer is customized as per the output representation format and fine-tuned along with a few other unfrozen layers upon the task-specific data. Unfreezing minimum number of layers minimizes the number of LLM parameters to fine-tune. Unfreezing more layers increases the computational requirements but can enhance the accuracy of the downstream task. This applies only if abundant data is available for fine-tuning. In most cases, fine-tuning only the last few layers can obtain desirable results \cite{rogers2021primer}. This is due to the fact the lower layers capture low-level, local features primarily related to the syntax. Whereas, the higher layers capture the global information involving high-level semantic abstractions specific to the task at hand. The efforts to fine-tune LLMs by involving fewer parameters, collectively referred to as \textit{Parameter-Efficient Fine-Tuning (PEFT)} have been discussed herein-below as well as portrayed in Figure \ref{fig:lit_surv9}.

\begin{figure*}
    \centering

\tikzset{
    basic/.style  = {draw, text width=2cm, align=center, fill=pink!10!blue!80!red!10, font=\sffamily, rectangle},
    root/.style   = {basic, rounded corners=2pt, thin, align=center, fill=green!30},
    onode/.style = {basic, thin, rounded corners=2pt, align=center, fill=green!90!yellow!10,text width=1.5cm,},
    tnode/.style = {basic, thin, align=left, fill=pink!20, text width=2.1cm, align=center},
    xnode/.style = {basic, thin, rounded corners=2pt, align=center,  fill=pink!60, text width=5.4cm,},
    wnode/.style = {basic, thin, align=left, fill=pink!0, text width=6cm},
    edge from parent/.style={draw=black, edge from parent fork right}

}
\scalebox{0.75}{%
\begin{forest} for tree={
    grow=east,
    growth parent anchor=west,
    parent anchor=east,
    child anchor=west,
    edge path={\noexpand\path[\forestoption{edge},->, >={latex}] 
         (!u.parent anchor) -- +(6pt,0pt) |-  (.child anchor) 
         \forestoption{edge label};}
}
[Fine-Tuning, basic,  l sep=6mm,
 	[Gradual Unfreezing,  xnode, l sep=6mm,
        [T5 \cite{raffel2020exploring}, wnode]]
 	 	[Parameter Subset Tuning (PST),  xnode, l sep=6mm,
			[BitFit  \cite{ben2022bitfit};
			Child Tuning \cite{xu2021raise}, wnode]]
 	 	[Adapter Tuning,  xnode, l sep=6mm,
			[Adapter Blocks \cite{houlsby2019parameter};
			Low-Rank Hypercomplex Adapter \cite{karimi2021compacter};
			Intrinsic Dimensionality \cite{aghajanyan2021intrinsic};
			Adaptable Adapters \cite{moosavi2022adaptable};
			AdaMix \cite{wang2022adamix}, wnode]]
 	 	[Prefix Tuning,  xnode, l sep=6mm,
			[Prefix Tuning \cite{li2021prefix}, wnode]]
 	 	[Instruction Tuning,  xnode, l sep=6mm,
			[FLAN \cite{wei2021finetuned}, wnode]]
 	 	[Low-Rank Adaptation (LoRA),  xnode, l sep=6mm,
			[LoRA \cite{hu2022lora};
			DyLoRA \cite{valipour2023dylora};
			AdaLoRA \cite{zhang2023adaptive};
			SoRA \cite{ding2023sparse} ;
			QLoRA \cite{dettmers2023qlora};
			QuAILoRA \cite{lawton2024quailora}, wnode]]] 
\end{forest}
}
    \caption{Taxonomy of the PEFT strategies for LLMs}
    \label{fig:lit_surv9}
\end{figure*}

\begin{itemize}
    \item \textbf{Gradual Unfreezing:} Here, the objective is to fine-tune the frozen model with limited data during initiation and unfreeze other layers in due course. In T5, the top layer is unfreezed \cite{raffel2020exploring} and subsequently the preceding layers are unfreezed until the entire network is fine-tuned. The rationale behind this is to prevent the model from "forgetting" its general language understanding and to make the fine-tuning process more efficient, especially with limited data.

    \item \textbf{Parameter Subset Tuning (PST):} It operates on the principle that the pre-trained weights of an LLM already contain a significant amount of linguistic knowledge, and fine-tuning a small, specific subset of parameters is enough to adapt this knowledge to a new task. In BitFit  \cite{ben2022bitfit}, only the bias terms are adjusted while keeping majority of the parameters frozen. Child Tuning \cite{xu2021raise} identifies a child network, i.e. a subset of network parameters relevant to a given task by estimating their Fisher information representing how much a parameter contributes to the final predictions. Thereafter, only the child network is fine-tuned.

    \item \textbf{Adapter Tuning:} It involves \textit{adapter modules}, i.e. an isolated network that is fine-tuned for any new task keeping most of the parameters of the LLM intact. Houlsby et al. \cite{houlsby2019parameter} appended adapter blocks consisting of dense layers with ReLU activation after the feed-forward layers in the Transformer. With only these task-specific layers being fine-tuned, the number of parameter updates reduces significantly. 
    
    Further variations include utilizing the Kronecker product of low-rank matrices for the construction of parameter matrices for the adapter \cite{karimi2021compacter}. Another variation involves reparameterization to low-dimensional subspaces for fine-tuning, enhancing efficiency by reducing the number of parameter updates \cite{aghajanyan2021intrinsic}. One drawback of the adapter approach is that it raises the overall model parameters leading to more computations during inference. This hindrance was resolved through Adaptable Adapters applying differing activations specific to each layer as well as datasets accompanied with a switch trained to select appropriate layers of the adapter module \cite{moosavi2022adaptable}. Furthermore, AdaMix combined various parameter-efficient adapters to provide SOTA results with an efficiency equivalent to fine-tuning with a single adapter module \cite{wang2022adamix}.

    \item \textbf{Prefix Tuning:} The key idea is to freeze the original parameters and instead fine-tune a small sequence of trainable, continuous, task-specific vector called "prefix." This prefix is pre-pended to the input sequence before it is fed into the frozen model. During fine-tuning, the model's weights remain fixed, and only the parameters of this prefix are updated. This approach essentially fine-tunes the model to condition its generation on the prefix, allowing the prefix to act as a "virtual prompt" that steers the frozen LLM toward a desired behavior. It is observed that fine-tuning prefix parameters accounting for just 0.1\% of the total LM parameters is sufficient to achieve appreciable performance \cite{li2021prefix}.

    \item \textbf{Instruction Tuning:} This technique was popularized by the Fine-tuned LAnguage Net (FLAN) \cite{wei2021finetuned}. FLAN involves fine-tuning a 137B parameter LM on over 60 datasets accompanied by instructional descriptions of the tasks in natural language. Instruction tuning enabled FLAN to outperform models with greater number of parameters like GPT-3 on zero-shot learning tasks. 
    
    \item \textbf{Low-Rank Adaptation (LoRA):} It freezes the original pre-trained weights and introduces trainable low-rank matrices into each linear layer of the Transformer architecture \cite{hu2022lora}. Instead of updating the full weight matrix, it approximates changes using a decomposition of the form $\Delta W \approx BA$, where $W \in \mathbb{R}^{d \times k}$ denotes the frozen weight matrix while $B \in \mathbb{R}^{d \times r}$ and $A \in \mathbb{R}^{r \times k}$ are trainable low-rank matrices having rank $r \ll \min(d, k)$. \textit{LoRA} \cite{hu2022lora} reduces the number of trainable parameters up to 10,000$\times$ in LLMs like GPT-3 with 3$\times$ lower hardware requirements compared to traditional fine-tuning. However, it applies a uniform rank across all layers, which may overlook the varying significance of different model components.

   Over time, various refinements to LoRA have been proposed. Dynamic Search-Free Low Rank Adaptation (DyLoRA) \cite{valipour2023dylora} trains LoRA adapter modules across a spectrum of ranks simultaneously, instead of a fixed rank. It sorts and evaluates the adapter outputs at various ranks, enabling the model to adaptively select effective low-rank structures eliminating the need for exhaustive rank search. This results in 4$\times$ to 7$\times$ faster fine-tuning—up while maintaining strong performance across tasks. Adaptive LoRA (AdaLoRA) \cite{zhang2023adaptive} facilitates dynamic rank allocation by applying Singular Value Decomposition (SVD) to determine the importance score of weight matrices. It then assigns rank to weight matrices based on its contribution to the model's performance. This adaptive approach ensures the most important parts of the LLM get the most fine-tuning attention. Sparse Low-rank Adaptation (SoRA) \cite{ding2023sparse} focuses on dynamically adjusting the rank of the incremental matrices. It uses a trainable gate to continuously modify the rank during fine-tuning. This gate is optimized using proximal gradient, allowing the LLM to determine the sparsity of matrices to be updated. This provides flexibility on the fly with added benefit of parameter-efficiency during fine-tuning. Quantized LoRA (QLoRA) \cite{dettmers2023qlora} further reduces memory consumption by combining quantization with LoRA adapters. It uses a specialized data type called NormalFloat (NF4), which compresses weights based on their normal distribution, preserving important values while minimizing memory. Additionally, it uses Double Quantization to quantize the quantization constants too, cutting overhead from 0.5 to 0.127 bits per parameter. During fine-tuning, weights are stored in NF4 format and dequantized into 16-bit BrainFloat (BF16) during computation. To preserve efficiency, only the gradients for the LoRA adapter weights are computed freezing the quantized weights of the pre-trained LM. Furthermore, Paged Optimizers manage memory dynamically between CPU and GPU, preventing crashes when GPU memory is exhausted. Altogether, QLoRA enables scalable fine-tuning of massive models on consumer-grade hardware without sacrificing performance. Quantization-Aware Initialization for LoRA (QuAILoRA) \cite{lawton2024quailora} is an enhancement to QLoRA that uses a data-driven method to align the quantized model’s output more closely with its full-precision counterpart from the start instead of initializing LoRA matrices randomly. It provides the benefits of 8-bit quantization while maintaining the memory efficiency of 4-bit LMs, making it a powerful tool for efficient LLM adaptation.    
\end{itemize}

    \begin{figure*}[h]
    \centering
    \includegraphics[width=0.8\linewidth]{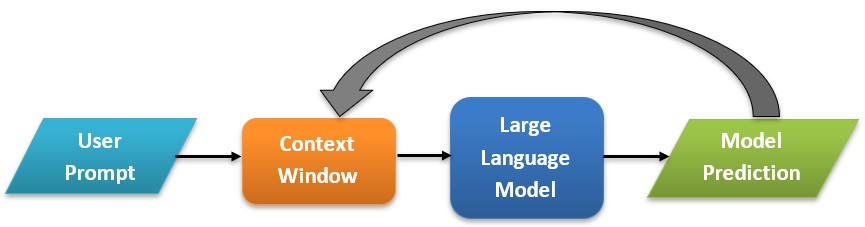}
    \caption{Illustration of prompt-engineering process}
    \label{fig7}       
    \end{figure*}

\subsubsection{Prompt-Engineering}
\label{pe}
The multi-task learning ability of generative LLMs was first demonstrated by GPT-2 \cite{radford2019language}. It was capable of performing various tasks out of the box minimizing manual effort during inference. Its subsequent version, GPT-3 \cite{brown2020language} was further able to perform few-shot or \textit{In-Context Learning (ICL)}, i.e. it could generate the required predictions just by providing the task description along with priming with a few use cases. ICL significantly reduced the computational complexity due to zero parameter updates in the pre-trained LLM. This led to the emergence of the term \textit{Prompt-Engineering} followed by a series of developments. Schick and Schutze \cite{schick2021generating} utilized a pre-trained LLM for ICL which excelled at tasks such as predicting the next sentence given the first sentence, generating the second sentence replacing "\_\_" in the prompt or even generating both the sentences by providing the description. Wei et al. \cite{wei2021finetuned} further demonstrated that an LLM's multi-task learning abilities can be enhanced by prompt-learning on several supervised datasets concerning various tasks. The categories of prompt-engineering have been mentioned herein-below, while the basic flow of steps has been depicted in Figure \ref{fig7}.

\begin{itemize}
    \item \textbf{Instruction-based Learning:} Also known as \textit{Priming}, requires the task instructions to be specified optionally with a few samples of the inputs and the corresponding outputs \cite{wei2021finetuned, schick2021few}. For instance, providing the instruction to perform translation accompanied with a few examples in the prompt to prime the LLM to generate a translation for any new sentence. 
    
    \item \textbf{Template-based Learning:} It deals with exploiting predefined structures, known as \textit{templates} to construct prompts. The templates can be designed as \textit{cloze styled}, i.e. inserting placeholders in the prompt text and attempting to fill in the blanks \cite{schick2021exploiting}, \textit{multiple-choice type}- providing multiple hypotheses in the template and asking the model to choose the correct one \cite{trinh2018simple} or \textit{prefix-type}- adding special prefixes before the input to denote the task to be performed on the input \cite{schick2021few}. 
    
    \item \textbf{Proxy-Task-based Learning:} It involves probing an LLM with a proxy-task, i.e. a related task sharing some attributes of the original task, to obtain the output of the original task through transferring the inference to the desired form. This enhances the efficiency and eases inference by utilizing simpler tasks closer to those upon which the LLM has been previously trained to perform tasks leveraging rigorous linguistic comprehension. Instances include applying textual entailment for topic detection \cite{yin2019benchmarking} or achieving coreference resolution through question answering \cite{wu2020corefqa}.

    \item \textbf{Chain-of-Thoughts (CoT) Prompting:} It enhances reasoning in LLMs by guiding them to generate intermediate steps before arriving at a final answer. It encourages the LLM to break down complex problems into a sequence of logical steps expressed in natural language \cite{wei2022chain}. This aids in better access and application of the pre-trained knowledge of LLMs. It significantly improves performance on tasks involving arithmetic, commonsense, and symbolic reasoning, especially in LLMs like PaLM and GPT-3, often surpassing traditional fine-tuning approaches. However, generating intermediate steps increase the overall context length. This translates into high computational requirements and high latency. 

\end{itemize}

\begin{figure*}
    \centering

\tikzset{
    basic/.style  = {draw, text width=2cm, align=center, fill=pink!10!blue!80!red!10, font=\sffamily, rectangle},
    root/.style   = {basic, rounded corners=2pt, thin, align=center, fill=green!30},
    onode/.style = {basic, thin, rounded corners=2pt, align=center, fill=green!90!yellow!10,text width=1.5cm,},
    tnode/.style = {basic, thin, align=left, fill=pink!20, text width=2.1cm, align=center},
    xnode/.style = {basic, thin, rounded corners=2pt, align=center,  fill=pink!60, text width=6cm,},
    wnode/.style = {basic, thin, align=left, fill=pink!0, text width=6cm},
    edge from parent/.style={draw=black, edge from parent fork right}

}
\scalebox{0.75}{%
\begin{forest} for tree={
    grow=east,
    growth parent anchor=west,
    parent anchor=east,
    child anchor=west,
    edge path={\noexpand\path[\forestoption{edge},->, >={latex}] 
         (!u.parent anchor) -- +(6pt,0pt) |-  (.child anchor) 
         \forestoption{edge label};}
}
[Prompt-Engineering, basic,  l sep=6mm,
 	[Prompt Pruning,  xnode, l sep=6mm,
		[DYNAICL \cite{zhou2023efficient};
		Selective Context \cite{li2023compressing};
		STDC \cite{yin2023did};
		PCRL \cite{jung2024discrete};
		LLMLingua \cite{jiang2023llmlingua};
		CoT-Influx \cite{huang2023fewer}, wnode]] 
 	[Prompt Summary	,  xnode, l sep=6mm,
		[RECOMP \cite{xu2024recomp};
		Semantic Compression \cite{fei2024extending}, wnode]] 
 	[Soft Prompt Compression,  xnode, l sep=6mm,
		[PromptCompression \cite{wingate2022prompt};
		AutoCompressors \cite{chevalier2023adapting};
		ICAE \cite{ge2024incontext}, wnode]] 
 	[Intermediate Reasoning Optimization,  xnode, l sep=6mm,
		[SoT \cite{ning2023skeleton};
		Coconut \cite{hao2024training};
		CCoT \cite{nayab2024concise};
		TALE \cite{han2025token};
		CoD \cite{xu2025chain}, wnode]]] 
\end{forest}
}
    \caption{Taxonomy of efficient prompt-engineering strategies for LLMs}
    \label{fig:lit_surv10}
\end{figure*}

Apart from these, certain prompt-engineering practices can enhance efficiency as illustrated in Figure \ref{fig:lit_surv10}. These efforts have been described herein-below.

\begin{itemize}
    \item \textbf{Prompt Pruning:} It comprises of strategies to reduce the size of input prompts by removing redundant or unimportant information (tokens) to save computational resources and improve inference speed. DYNAICL \cite{zhou2023efficient} uses a meta-controller LLM to dynamically select an optimal number of in-context examples for a given prompt, adapting to the available computational budget. It focuses on optimizing the number of examples rather than the content of the prompt itself. Selective Context \cite{li2023compressing} merges tokens into larger units and then prunes these units based on their self-information, which is measured as the negative log-likelihood. The core idea is that units with high self-information are more informative and should be retained, while redundant or low-information units are removed. Syntax-guided Task Definition Compression (STDC) \cite{yin2023did} prunes prompts based on their parse tree. It iteratively removes phrase nodes that have the smallest impact on the LLM's performance, effectively simplifying the prompt while retaining its most critical structural and semantic information. Prompt Compression with Reinforcement Learning (PCRL) \cite{jung2024discrete} employs a Reinforcement Learning (RL) approach. It trains a separate \textit{policy LLM} to decide which tokens to remove from a prompt. The reward function is the trade-off between compression ratio, i.e. reduction the prompt length; and faithfulness, i.e. ensuring that the output from the pruned prompt is similar to the output from the original unpruned prompt. LLMLingua \cite{jiang2023llmlingua} uses a multi-stage pruning approach progressing from coarse to fine. Firstly, a broad demonstration-level pruning is carried out, removing entire examples from the prompt. Then token-level pruning is performed based on perplexity, i.e. a measure of how well an LLM predicts a sequence of tokens. Moreover, it employs a budget controller that intelligently distributes the pruning budget across different sections of the prompt. CoT-Influx \cite{huang2023fewer} applies a similar coarse-to-fine pruning methodology specifically to Chain-of-Thought (CoT) prompts. It uses RL to determine which parts of the CoT prompt to remove. Initially, it prunes less important CoT examples entirely. Then, it prunes unimportant tokens from the remaining, more critical examples. This two-step process streamlines complex CoT prompts, reducing the computational load while preserving the reasoning steps necessary for accurate a response.

\item \textbf{Prompt Summary:} It focuses on condensing the original prompt while retaining its semantic integrity. Unlike prompt pruning that selectively removes less important elements while keeping the rest intact, summarization transforms the entire prompt into a compact representation. RECOMP (abbreviation for Retrieve, Compress, Prepend) \cite{xu2024recomp} proposes an Abstractive Compressor that generates concise summaries from input questions and retrieved documents. It leverages extreme-scale LLMs to distill a lightweight summarization model. Semantic Compression \cite{fei2024extending} adopts a topic-based approach: it first segments the text into individual sentences, clusters them by topic, and then summarizes each group to produce a semantically compressed prompt.

    \item \textbf{Soft Prompt Compression:} It focuses on replacing lengthy input prompts with compact, learnable sequences known as soft prompts. These soft prompts consist of continuous, trainable tokens that serve as efficient substitutes for fixed prompts like system or task-specific instructions. PromptCompression \cite{wingate2022prompt} introduces a technique that prepends a set of soft tokens to the input. These tokens are optimized through backpropagation on a prompt dataset. Once tuned, the resulting soft token sequence effectively captures the essence of the original prompt in a much shorter form. AutoCompressors \cite{chevalier2023adapting} transform LLMs into context-aware compressors by training them to generate summary vectors from long documents. These vectors act as soft prompts for subsequent segments, enabling the model to handle sequences with 30k tokens. It intuitively mitigates context window limitations by converting prior context into latent representations. In-context Autoencoder (ICAE) \cite{ge2024incontext} introduces a lightweight autoencoder architecture where a LoRA-adapted LLM serves as the encoder and the target LLM acts as the decoder. It compresses long contexts into compact memory slots, achieving up to 4$\times$ compression.

    \item \textbf{Intermediate Reasoning Optimization:} It  refers to techniques that improve the efficiency and effectiveness of LLMs during multi-step reasoning. These techniques streamline intermediate reasoning steps by condensing, parallelizing, or imposing constraints. This enables human-like problem-solving efficiency with fewer tokens, and lower latency. Skeleton-of-Thought (SoT) \cite{ning2023skeleton} mimics how humans plan and write by first creating a high-level outline before filling in the details. At first, an LLM is prompted to first generate a concise, high-level "skeleton" or outline of the answer. Each of the points in the skeleton is used as a separate prompt. Then the LLM generates the content for all points in parallel using either batched decoding (for open-source models) or parallel API calls (for closed-source models). The final answer is then assembled from these simultaneously generated segments. It significantly cuts down on end-to-end generation time by replacing a single, long sequential decoding process with multiple, shorter parallel ones. However for tasks requiring step-by-step reasoning having causal dependencies among the points, SoT might not be an effective solution. Chain of Continuous Thought (Coconut) \cite{hao2024training} allows the LLM to reason in its internal, continuous latent space. By reasoning in the latent space, LLMs can perform a more flexible, breadth-first search of possible solutions. This enhances accuracy while requiring fewer inference by simultaneously exploring multiple reasoning trajectories. Instead of generating a word token for each reasoning step as in CoT, the model's last hidden state is fed back directly as the next input embedding. This process, referred to as "continuous thought," enables the model to explore reasoning paths that are not constrained by natural language syntax or semantics. Constrained-CoT (CCoT) \cite{nayab2024concise} aims to generate shorter, more focused reasoning steps that capture the core logic without unnecessary filler words or redundant explanations. The prompt is designed with an explicit instruction to limit the output to a specific word count. This encourages the LLM to condense its reasoning and provide a more concise response with reduced latency. Token-Budget-Aware LLM Reasoning (TALE) \cite{han2025token} makes reasoning more efficient by explicitly considering a token budget as a hard constraint and tunes the LLM to reason effectively within that limit. It uses RL with a reward function as the trade-off between the quality of the reasoning, i.e. correctness and its adherence to the budget. It enforces an adaptive approach encouraging the LLM to not only generate concise reasoning with tight budget but also generate extensive outputs with generous budget. Chain-of-Draft (CoD) \cite{xu2025chain} overcomes the verbose reasoning process of CoT by producing concise reasoning outputs highlighting critical insights only as intermediate steps. It mimics the human reasoning process of jotting down key intermediate outcomes during problem-solving. It delivers performance on par with CoT while consuming just 7.6\% tokens with respect to CoT.

\end{itemize}

\begin{figure}[!t]
    \centering
    \includegraphics[width=0.75\linewidth]{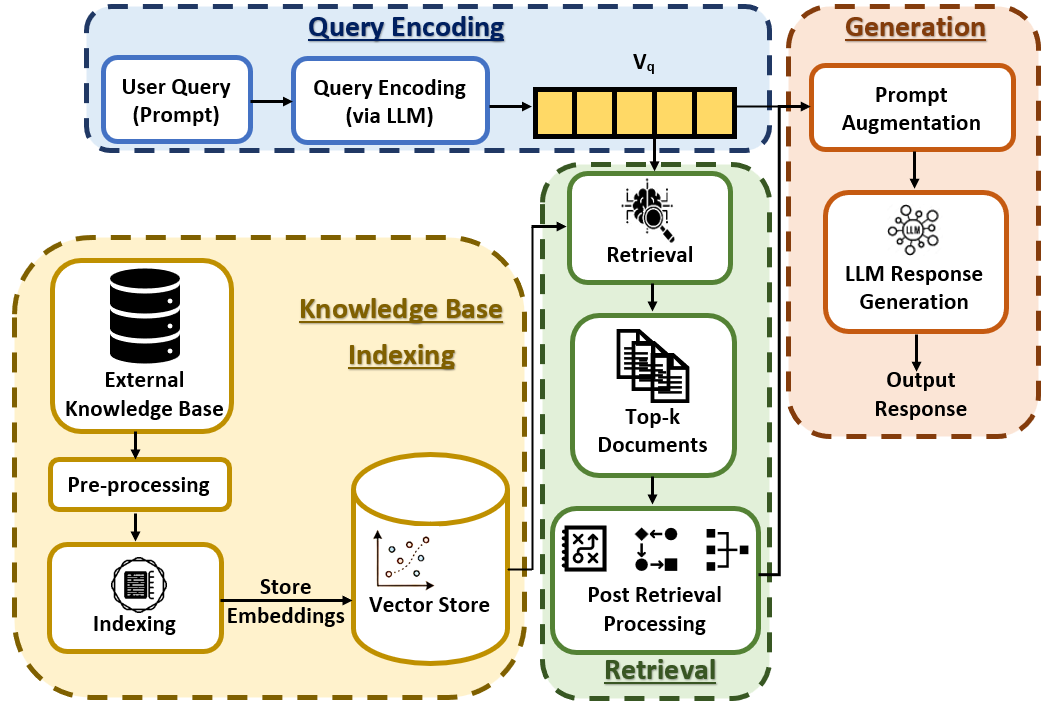}
    \caption{Illustration of RAG process.}
    \label{fig12}       
    \end{figure}

\subsubsection{Retrieval-Augmented Generation (RAG)}
\label{rag}
Conventionally, the knowledge of LLMs were limited to the data on which its learning has been done. This necessitated regularly updating the LLM on new data to obtain responses on a new context. However, tuning LLMs is computationally intensive as well as time consuming. Also, it is practically not possible to keep on tuning the LLM to ensure that it is always up-to-date with the current state of happenings in the real world. To deal with this issue in a practical and efficient way, RAG was formulated. RAG searches one or more external knowledge bases up-to-date with the information required for a task, and augments this knowledge with the context fed to the LLM prompt \cite{lewis2020retrieval}. A typical RAG-based LLM has been illustrated in Figure \ref{fig12}. It first encodes a user query (prompt) into a vector, retrieves top-k relevant documents from a vector store (knowledge-base) using similarity search. It then feeds both the query and retrieved content into a generator (often an LLM) to produce the final output by reasoning on the additional context provided by the retriever. This enables the LLM to produce grounded, accurate, and up-to-date responses without requiring retraining. Various recent works demonstrate a significant progress in RAG as depicted in Figure \ref{fig:lit_surv11} and described herein-below.

\begin{figure*}
    \centering

\tikzset{
    basic/.style  = {draw, text width=3.6cm, align=center, fill=pink!10!blue!80!red!10, font=\sffamily, rectangle},
    root/.style   = {basic, rounded corners=2pt, thin, align=center, fill=green!30},
    onode/.style = {basic, thin, rounded corners=2pt, align=center, fill=green!90!yellow!10,text width=1.5cm,},
    tnode/.style = {basic, thin, align=left, fill=pink!20, text width=2.1cm, align=center},
    xnode/.style = {basic, thin, rounded corners=2pt, align=center,  fill=pink!60, text width=4.5cm,},
    wnode/.style = {basic, thin, align=left, fill=pink!0, text width=6cm},
    edge from parent/.style={draw=black, edge from parent fork right}

}
\scalebox{0.75}{%
\begin{forest} for tree={
    grow=east,
    growth parent anchor=west,
    parent anchor=east,
    child anchor=west,
    edge path={\noexpand\path[\forestoption{edge},->, >={latex}] 
         (!u.parent anchor) -- +(6pt,0pt) |-  (.child anchor) 
         \forestoption{edge label};}
}
[Retrieval-Augmented Generation (RAG), basic,  l sep=6mm,
 	[Querying and Retrieval,  xnode, l sep=6mm,
		[Self-RAG \cite{asai2023self};
		CRAG \cite{yan2024corrective};
		Adaptive-RAG \cite{jeong2024adaptive};
		AT-RAG \cite{rezaei2024rag};
		Query Rewriter+ \cite{shi2024enhancing}, wnode]] 
 	[Post-Retrieval Restructuring,  xnode, l sep=6mm,
		[RQ-RAG \cite{chan2024rqrag};
		G-RAG \cite{dong2024don};
		Refiner \cite{li2024refiner};
		MapReduce Approach \cite{zhang2024mapreduce}, wnode]]]
\end{forest}
}

    \caption{Taxonomy of RAG strategies}
    \label{fig:lit_surv11}
\end{figure*}

\begin{itemize}

    \item {\textbf{Querying and Retrieval:}} It deals with how relevant information is fetched from external databases or knowledge-bases to support an LLM's context enabling it to provide a well-informed response. Various research works have been carried out to ensure the LLM gets high-quality, context-rich input with minimal compute or memory utilization as discussed herein. Self-RAG \cite{asai2023self} boosts the factuality and adaptability by integrating on-demand retrieval with self-reflection mechanisms. Instead of blindly incorporating a fixed number of retrieved documents, Self-RAG tunes an LLM to decide what to retrieve, generate responses, and then critique its own outputs using special reflection tokens. These tokens allow the model to assess the relevance and accuracy of both the retrieved content and its own generation, enabling controlled and context-sensitive behavior during inference. Corrective Retrieval Augmented Generation (CRAG) \cite{yan2024corrective} uses a lightweight retrieval evaluator that assesses the quality of retrieved documents and assigns a confidence score. Based on this score, it triggers different actions like carrying on further search or refining the retrieved content through a "decompose-then-recompose" algorithm that filters out noise and focuses on key information. This adaptive mechanism makes it highly resilient to retrieval errors and hallucinations. Adaptive-RAG \cite{jeong2024adaptive} applies a small LM classifier trained on auto-annotated data to predict the complexity level of each query. Based on it, it selects the most appropriate strategy among options like skipping retrieval for simple factual questions, using single-step retrieval for moderately complex ones, or applying multi-hop iterative retrieval for deeply layered queries. This adaptive mechanism improves efficiency by reducing unnecessary computational overhead while ensuring accuracy through nuanced responses for complex tasks. AT-RAG \cite{rezaei2024rag} enhances query efficiency and answer relevance through two core innovations: topic filtering and iterative reasoning. By leveraging BERTopic \cite{grootendorst2022bertopic}, it dynamically narrows the retrieval space based on the semantic topic of each query, reducing noise and improving precision. It then applies multi-hop iterative reasoning, allowing the system to refine its understanding and retrieval over successive steps. These contributions aid in reducing retrieval time while ensuring relevance and correctness for complex queries. Shi et al. \cite{shi2024enhancing} proposed four synergistic modules such as Query Rewriter+, which generates diverse, instruction-tuned queries to overcome ambiguity; Knowledge Filter, which uses natural language inference to discard irrelevant information; Memory Knowledge Reservoir, which caches previously retrieved knowledge to reduce redundancy and latency; and Retriever Trigger, which intelligently decides whether external retrieval is necessary based on query confidence. Together, these modules reduce retrieval costs by 71\%, response time by around 50\%, and improve answer quality.

    \item {\textbf{Post-Retrieval Restructuring:}} It involves an array of approaches to restructure the context post-retrieval to make it concise and enrich its informativeness. Refined Query Retrieval-Augmented Generation (RQ-RAG) \cite{chan2024rqrag} addresses the limitation of reliance on raw, often ambiguous or overly broad queries that can lead to poor retrieval and hallucinated answers. It tunes the LLM to explicitly rewrite, decompose, and disambiguate queries before retrieval. This ensures that the retrieval step pulls in more relevant and focused documents. Graph RAG (G-RAG) \cite{dong2024don} is a graph-based re-ranking method leveraging inter-document relationships and semantic structure. It constructs a graph neural network (GNN) between retrieved documents, enriched with Abstract Meaning Representation (AMR) graphs, to model semantic and structural connections. The re-ranker sits between the retriever and the generator, elevating documents that are not only individually relevant but also contextually connected to others. Refiner \cite{li2024refiner} is an end-to-end framework that extracts and reorganizes query-relevant content using a single decoder-only LM, effectively mitigating the “lost-in-the-middle” issue common in long-context inputs. To help LLMs better align with the original context, Refiner preserves query-relevant content verbatim and selectively retains surrounding context. It organizes extracted information into hierarchical sections, grouping related content together while separating unrelated pieces, thereby enhancing the clarity and coherence of chunk-level information for improved comprehension. In another work, a MapReduce Approach \cite{zhang2024mapreduce} was designed to address the “lost-in-the-middle” problem by context mapping, i.e. breaking down long documents into smaller, manageable segments to process them individually. Then context reduction, i.e. aggregating the most relevant insights for final generation is performed. This approach ensures efficiency and scalability while improving robustness, factuality, and safety of LLM outputs, especially in critical domains like healthcare.

\end{itemize}

\section{RQ5: Evaluation Benchmarks and Measures}
\label{ev}

\subsection{LLM Benchmarks}
\label{bm}
The benchmarks and tasks for the evaluation of the efficacy of LLMs have been discussed herein-below.

\begin{itemize}
    \item \textbf{Arena Elo:} It is a crowdsourced leaderboard system used to rank LLMs based on head-to-head comparisons in a democratic way\footnote{\url{https://openlm.ai/chatbot-arena/}}. It compiles the votes of over 5M users to compute Elo rating relying on LLM responses in randomized match-ups. The Elo rating which has been widely used in ranking chess players, reflects a model’s relative performance across millions of interactions. Given the Elo ratings $R_E^A$ and $R_E^B$ of two players $A$ and $B$, the winning probability $P_W(A)$ of $A$ is computed as follows:

\begin{equation}
\label{eq28}
P_W(A)= \frac{1}{1+10^{(R_E^B -R_E^A)/400}}
\end{equation}

\item \textbf{Artificial Analysis Intelligence Index (AAII):} It is a comprehensive benchmarking suite for English texts developed by Artificial Analysis\footnote{\url{https://artificialanalysis.ai/methodology/intelligence-benchmarking}} to evaluate LLMs across capabilities like reasoning, knowledge, mathematics, and programming. It aggregates results from eight datasets, including MMLU-Pro, AIME 2025, and LiveCodeBench, offering holistic assessment of an LLM’s capabilities. The LLMs are evaluated under uniform conditions with consistent prompts and temperature settings. The evaluations are zero-shot, relying solely on instruction prompts without examples, aligning with modern chat-style models.

\item \textbf{Terminal-Bench (TB) Hard:} It represents the 'Hard' variant of the Terminal-Bench (TB) Core\footnote{\url{https://www.tbench.ai/}}, a coding benchmark that tests LLMs on complex command-line tasks. It includes 47 tasks simulating real-world terminal environments where models must execute shell commands, manage files, and handle system-level operations. This benchmark is part of the AAII and is designed to push models beyond basic scripting into advanced system interactions.

\item \textbf{Artificial Analysis Long Context Reasoning (AA-LCR):} It consists of 100 challenging text-based questions designed to test reasoning across multiple real-world documents, each averaging around 100,000 tokens\footnote{\url{https://huggingface.co/datasets/ArtificialAnalysis/AA-LCR}}. Questions were crafted to require multi-document synthesis and validated using non-frontier models to ensure difficulty. Human evaluators confirmed the clarity and validity of each question, with modest first-attempt accuracy but strong agreement on correct answers. It emphasizes genuine reasoning over retrieval and spans seven document types, demanding both general and mathematical analysis.

\item \textbf{Humanity’s Last Exam (HLE): } It is a comprehensive multi-modal benchmark designed to test LLMs with human academic knowledge \cite{phan2025humanity}. It features 2,500 challenging closed-ended questions spanning over 100 subjects, developed through a global collaboration of nearly 1,000 expert contributors from more than 500 institutions across 50 countries. The dataset is publicly released, with a private test set reserved to monitor overfitting.

\item \textbf{Massive Multitask Language Understanding (MMLU) Pro:} It is an advanced version of the MMLU benchmark \cite{wang2024mmlu}. It includes 12,032 reasoning-intensive questions across Math, Physics, Chemistry, Law, Engineering, Economics, Health, Psychology, Business, Biology, Philosophy, and Computer Science. It expands answer choices from four to ten accompanied with removal of trivial and noisy items, resulting in a more challenging dataset causing a 16–33\% drop in model accuracy compared to MMLU. Despite the increased difficulty, MMLU-Pro shows greater robustness across 24 prompt styles, reducing score sensitivity by over 2\%. 

\item \textbf{Graduate-Level Google-Proof Q\&A (GPQA) Diamond:} It is the "Diamond" subset of the GPQA dataset \cite{rein2024gpqa} of 198 multiple-choice questions in physics, chemistry and biology curated by domain experts to test deep scientific reasoning. Designed to be unsolvable via simple web searches, the questions challenge both humans and AI. Its difficulty makes it a valuable resource for developing scalable oversight methods, helping human experts supervise advanced AI systems in high-stake scientific contexts.

\item \textbf{LiveCodeBench (LCB):} It tests LLM's ability to write and execute code in real-time environments \cite{jain2025livecodebench} . It includes tasks across various programming languages and assesses debugging, syntax accuracy, and runtime behavior. It continuously aggregates new problems from LeetCode, AtCoder, and CodeForces. It currently includes 400 high-quality challenges published between May 2023 and May 2024. Unlike traditional benchmarks focused solely on code generation, it also assesses capabilities like self-repair, code execution, and test output prediction. 

\item \textbf{SciCode:} It is a benchmark focused on scientific programming tasks \cite{tian2024scicode} developed through collaborative effort by experts across 16 disciplines, including mathematics, physics, chemistry, biology, and materials science. It comprises 80 complex problems broken down into 338 subproblems, each requiring knowledge recall, reasoning, and code synthesis. The dataset includes optional scientific background descriptions and gold-standard solutions with test cases for evaluation. 

\item \textbf{Instruction Following Benchmark (IFBench):} It evaluates LLMs on precise instruction-following across 58 diverse and provable out-of-domain constraints \cite{pyatkin2025generalizing}. It supports rigorous generalization testing and includes custom-built constraint verification modules. It includes tasks on counting, analogy, syllogism, and deductive reasoning.

\item \textbf{American Invitational Mathematics Examination (AIME) 2025:} It evaluates advanced mathematical reasoning in LLMs\footnote{\url{https://huggingface.co/datasets/yentinglin/aime_2025}}. It consists of 30 challenging problems that require creative problem-solving across algebra, geometry, number theory, and combinatorics. It emphasizes competition-style rigor, with questions crafted to be difficult even for top-performing students.

\item \textbf{MATH 500:} It is often used alongside AIME 2025 to form a robust math evaluation suite. It is a curated subset of 500 competition-level problems from the MATH dataset \cite{hendrycksmath2021}, designed to evaluate advanced reasoning across mathematical domains including geometry, algebra, number theory, pre-calculus, and probability. Each problem requires detailed step-by-step solution with LaTeX code to assess symbolic manipulation and logical precision.

\item \textbf{HumanEval:} It is a gold standard for evaluating and comparing the coding capabilities of various LMs \cite{chen2021evaluating}. It includes 164 programming tasks paired with prompts in natural language and reference solutions. These problems test the LLM’s ability for algorithmic reasoning and control flow in zero-shot settings.

\end{itemize}

\subsection{Efficiency Measures}
\label{em}
To measure the efficiency $\eta$, the trade-off between the model performance and cost factors needs to be calculated as shown in Equation (\ref{eq29}). 

\begin{equation}
\label{eq29}
\eta = \frac{Performance}{Cost Factors}
\end{equation}

For performance, it is necessary to analyze the Pareto-improvement through comparison with a benchmark (discussed in Section \ref{bm}), i.e. attaining higher accuracy at lower cost \cite{durlich2023concept}. Schwarts et al. \cite{schwartz2020green} formulated cost factors proportional to the time and resources for execution on a single sample $E^{s}$, data size $D^{s}$ and the number of epochs $n$ required for training as depicted in Equation (\ref{eq30}).

\begin{equation}
\label{eq30}
Cost Factors \propto E^{s} \cdot D^{s} \cdot n 
\end{equation}

Durlich et al. \cite{durlich2023concept} further refined Equation (\ref{eq30}) to arrive at the cost for inference by separately accounting $E^{s}$ and $D^{s}$ for training ($E^{s}_{T}$, $D^{s}_{T}$) and inference ($E^{s}_{I}$, $D^{s}_{I}$) as follows:

\begin{equation}
\label{eq31}
Cost Factors \propto E^{s}_{T} \cdot D^{s}_{T} \cdot n + E^{s}_{I} \cdot D^{s}_{I}
\end{equation}

The cost factors can be defined with respect to various metrics as follows:

\begin{itemize}
    \item \textbf{Floating-point Operations (FLOPs):} FLOPs needed for a single instance computation can serve as a consistent benchmark irrespective of the hardware for deployment\cite{treviso2023efficient}. However, existing HPCs with support for parallel processing might lead to non-uniform execution times even with the same number of FLOPs.

    \item \textbf{Inference Time:} It denotes the time required by the model to process a test input and generate a suitable response \cite{xu2023survey}. Unlike FLOPs, it is hardware-dependent, i.e. it depends upon the configuration of the HPC and support for parallel execution. From the evaluation perspective, it enables a real-time measure of various algorithms based on execution upon identical HPC.

    \item \textbf{Latency:} The delay between sending a request and receiving the first token from the LLM. For reasoning models, this corresponds to the first reasoning token. For non‑streaming models, it reflects the time until the entire output is received.

    \item \textbf{End-to-End Response Time (E2E RT):} Refers to the total time from when the user submits a query until they receive the complete response. It accounts for the inference time together with network bandwidth and delay due to request handling, pre-processing, post-processing and client-side rendering.
    
    \item \textbf{Throughput:} The speed at which tokens are delivered during inference, measured in tokens per second once streaming begins.

    \item \textbf{Context Window:} The maximum number of tokens that can be processed in a single request, combining both input and output. The output token capacity varies by model and is typically lower than the input token capacity.

    \item \textbf{Number of Parameters:} Refers to the total number of parameters comprising all weights and biases in the network. It includes both trainable and non-trainable parameters. For instance, during fine-tuning, a portion of model parameters is frozen (non-trainable), while the remaining parameters are updated (trainable).

    \item \textbf{Size:} The size is stated in terms of the amount of storage space required by the model. Larger model size generally means more capacity to learn complex patterns, but also higher computational and memory requirements. It is often estimated by the number of parameters. However, this might not hold true for quantized or mixed-precision models where the precision of the parameters is different from the default 32-bit floating point (FP32) precision.

    \item \textbf{Carbon Emissions: } Refer to the amount of greenhouse gases (measured in $CO_{2}$ equivalent) released into the atmosphere as a result of the energy consumed during pre-training, fine‑tuning, and inference of LLMs. It is the most significant indicator of the environmental impact due to LLMs. 

    \item \textbf{Power Consumption:} The amount of electrical energy consumed by an LLM during training or inference. It depends upon multiple factors like the FLOPs per watt of GPUs/ TPUs, number of GPUs/ TPUs deployed, cooling systems and data center overhead. A detailed discussion on estimating power consumption and carbon emissions of LLM has been presented in Section \ref{ecp}. 

    \item \textbf{Pricing:} It indicates cost incurred by the end user for availing LLM inferencing services. It is calculated as the average price per million tokens using a weighted mix of input and output token rates with the common ratio being 3:1\footnote{\url{https://www.asad.pw/llm-subscriptions-vs-apis-value-for-money/}}.
    
\end{itemize}

Despite the progress, the current approaches for measuring efficiency are not fool-proof. There lies a disparity in the carbon emissions reported by various monitoring applications as discussed in Section \ref{ecp}. Most of the studies only account for the computing resources and do not consider the cooling, networking or other operational costs. Moreover, relying on only one or two metrics to estimate efficiency might be inadequate, given the weak correlations among them. Some models might have high FLOPs even with fewer parameters. This can be observed in architectures employing parameter sharing as in ALBERT \cite{lan2019albert}. Conversely, some LLMs exhibit a very high parameter count but relatively low FLOPs, as in the case of MoE-based models such as Mixtral 8×7B \cite{jiang2024mixtral}. During inference, these models activate only a subset of parameters, thereby improving efficiency. However, the training phase necessitates updating all parameters, which results in substantially higher FLOPs. Similarly, models like QuantGPT \cite{tao2022compression} created via QAT might be significantly smaller in size and but the training process is both computationally as well as memory intensive due to maintaining and updating additional full-precision latent weights. Moreover, training these models via KD increases the overall training FLOPs, since each iteration requires an additional forward pass through the large teacher LLM. Furthermore, the cost of production of hardware and infrastructure for deployment of these models is often unaccounted for. A study by Gupta et al. \cite{gupta2021chasing} reveals that the environmental impact due to setting up infrastructure and manufacturing hardware equipment is maximum compared to other life-cycle stages.

\section{RQ6: Results and Discussion}
\label{rr}
In this section, a multi-faceted statistical analysis of the surveyed articles has been presented. This is followed by an evaluation of the performance of the SOTA NLP models concerning their efficiency to depict the trend of research. Furthermore, the implications as well as the limitations of this survey and the threats to validity have been studied.

\subsection{Statistical Insights}
\label{si}

\subsubsection{Year-wise Distribution of Articles}
\label{yda}
\begin{figure}[t!]
\centering
\includegraphics[width=0.6\linewidth]{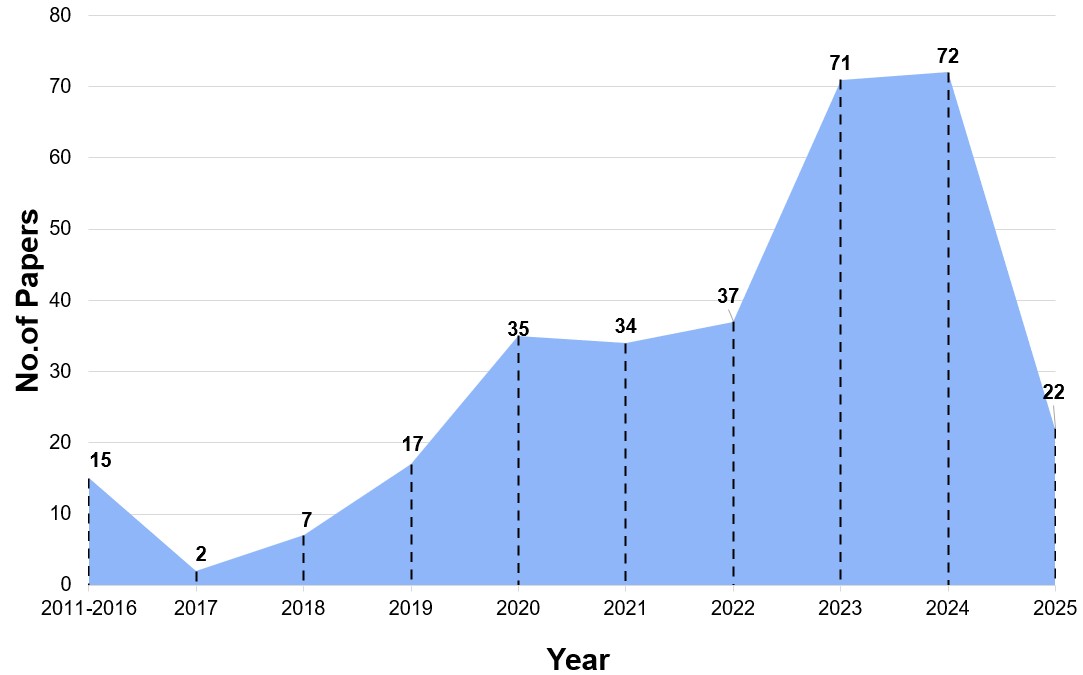}
\caption{Year-wise distribution of papers}
\label{fig8}       
\end{figure}

Figure \ref{fig8} presents the year-wise distribution of surveyed articles. It reflects the rise in interest in Transformer-based LM after the seminal paper by Vaswani et al. \cite{vaswani2017attention} in the year 2017. It can be perceived that the major share of articles is from the last five years. The maximum number of articles is between 2022 and 2024. This indicates a substantial increase in research activity in recent years, driven by the growing interest in Transformer‑based LLMs. Contrary to our expectations of a rising trend, in 2025 the number of articles has declined. This can be attributed to the fact that to assess the quality of articles, the number of citations has been considered to be one of the important indicators in our survey. However, considering the delays associated with publication workflows and the gradual process through which an article gains recognition, it is challenging for recently published papers to attract a significant number of citations within a short time frame.

\begin{figure*}[h!]
\centering
\includegraphics[width=\linewidth]{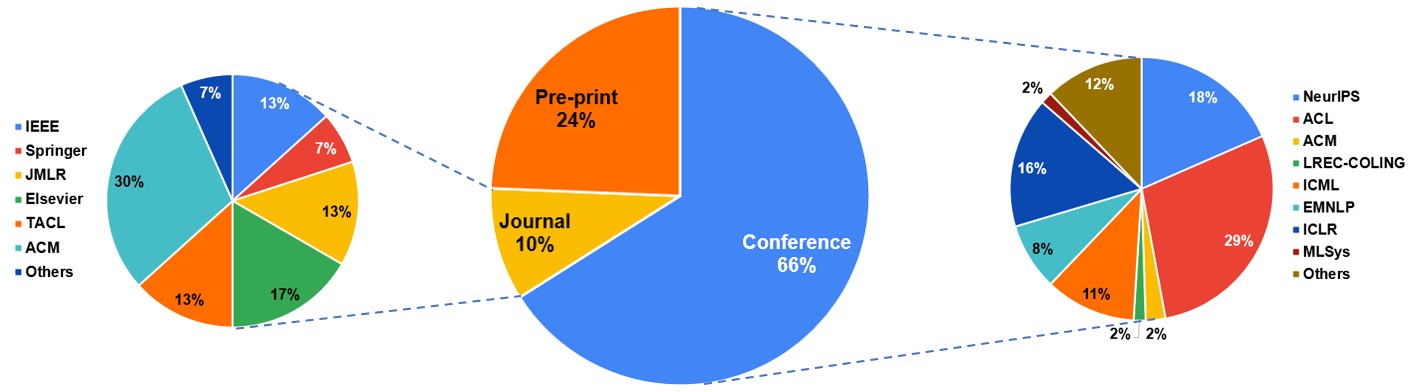}
\caption{Percentage share of article-types}
\label{fig9}       
\end{figure*}

\subsubsection{Share of Article Types}
\label{sat}
Figure \ref{fig9} illustrates the share of article types in this survey. The central pie chart reflects that conferences account for a dominant 66\% of articles while journals have only 10\% share. This can be attributed to the fact that NLP researchers often favor prestigious conferences over journals, as fixed timelines of conferences enable faster dissemination compared to the lengthy, multi‑stage review cycles of journal publications in this rapidly evolving field. Among the conferences, ACL (29\%) and NeurIPS (18\%) together account for 47\% of all conference publications. This is followed by ICLR (16\%), ICML (11\%) EMNLP (8\%) and LREC-COLING (2\%). These conferences being ranked CORE A* for NLP, become the venue for many seminal papers. Among the journal articles, ACM (30\%) and Elsevier (17\%) stand out as major publishers of high impact works. JMLR, IEEE and TACL have equal share of 13\%, while Springer holds around 7\% share of articles. Interestingly, 24\% share of articles is from pre-print platforms like arXiv\footnote{https://arxiv.org/}. A closer examination reveals that many pre‑prints are noteworthy contributions from leading researchers affiliated with reputed institutions. Offering visibility within a few days by bypassing the lengthy review and publication timelines of journals and conferences, it has become an attractive venue for establishing authorship. Consequently, it has become common practice to upload work to pre‑print platforms prior to formal publication, with many of the latest advances now appearing first as pre‑prints rather than in traditional venues.

\begin{figure*}[h!]
    \centering
    \begin{subfigure}[h]{0.45\textwidth}
        \centering
        \includegraphics[width=\linewidth]{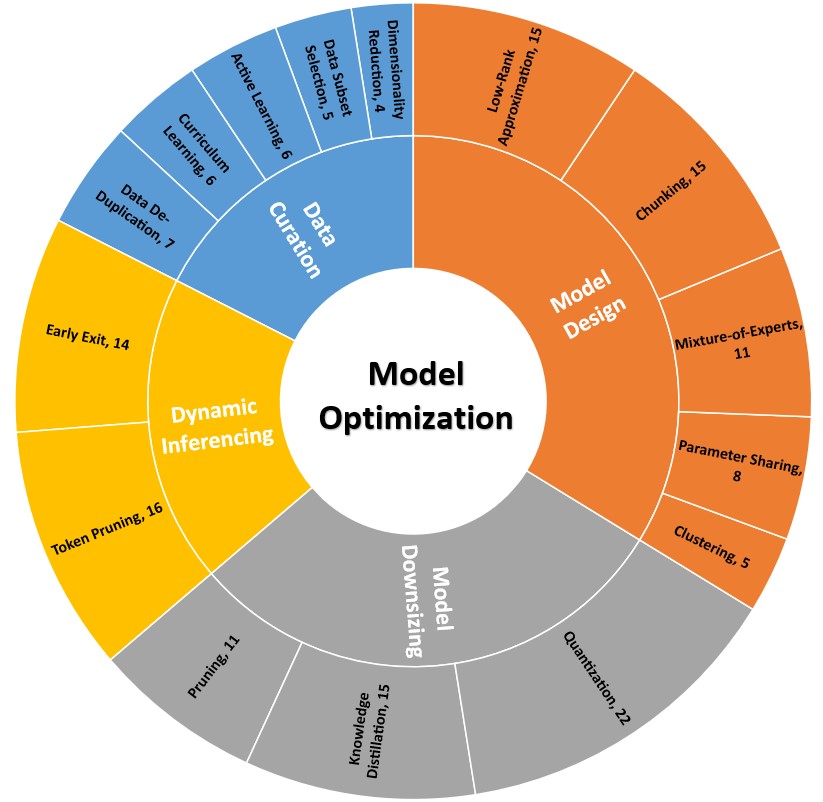}
        \caption{Model optimization}
        \label{fig11a}       
    \end{subfigure}%
    ~
    \begin{subfigure}[h]{0.45\textwidth}
        \centering
        \includegraphics[width=\linewidth]{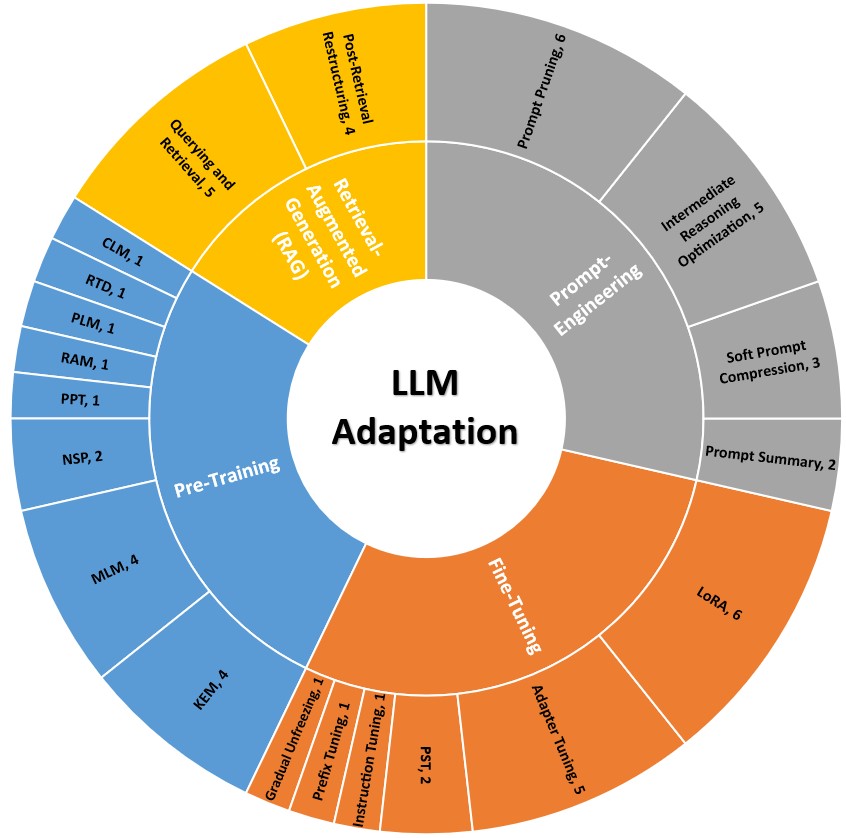}
        \caption{LLM adaptation}
        \label{fig11b}       
    \end{subfigure}

    \caption{Category-wise distribution of articles}
    \label{fig11}
    
\end{figure*}

\subsubsection{Category-wise Distribution of Articles}
\label{cda}
For a more detailed analysis, the distribution of articles related to both model optimization and LLM adaptation has been performed. For model adaptation, the analysis of the sunburst chart in Figure \ref{fig11a} reveals a clear prioritization of architectural efficiency and compression. Model design (54 articles) and model downsizing (48 articles) are the dominant drivers, with quantization (22 articles) emerging as the leading individual technique for reducing model footprint. Dynamic inferencing (30 articles) serves as a significant secondary pillar, focusing purely on runtime speed via token pruning and early exit. Furthermore, data curation (28 articles) remains a niche area, characterized by a fragmented spread of techniques like data de-duplication and curriculum learning, suggesting that current optimization research is more heavily weighted toward the model itself rather than the data it consumes.

The analysis of the sunburst chart in Figure \ref{fig11b} reveals a competitive and balanced landscape in LLM adaptation. The distribution is led by a tie between fine-tuning and prompt-engineering (16 articles each), closely followed by pre-training (15 articles), with RAG (9 articles) representing a smaller but emerging segment. Within the dominant categories, trends favor efficient practices. For instance, fine-tuning heavily relies on parameter-efficient methods like LoRA (6 articles) and adapter tuning (5 articles), while prompt-engineering prioritizes token reduction via prompt pruning (6 articles) and intermediate reasoning optimization (5 articles). Conversely, pre-training exhibits the most fragmentation across eight diverse sub-techniques such as MLM (4 articles) and KEM (4 articles), suggesting a wide experimental breadth in foundational model updates compared to the more consolidated approaches in fine-tuning and prompting.

\begin{table*}[h!]
  \begin{center}
    \caption{Comparison of Performance of Popular LLMs}
    \label{table:table1}
    \setlength{\tabcolsep}{0.3em} 
    \renewcommand{\arraystretch}{1} 
 \rotatebox{0}{ \resizebox{\textwidth}{!}{
 \begin{tabular}{p{2.4cm} p{1cm}p{1cm}p{1cm}p{1cm}p{1cm}p{1cm}p{1.2cm}p{1cm}p{1cm}p{1cm}p{1cm}p{1cm}p{1cm}} 
 \hline
Model & Arena Elo & AAII & TB Hard & AA-LCR  & HLE & MMLU-Pro & GPQA Diamond & LCB & SciCode & IFBench & AIME 2025 & MATH 500 & Human-Eval \\
\hline  
Claude 4 Sonnet & 1335 & 57 & 30\% & 65\% & 10\% & 84\% & 78\% & 66\% & 40\% & 55\% & 74\% & 99\% & - \\ 
\hline 
Claude 4.1 Opus & 1419 & 59 & 32\% & 66\% & 12\% & 88\% & 81\% & 65\% & 41\% & 55\% & 80\% & - & - \\ 
\hline 
Claude 4.5 Sonnet & 1420 & 63 & 33\% & 66\% & 17\% & 88\% & 83\% & 71\% & 45\% & 57\% & 88\% & - & - \\ 
\hline DeepSeek-r1 \cite{guo2025deepseek} & 1426 & 52 & 15\% & 55\% & 15\% & 85\% & 81\% & 77\% & 40\% & 40\% & 76\% & 98\% & 97\% \\ 
\hline DeepSeek V3.1 Terminus \cite{liu2024deepseek3} & 1420 & 58 & 28\% & 65\% & 15\% & 85\% & 79\% & 80\% & 41\% & 57\% & 90\% & - & - \\ 
\hline Gemini 2.5 Flash \cite{comanici2025gemini} & 1412 & 54 & 16\% & 64\% & 13\% & 84\% & 79\% & 71\% & 41\% & 52\% & 78\% & - & - \\ 
\hline Gemini 2.5 Pro \cite{comanici2025gemini} & 1466 & 60 & 25\% & 66\% & 21\% & 86\% & 84\% & 80\% & 43\% & 49\% & 88\% & 97\% & - \\ 
\hline 
Gemma 3 12B \cite{team2025gemma} & 1335 & 20 & 1\% & 7\% & 5\% & 60\% & 35\% & 14\% & 17\% & 37\% & 18\% & 85\% & 83\% \\ 
\hline 
Gemma 3 27B \cite{team2025gemma} & 1357 & 22 & 4\% & 0\% & 5\% & 67\% & 43\% & 14\% & 21\% & 32\% & 21\% & 88\% & 89\% \\ 
\hline 
Gemma 3 4B \cite{team2025gemma} & 1293 & 15 & 1\% & 6\% & 5\% & 42\% & 29\% & 11\% & 7\% & 28\% & 13\% & 77\% & 72\% \\ 
\hline 
GLM-4.6 & 1442 & 56 & 23\% & 54\% & 13\% & 83\% & 78\% & 70\% & 38\% & 43\% & 86\% & - & - \\ 
\hline 
GPT-5 & 1443 & 68 & 31\% & 76\% & 27\% & 87\% & 85\% & 85\% & 43\% & 73\% & 94\% & 99\% & 99\% \\ 
\hline 
GPT-5 mini & 1375 & 64 & 31\% & 68\% & 20\% & 84\% & 83\% & 84\% & 39\% & 75\% & 91\% & - & - \\ 
\hline 
GPT-5 nano (high) & 1333 & 51 & 11\% & 42\% & 8\% & 78\% & 68\% & 79\% & 37\% & 68\% & 84\% & - & - \\ 
\hline 
Grok 4 & 1446 & 65 & 38\% & 68\% & 24\% & 87\% & 88\% & 82\% & 46\% & 54\% & 93\% & 99\% & 98\% \\ 
\hline 
Grok 4 Fast & 1419 & 60 & 18\% & 65\% & 17\% & 85\% & 85\% & 83\% & 44\% & 51\% & 90\% & - & - \\ 
\hline 
Kimi K2 0905 & 1382 & 50 & 23\% & 52\% & 6\% & 82\% & 77\% & 61\% & 31\% & 42\% & 57\% & - & - \\ 
\hline 
Llama 3.3 70B \cite{dubey2024llama} & 1276 & 28 & 3\% & 15\% & 4\% & 71\% & 50\% & 29\% & 26\% & 47\% & 8\% & 77\% & 86\% \\ 
\hline 
Llama 3.3 Nemotron Super 49B v1.5 \cite{dubey2024llama} & 1340 & 45 & 5\% & 34\% & 7\% & 81\% & 75\% & 74\% & 35\% & 37\% & 77\% & 98\% & 95\% \\ 
\hline 
Llama 4 Maverick & 1292 & 36 & 6\% & 46\% & 5\% & 81\% & 67\% & 40\% & 33\% & 43\% & 19\% & 89\% & 88\% \\ 
\hline 
Llama 4 Scout & 1276 & 28 & 1\% & 26\% & 4\% & 75\% & 59\% & 30\% & 17\% & 40\% & 14\% & 84\% & 83\% \\ 
\hline 
Magistral Medium 1.2 \cite{rastogi2025magistral} & 1253 & 52 & 13\% & 51\% & 10\% & 82\% & 74\% & 75\% & 39\% & 43\% & 82\% & - & - \\ 
\hline 
Mistral Medium 3.1 & 1402 & 35 & 10\% & 20\% & 4\% & 68\% & 59\% & 41\% & 34\% & 40\% & 38\% & - & - \\ 
\hline 
Mistral Small 3.2 & 1337 & 29 & 6\% & 17\% & 4\% & 68\% & 51\% & 28\% & 26\% & 34\% & 27\% & 88\% & 85\% \\ 
\hline 
Mixtral 8x7B \cite{jiang2024mixtral} & 1138 & 3 & - & - & 5\% & 39\% & 29\% & 7\% & 3\% & - & - & 30\% & 1\% \\ 
\hline 
o3 \cite{el2025competitive} & 1441 & 65 & 35\% & 69\% & 20\% & 85\% & 83\% & 81\% & 41\% & 71\% & 88\% & 99\% & 99\% \\ 
\hline 
Phi-4 \cite{abdin2024phi} & 1222 & 25 & 4\% & 0\% & 4\% & 71\% & 57\% & 23\% & 26\% & 24\% & 18\% & 81\% & 87\% \\ 
\hline 
Qwen3 235B \cite{yang2025qwen3} & 1418 & 57 & 13\% & 67\% & 15\% & 84\% & 79\% & 79\% & 42\% & 51\% & 91\% & 98\% & 98\% \\ 
\hline 
Qwen3 Max \cite{yang2025qwen3} & 1440 & 55 & 19\% & 47\% & 11\% & 84\% & 76\% & 77\% & 38\% & 44\% & 81\% & - & - \\ 
\hline 
Qwen3 Next 80B \cite{yang2025qwen3} & 1417 & 54 & 9\% & 60\% & 12\% & 82\% & 76\% & 78\% & 39\% & 61\% & 84\% & - & - \\ 

\hline 
\end{tabular}}}
\end{center}
\end{table*}

\subsection{Performance and Efficiency Analysis}
\label{pea}

\subsubsection{Comparison of Performance}
\label{cp}
Table \ref{table:table1} compares the performance of popular LLMs on 13 different benchmarks to assess their capabilities on a wide range of tasks. The comparison of Arena Elo depicts a highly saturated frontier with marginal variation in performance. Gemini 2.5 Pro establishes a distinct lead with an Arena Elo of 1466. However, the immediate contenders comprising Grok 4, GPT-5, GLM-4.6, and o3 lie within a narrow $\approx$25-point band, indicating functional parity for most real-world applications. The table highlights a paradigm shift toward high-performance lightweight LLMs, with efficiency-optimized variants like Gemini 2.5 Flash and Grok 4 Fast outscoring flagships like Claude 4 Sonnet, thereby validating the industry's pivot toward cost-effective intelligence. Furthermore, the presence of DeepSeek, GLM, and Qwen in the top tier confirms a multi-polar market where international open-weights and proprietary LLMs compete on an equal footing.

A comparison of the AAII scores highlights aconcentration in frontier model performance. GPT‑5 secures the top position with an index of 68, while Grok 4 and o3 are tied at 65. A significant observation is the exceptional performance of distilled architectures. For instance, GPT-5 mini achieves a score of 64, outperforming flagship LLMs like Claude 4.5 Sonnet (63) and Gemini 2.5 Pro (60). Besides, the table underscores a capability gap for open-weights LLMs, as the Gemma 3 series and standard Llama variants generally score significantly lower (mostly below 40).

A comparison of the average performance across all other benchmarks reveals the domination by the latest proprietary frontiers, with GPT-5 achieving a commanding average performance of 72.6\%. It is closely trailed by Grok 4 (70.6\%) and o3 (70.1\%). This indicates that the reasoning capabilities of LLMs have reached unforeseen heights. In addition, a distinct Pareto improvement in both performance and efficiency is observed, notably demonstrating the power of modern distillation techniques. For instance, GPT-5 mini (63.9\%) not only rivals but surpasses established heavyweight LLMs like Claude 4 Sonnet (60.1\%) and DeepSeek V3.1 (60.0\%). In contrast, the table reveals a marked performance drop among smaller or open‑weight LLMs, with Gemma 3 and Llama 3.3 variants typically scoring below 40.0\%. This underscores that, while open‑weight LLMs continue to improve, they remain far from the reasoning capabilities of the high‑compute frontier LLMs.

\subsubsection{Analysis of Efficiency during Inference}
\label{aei}

Table \ref{table:table6} provides a granular comparison of the operational characteristics of popular LLMs during inference considering factors such as accessibility, cost, and speed. Observing the providers of LLMs, it becomes evident that most of the widely adopted LLMs are either provided or supported by a small group of high-capital organizations with net worth in the trillions of dollars, such as Google, Meta, and Microsoft. It underscores a disparity in the opportunity for research as the creation of SOTA LLMs demands immense financial investment and computational infrastructure. This limits the participation of smaller institutions and independent researchers.

\begin{table*}[h!]
  \begin{center}
    \caption{Comparison of Technical Details and Efficiency of Popular LLMs}
    \label{table:table6}
    
 \rotatebox{0}{ \resizebox{\textwidth}{!}{  
 \begin{tabular}{p{2.7cm} p{1.5cm} p{1.5cm} p{1.5cm} p{1.5cm} p{1.5cm} p{1.5cm} p{1.5cm} p{1.5cm}} 
 \hline
Model & Provider & License & Context Window & Pricing- Input & Pricing- Output & Median Output Tokens/s & Latency (s) & E2E RT (s) \\
\hline  
Claude 4 Sonnet & Anthropic & Proprietary & 1M & \$3.00 & \$15.00 & 53.2 & 38.49 & 47.88 \\ 
\hline 
Claude 4.1 Opus & Anthropic & Proprietary & 200K & \$15.00 & \$75.00 & 42.1 & 49.02 & 60.89 \\ 
\hline 
Claude 4.5 Sonnet & Anthropic & Proprietary & 1M & \$3.00 & \$15.00 & 59.9 & 35.1 & 43.44 \\ 
\hline 
DeepSeek-r1 \cite{guo2025deepseek} 0528 & DeepSeek & Open & 128K & \$0.80 & \$3.00 & 0 & 0 & 0 \\ 
\hline 
DeepSeek V3.1 Terminus \cite{liu2024deepseek3} & DeepSeek & Open & 128K & \$1.64 & \$2.75 & 0 & 0 & 0 \\ 
\hline 
Gemini 2.5 Flash \cite{comanici2025gemini} & Google & Proprietary & 1M & \$0.30 & \$2.50 & 260.4 & 8.61 & 10.53 \\ 
\hline 
Gemini 2.5 Pro \cite{comanici2025gemini} & Google & Proprietary & 1M & \$1.25 & \$10.00 & 150.1 & 30.35 & 33.68 \\ 
\hline 
Gemma 3 12B \cite{team2025gemma} & Google & Open & 128K & \$0.00 & \$0.00 & 52.8 & 1.37 & 10.84 \\ 
\hline 
Gemma 3 27B \cite{team2025gemma} & Google & Open & 128K & \$0.00 & \$0.00 & 52.4 & 0.59 & 10.13 \\ 
\hline 
Gemma 3 4B \cite{team2025gemma} & Google & Open & 128K & \$0.00 & \$0.00 & 51.2 & 0.98 & 10.74 \\ 
\hline 
GLM-4.6 & Z.ai & Open & 200K & \$0.60 & \$2.05 & 45.9 & 44.29 & 55.17 \\ 
\hline 
GPT-5 & OpenAI & Proprietary & 400K & \$1.25 & \$10.00 & 159 & 66.94 & 70.08 \\ 
\hline 
GPT-5 mini & OpenAI & Proprietary & 400K & \$0.25 & \$2.00 & 73.8 & 94.29 & 101.07 \\ 
\hline 
GPT-5 nano (high) & OpenAI & Proprietary & 400K & \$0.05 & \$0.40 & 117.9 & 71.44 & 75.69 \\ 
\hline 
Grok 4 & xAI & Proprietary & 256K & \$3.00 & \$15.00 & 42.3 & 13.79 & 25.62 \\ 
\hline 
Grok 4 Fast & xAI & Proprietary & 2M & \$0.20 & \$0.50 & 197.8 & 3.64 & 6.16 \\ 
\hline 
Kimi K2 0905 & Kimi & Open & 256K & \$0.99 & \$2.75 & 56.2 & 0.47 & 9.36 \\ 
\hline 
Llama 3.3 70B \cite{dubey2024llama} & Meta & Open & 128K & \$0.54 & \$0.68 & 92.6 & 0.45 & 5.84 \\ 
\hline 
Llama 3.3 Nemotron Super 49B v1.5 \cite{dubey2024llama} & Meta & Open & 128K & \$0.10 & \$0.40 & 79.7 & 25.64 & 31.92 \\ 
\hline 
Llama 4 Maverick & Meta & Open & 1M & \$0.24 & \$0.85 & 142.7 & 0.37 & 3.88 \\ 
\hline 
Llama 4 Scout & Meta & Open & 10M & \$0.15 & \$0.59 & 114.3 & 0.51 & 4.89 \\ 
\hline 
Magistral Medium 1.2 \cite{rastogi2025magistral} & Mistral AI & Proprietary & 128K & \$2.00 & \$5.00 & 147.5 & 13.92 & 17.31 \\ 
\hline 
Mistral Medium 3.1 & Mistral AI & Proprietary & 128K & \$0.40 & \$2.00 & 96.5 & 0.37 & 5.55 \\ 
\hline 
Mistral Small 3.2 & Mistral AI & Open & 128K & \$0.10 & \$0.30 & 123.5 & 0.28 & 4.32 \\ 
\hline 
Mixtral 8x7B \cite{jiang2024mixtral} & Mistral AI & Open & 33K & \$0.70 & \$0.70 & 61 & 0.35 & 8.51 \\ 
\hline 
o3 \cite{el2025competitive} & OpenAI & Proprietary & 200K & \$2.00 & \$8.00 & 216.2 & 12.29 & 14.61 \\ 
\hline 
Phi-4 \cite{abdin2024phi} & Microsoft & Open & 16K & \$0.13 & \$0.50 & 36.3 & 0.41 & 14.2 \\ 
\hline 
Qwen3 235B \cite{yang2025qwen3} & Alibaba & Open & 256K & \$0.70 & \$8.40 & 83.7 & 25 & 30.98 \\ 
\hline 
Qwen3 Max \cite{yang2025qwen3}& Alibaba & Proprietary & 262K & \$1.20 & \$6.00 & 36.8 & 1.63 & 15.22 \\ 
\hline 
Qwen3 Next 80B \cite{yang2025qwen3} & Alibaba & Open & 262K & \$0.50 & \$6.00 & 182.2 & 12.04 & 14.78 \\
\hline 
 \multicolumn{9}{p{14.4cm}}{Note: Pricing is calculated in terms of USD per 1M Tokens\newline
 Latency is calculated as time to receive the first answer token \newline
 E2E RT indicates End-to-End Response Time}
\end{tabular}}}
\end{center}
\end{table*}

The license-types show a bifurcated market structure. Proprietary LLMs from established companies such as Anthropic, Google, and OpenAI dominate the high-performance tier. In contrast, the table reveals an emerging ecosystem of "Open" license LLMs, primarily driven by Meta, Mistral AI, and DeepSeek. The increased availability of open-weight LLMs with enterprise-grade capabilities accelerates innovation by  researchers and smaller organizations while significantly reducing operational costs.

Analyzing the amount of context supported, it can be said that the context windows of LLMs have largely stabilized at 128k tokens (Gemma 3, Mistral Small), establishing this as the new baseline for "short" context. However, a distinct "long-context" tier has emerged, led by Gemini 2.5 Flash and Llama 4 Maverick with 1 million token windows, and notably Grok 4 Fast pushing to 2 million. Llama 4 Scout serves as an exception, offering a massive 10 million token window. This indicates a paradigm-shift in architectural design to support massive data.

\begin{figure*}[h]
\centering
\includegraphics[width=\linewidth]{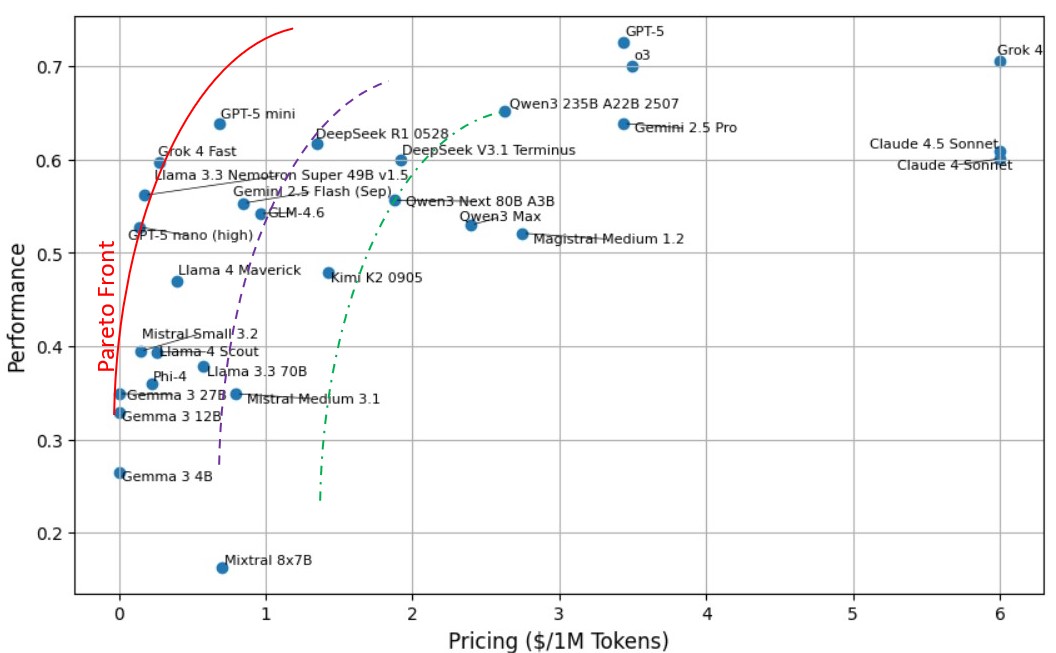}
\caption{Analysis of the price-performance tradeoff of popular LLMs}
\label{fig16}       
\end{figure*}

The pricing dynamics for input and output vary wildly, reflecting the "premium for intelligence" of different LLMs. The most expensive model listed is Claude 4.1 Opus, commanding \$15.00 per million input tokens and a staggering \$75.00 per million output tokens. This pricing signals a target market of high-value, low-volume reasoning tasks where accuracy is paramount. In contrast, efficiency-focused LLMs like Gemini 2.5 Flash (\$0.30 input / \$2.50 output) and Grok 4 Fast (\$0.20 input / \$0.50 output) offer orders-of-magnitude lower costs, democratizing access to capable intelligence for high-volume applications. Interestingly, GPT-5 adopts a symmetric pricing model (\$10.00 for both input and output), a departure from the standard industry practice of discounting input tokens.

The throughput (Median Output Tokens/s) metrics highlight the divide between "reasoning" and "general-purpose" LLMs. Gemini 2.5 Flash leads with an impressive 260.4 tokens/s, followed by o3 at 216.2 tokens/s serving as the optimal choice for real-time applications like voice agents. Conversely, heavier LLMs like Claude 4.1 Opus (42.1 tokens/s) and Llama 3.3 70B (56.2 tokens/s) are significantly slower. This makes them suitable for applications where longer wait times are acceptable in exchange for higher quality outputs.

The latency values show extreme variance based on model size and optimization. Several open-weight and smaller LLMs have minimal latency, with Llama 4 Maverick (0.37s), Mistral Small 3.2 (0.28s), and Gemma 3 27B (0.59s) delivering near-instantaneous responses. On the other end of the spectrum, GPT-5 exhibits a high latency of 66.94s and GPT-5 mini requires 94.29s. These unusually high figures indicate deployment of extensive reasoning mechanisms before outputting the first visible character to the user.

The End-to-End Response Time (E2E RT) generally correlates with the complexity of the task and the LLM's generation speed. While lighter LLMs like Grok 4 Fast complete generations in just 6.16s, the heavy reasoning LLMs require significantly more time, with GPT-5 mini taking over 100 seconds to complete a response. E2E RT plays a pivotal role in determining the application LLMs. For instance, the minute-long generation time of Claude 4.1 Opus limits its utility in interactive chat interfaces compared to LLMs having E2E RT around 10 seconds.

\begin{table*}[h!]
  \begin{center}
    \caption{Comparative Analysis of Computational Resources, Energy Consumption, and Carbon Emissions for Pre-Training LLMs}
    \label{table:table3}
 \rotatebox{0}{ \resizebox{\textwidth}{!}{    
 \begin{tabular}{lllllll}
 \hline
 Model & \#Parameters & \#Tokens & GPU/TPU & GPU/TPU Hours & Energy & Emissions \\  
 \hline
Transformer\_base \cite{vaswani2017attention} & 65M & $\approx$2.43B & GPU-P100 & 96 & 1.416 & 0.0117  \\  \hline   
Transformer\_big \cite{vaswani2017attention} & 213M & $\approx$2.43B & GPU-P100 & 672 & 1.515 & 0.0864  \\  \hline   
ELMo \cite{peters-etal-2018-deep} & 93M & 5.5B & GPU-P100 x 3 & 1008 & 0.51766 & 0.118  \\  \hline   
BERT\_base \cite{kenton2019bert} & 110M & $\approx$3.3B & TPU-v3 & 384 & $\approx$50 & 5  \\  \hline   
BERT\_large \cite{kenton2019bert} & 340M & $\approx$3.3B & TPU-v3 & 1536 & $\approx$200 & 20  \\  \hline   
GPT-1 \cite{radford2018improving} & 117M & $\approx$4.5B & GPU-V100 & - & $\approx$30 & $\approx$25  \\  \hline   
GPT-2 \cite{radford2019language} & 1.5B & $\approx$8B & TPU-v3 & 5376 & - & -  \\  \hline   
GPT-3 \cite{brown2020language} & 175B & 300B & GPU-V100 & 3.5M & 1,287 & 552.1  \\  \hline  
GPT-4 \cite{achiam2023gpt} & 1.8T & $\approx$13T & A100 & 54M & $\approx$10000 & $\approx$2000  \\  \hline   
GPT-5 & 10T-50T & - & GPU-H100 & $\approx$108M & - & -  \\  \hline    
Gopher \cite{rae2021scaling} & 280B & $\approx$300B & TPU-v4 & $\approx$3M & $\approx$1200 & $\approx$450  \\  \hline   
GShard \cite{lepikhingshard} & 600B & - & TPU-v3 & 76K & 24.1 & 4.8  \\  \hline   
LLaMA-7B \cite{touvron2023llama} & 7B & 1.0T & GPU-A100   & 82K & 36 & 14  \\  \hline   
LLaMA-13B \cite{touvron2023llama} & 13B & 1.0T & GPU-A100   & 135K & 59 & 23  \\  \hline   
LLaMA-33B \cite{touvron2023llama} & 33B & 1.4T & GPU-A100   & 530K & 233 & 90  \\  \hline   
LLaMA-65B \cite{touvron2023llama} & 65B & 1.4T & GPU-A100   & 1.0M & 449 & 173  \\  \hline   
T5 \cite{raffel2020exploring} & 11B & 34B & TPU-v3 & 245K & 85.7 & 46.7  \\  \hline   
XLM \cite{conneau2019cross} & 10.7B & 6.3B & GPU-V100 & 250K & 167.4 & 39  \\  \hline   
Switch Transformer \cite{fedus2022switch} & 1.6T & $\approx$32T & TPU-v3 & 663K & 179 & 72.2  \\  \hline  
OPT \cite{zhang2022opt} & 175B & 180B & GPU-A100  & 809K & 356 & 137  \\  \hline   
BLOOM \cite{workshop2022bloom} & 175B & $\approx$366B & GPU-A100  & 1.1M & 475 & 183  \\  \hline   
LaMDA \cite{thoppilan2022lamda} & 137B & - & TPU-V3 S & 1.4M & 451 & 25.2  \\  \hline   
PaLM \cite{chowdhery2023palm} & 540B & 780B & TPU-v4 & 2.5M & $\approx$3500 & $\approx$1000  \\  \hline   
Chinchilla \cite{hoffmann2022training} & 70B & $\approx$1.4T & TPU-v4 & $\approx$1.2M & $\approx$600 & $\approx$250  \\  
    \hline
 \multicolumn{7}{p{0.9\linewidth}}{Note: Emissions- Co2 emitted (metric tons), Energy- Power Consumption (MWh), \#Parameters- Number of parameters, \#Tokens- Number of training tokens}
\end{tabular}}}
\end{center}
\end{table*}

\subsubsection{Analysis of the Price-Performance Tradeoff}
\label{ppt}
The Pareto chart in Figure \ref{fig16} compares LLMs by performance and pricing\footnote{Pricing is calculated as a blend of input and output pricing per million tokens taken in a 3:1 ratio.}. It highlights a Pareto Front where the most efficient models such as GPT‑5 mini, Grok 4 Fast, and GPT‑5 nano lie on the Pareto front (marked by a red curve). Just behind them are near‑optimal contenders like Llama 3.3 Nemotron Super 49B v1.5 and Gemini 2.5 Flash. The landscape appears to be divided into three tiers as follows. The top-tier comprises ultra‑high‑performance models like GPT‑5, Grok 4, and Claude 4.5 Sonnet having steep prices. The mid‑range models such as Gemini 2.5 Pro and Qwen3 235B balance cost and capability. Whereas, the last tier consists of budget‑friendly small models like Llama 4 Maverick, Mistral‑Small, and Phi‑4 offering low‑cost, low‑latency options. Models far from the Pareto front are inefficient, offering worse performance for similar prices, while pricing clusters reveal most models fall below \$4 per million tokens, with a few high‑cost outliers serving specialized needs. Overall, the chart underscores the trade‑offs between performance and economic efficiency when choosing an LLM.

\subsubsection{Analysis of Efficiency during Pre-Training}
\label{aep}
Table \ref{table:table3} illustrates the dramatic rise in computational requirements with the evolution of LLMs. It accounts for the number of model parameters, tokens used, and training time for pre-training popular LLMs. It also accounts for sustainability factors like power consumption along with environmental impact due to carbon emissions. As the LLMs have scaled from the 65M parameter Transformer\_base to the massive 1.8T parameter GPT-4, pre-training demands have surged from less than 100 GPU hours on legacy P100 hardware to tens of millions of hours on modern A100 and H100 clusters. This exponential growth in parameters and training tokens correlates directly with energy consumption. Early models like BERT\_large resulted in negligible emissions (5 tons $CO_2$), while recent LLMs like PaLM and GPT-4 are estimated to generate thousands of tons of $CO_2$ ($\approx$1,000 and $\approx$2,000 tons, respectively). Furthermore, the table highlights a shift in training strategies exemplified by Chinchilla and the LLaMA versions where high token counts on high-performance hardware (TPU-v4, A100) drive significant energy costs even for LLMs with fewer parameters. This reflects that the pursuit of enhancing model capability currently necessitates massive energy requirements, resulting in huge carbon emissions despite improvements in hardware efficiency.

\subsection{Trend of Research}
\label{tr}
It is evident that Transformer-based LLMs have achieved new heights in the recent years. Modern LLMs demonstrate remarkable improvements in reasoning capabilities and in handling extended contexts. However, their immense resource requirements increase carbon emissions and energy consumption, limiting research in resource-constrained environments. This can be visualized from Table \ref{table:table3} showing that training the SOTA NLP models can lead to several tons of carbon emissions. It can be observed that there has been an overall rising trend in the size of the pre-training data as well as the parameter count. The initial developments incrementally raised the model parameters and pre-training corpus obeying a linear relationship among them. Subsequent LLMs focused majorly on increasing the model parameters without much rise in the size of the pre-training data. The recent LLMs are focusing on striking a balance between the size of data and the model size if not reducing the model size compared to the volume of pre-training data. Such efforts have exhibited improved efficiency without compromising efficacy. Moreover, there is a growing inclination towards open‑weight models. The DeepSeek, Gemma, LLaMA and Mistral family of LLMs have demonstrated appreciable performance on most general-purpose tasks (as seen in Table \ref{table:table1}), while offering added advantages of reduced training costs and the flexibility to experiment with the model for further advancement. This signals a ray of hope, suggesting that awareness of efficient and low‑cost LLMs is proliferating steadily within the NLP research community.

\subsection{Discussion}
\label{dis}

\subsubsection{Implications of this Survey}
\label{ios}
This paper presents a commentary on Transformer-based LLMs in NLP along with discussion on aspects like computational requirements, scaling laws, carbon emissions and energy usage of LLMs. Following this, the research contributions on enhancing the efficiency of Transformer-based LLM at various stages of model development and LLM adaptation have been discussed. Furthermore, performance comparison of popular LLMs on various benchmarks together with multifaceted efficiency estimation highlight the current research scenario and also predict the future trend. This paper aims at researchers and practitioners of NLP, AI, as well as scholars who desire to develop an in-depth understanding of Transformers and LLMs, along with their computational overheads, and efficiency considerations. The goal of this survey is to determine how current the NLP landscape contributes towards a sustainable society and establish a foundation for future research.

\subsubsection{Limitations of this Survey}
\label{lim}
Despite curating the research works with meticulous care and making this paper as comprehensive as possible, certain limitations persist. There might be certain biases in article selection. Due to English being the most representative language, the study has been limited to articles published in the English language. As the articles were retrieved through keyword-matching, some relevant articles might have been omitted due to the absence of those keywords. Besides, the survey covers only journal, conference, and pre-print articles, excluding other article types like thesis, reports, etc. However, none of these limitations diminish the significance of this survey. Furthermore, the limitations herald possibilities for expanding this work in the future.

\subsubsection{Threats to Validity}
\label{ttv}
To minimize the threats to validity, certain measures have been undertaken. To ensure a methodical review of the articles, a survey strictly adhering to the research questions has been adopted. The papers have been obtained from all reputable venues of publications to maximize the scope of the survey. The keywords and selection criteria for the articles have been finalized after thorough deliberation among all the authors to ensure maximal coverage of high-quality relevant articles. To further eliminate biases and ensure optimal organization of the findings, several brainstorming sessions were conducted among the authors.

\section{Conclusion and Future Avenues}
\label{concl}

This survey underscores the pivotal role of efficiency in the evolution of Transformer‑based LLMs. Despite remarkable advances in the performance and capabilities of Transformer‑based LLMs in recent years, their escalating resource demands present critical challenges, particularly with respect to carbon emissions, energy consumption, and monetary costs. The rising trend in model parameters and volume of data highlights the urgency of striking a balance between scale and sustainability. By synthesizing insights from 324 studies, it can be inferred that enhancing efficiency of LLM is inclusive of all stages of model development such as data curation, model design, model downsizing and dynamic inferencing. Moreover, it depends on the LLM adaptation strategies like pre-training, fine-tuning, prompt-engineering, and RAG. Consolidating these approaches is essential to reduce monetary costs, energy consumption, and carbon emissions to ensure sustainable progress.

Encouragingly, the recent advances demonstrate that efficiency can be enhanced without compromising efficacy, especially through innovations that align model size with the volume of data. Furthermore, the growing adoption of open‑weight models such as DeepSeek, Gemma, LLaMA, and Mistral reflects a promising shift toward democratizing access, reducing costs, and fostering experimentation in resource‑constrained environments. This trajectory signals increasing awareness within the NLP community of the need for low‑cost, environmentally responsible LLMs. Future research must prioritize the efficiency of LLMs throughout their life cycle also accounting with the environmental footprint of the infrastructure involved. This would ensure that the advancements in the given domain remains aligned with global sustainability objectives.

\bibliographystyle{unsrt}  
\bibliography{sample-base}

@string{BIT = "{BIT}" }

@string{Computing = "Computing" }

@string{Computer = "{IEEE} Computer" }

@string{Springer = "Springer-Verlag" }

@book{liu2020sentiment,
	title        = {Sentiment analysis: Mining opinions, sentiments, and emotions},
	author       = {Liu, Bing},
	year         = 2020,
	publisher    = {Cambridge university press}
}

@inproceedings{cho2014properties,
	title        = {On the Properties of Neural Machine Translation: Encoder--Decoder Approaches},
	author       = {Cho, Kyunghyun and van Merri{\"e}nboer, Bart and Bahdanau, Dzmitry and Bengio, Yoshua},
	year         = 2014,
	booktitle    = {Proceedings of SSST-8, Eighth Workshop on Syntax, Semantics and Structure in Statistical Translation},
	pages        = {103--111}
}

@inproceedings{bahdanau2015neural,
	title        = {Neural machine translation by jointly learning to align and translate},
	author       = {Bahdanau, Dzmitry and Cho, Kyung Hyun and Bengio, Yoshua},
	year         = 2015,
	booktitle    = {3rd International Conference on Learning Representations, ICLR 2015}
}

@article{weld2022survey,
	title        = {A survey of joint intent detection and slot filling models in natural language understanding},
	author       = {Weld, Henry and Huang, Xiaoqi and Long, Siqu and Poon, Josiah and Han, Soyeon Caren},
	year         = 2022,
	journal      = {ACM Computing Surveys},
	publisher    = {ACM New York, NY},
	volume       = 55,
	number       = 8,
	pages        = {1--38}
}

@article{vaswani2017attention,
	title        = {Attention is all you need},
	author       = {Vaswani, Ashish and Shazeer, Noam and Parmar, Niki and Uszkoreit, Jakob and Jones, Llion and Gomez, Aidan N and Kaiser, {\L}ukasz and Polosukhin, Illia},
	year         = 2017,
	journal      = {Advances in neural information processing systems},
	volume       = 30
}

@inproceedings{kenton2019bert,
	title        = {Bert: Pre-training of deep bidirectional transformers for language understanding},
	author       = {Kenton, Jacob Devlin Ming-Wei Chang and Toutanova, Lee Kristina},
	year         = 2019,
	booktitle    = {Proceedings of naacL-HLT},
	volume       = 1,
	pages        = 2
}

@article{yang2019xlnet,
	title        = {Xlnet: Generalized autoregressive pretraining for language understanding},
	author       = {Yang, Zhilin and Dai, Zihang and Yang, Yiming and Carbonell, Jaime and Salakhutdinov, Russ R and Le, Quoc V},
	year         = 2019,
	journal      = {Advances in neural information processing systems},
	volume       = 32
}

@inproceedings{lewis2020bart,
	title        = {BART: Denoising Sequence-to-Sequence Pre-training for Natural Language Generation, Translation, and Comprehension},
	author       = {Lewis, Mike and Liu, Yinhan and Goyal, Naman and Ghazvininejad, Marjan and Mohamed, Abdelrahman and Levy, Omer and Stoyanov, Veselin and Zettlemoyer, Luke},
	year         = 2020,
	booktitle    = {Proceedings of the 58th Annual Meeting of the Association for Computational Linguistics},
	pages        = {7871--7880}
}

@article{radford2018improving,
	title        = {Improving language understanding by generative pre-training},
	author       = {Radford, Alec and Narasimhan, Karthik and Salimans, Tim and Sutskever, Ilya and others},
	year         = 2018,
	publisher    = {OpenAI}
}

@article{radford2019language,
	title        = {Language models are unsupervised multitask learners},
	author       = {Radford, Alec and Wu, Jeffrey and Child, Rewon and Luan, David and Amodei, Dario and Sutskever, Ilya and others},
	year         = 2019,
	journal      = {OpenAI blog},
	volume       = 1,
	number       = 8,
	pages        = 9
}

@article{brown2020language,
	title        = {Language models are few-shot learners},
	author       = {Brown, Tom and Mann, Benjamin and Ryder, Nick and Subbiah, Melanie and Kaplan, Jared D and Dhariwal, Prafulla and Neelakantan, Arvind and Shyam, Pranav and Sastry, Girish and Askell, Amanda and others},
	year         = 2020,
	journal      = {Advances in neural information processing systems},
	volume       = 33,
	pages        = {1877--1901}
}

@article{chowdhary2020natural,
	title        = {Natural language processing},
	author       = {Chowdhary, KR},
	year         = 2020,
	journal      = {Fundamentals of artificial intelligence},
	publisher    = {Springer},
	pages        = {603--649}
}

@inproceedings{ribeiro2020beyond,
	title        = {Beyond exploding and vanishing gradients: analysing RNN training using attractors and smoothness},
	author       = {Ribeiro, Ant{\^o}nio H and Tiels, Koen and Aguirre, Luis A and Sch{\"o}n, Thomas},
	year         = 2020,
	booktitle    = {International conference on artificial intelligence and statistics},
	pages        = {2370--2380},
	organization = {PMLR}
}

@inproceedings{strubell2019energy,
	title        = {Energy and Policy Considerations for Deep Learning in NLP},
	author       = {Strubell, Emma and Ganesh, Ananya and McCallum, Andrew},
	year         = 2019,
	booktitle    = {Proceedings of the 57th Annual Meeting of the Association for Computational Linguistics},
	organization = {Association for Computational Linguistics}
}

@article{patterson2021carbon,
	title        = {Carbon emissions and large neural network training},
	author       = {Patterson, David and Gonzalez, Joseph and Le, Quoc and Liang, Chen and Munguia, Lluis-Miquel and Rothchild, Daniel and So, David and Texier, Maud and Dean, Jeff},
	year         = 2021,
	journal      = {arXiv preprint arXiv:2104.10350}
}

@article{schwartz2020green,
	title        = {Green ai},
	author       = {Schwartz, Roy and Dodge, Jesse and Smith, Noah A and Etzioni, Oren},
	year         = 2020,
	journal      = {Communications of the ACM},
	publisher    = {ACM New York, NY, USA},
	volume       = 63,
	number       = 12,
	pages        = {54--63}
}

@article{treviso2023efficient,
	title        = {Efficient methods for natural language processing: A survey},
	author       = {Treviso, Marcos and Lee, Ji-Ung and Ji, Tianchu and Aken, Betty van and Cao, Qingqing and Ciosici, Manuel R and Hassid, Michael and Heafield, Kenneth and Hooker, Sara and Raffel, Colin and others},
	year         = 2023,
	journal      = {Transactions of the Association for Computational Linguistics},
	publisher    = {MIT Press One Broadway, 12th Floor, Cambridge, Massachusetts 02142, USA~…},
	volume       = 11,
	pages        = {826--860}
}

@inproceedings{lewis2018generative,
	title        = {Generative question answering: Learning to answer the whole question},
	author       = {Lewis, Mike and Fan, Angela},
	year         = 2018,
	booktitle    = {International Conference on Learning Representations}
}

@article{nallapati2016sequence,
	title        = {Sequence-to-sequence RNNs for text summarization},
	author       = {Nallapati, Ramesh and Xiang, Bing and Zhou, Bowen},
	year         = 2016,
	journal      = {arXiv preprint arXiv:1602.06023}
}

@article{ansar2021combating,
	title        = {Combating the menace: A survey on characterization and detection of fake news from a data science perspective},
	author       = {Ansar, Wazib and Goswami, Saptarsi},
	year         = 2021,
	journal      = {International Journal of Information Management Data Insights},
	publisher    = {Elsevier},
	volume       = 1,
	number       = 2,
	pages        = 100052
}

@article{mikolov2013distributed,
	title        = {Distributed representations of words and phrases and their compositionality},
	author       = {Mikolov, Tomas and Sutskever, Ilya and Chen, Kai and Corrado, Greg S and Dean, Jeff},
	year         = 2013,
	journal      = {Advances in neural information processing systems},
	volume       = 26
}

@inproceedings{pennington2014glove,
	title        = {Glove: Global vectors for word representation},
	author       = {Pennington, Jeffrey and Socher, Richard and Manning, Christopher D},
	year         = 2014,
	booktitle    = {Proceedings of the 2014 conference on empirical methods in natural language processing (EMNLP)},
	pages        = {1532--1543}
}

@inproceedings{irsoy2014opinion,
	title        = {Opinion mining with deep recurrent neural networks},
	author       = {Irsoy, Ozan and Cardie, Claire},
	year         = 2014,
	booktitle    = {Proceedings of the 2014 conference on empirical methods in natural language processing (EMNLP)},
	pages        = {720--728}
}

@article{hinton2015distilling,
	title        = {Distilling the Knowledge in a Neural Network},
	author       = {Hinton, Geoffrey and Vinyals, Oriol and Dean, Jeff},
	year         = 2015,
	journal      = {stat},
	volume       = 1050,
	pages        = 9
}

@article{sharma2020sentimental,
	title        = {Sentimental short sentences classification by using CNN deep learning model with fine tuned Word2Vec},
	author       = {Sharma, Amit Kumar and Chaurasia, Sandeep and Srivastava, Devesh Kumar},
	year         = 2020,
	journal      = {Procedia Computer Science},
	publisher    = {Elsevier},
	volume       = 167,
	pages        = {1139--1147}
}

@inproceedings{howard2018universal,
	title        = {Universal Language Model Fine-tuning for Text Classification},
	author       = {Howard, Jeremy and Ruder, Sebastian},
	year         = 2018,
	booktitle    = {Proceedings of the 56th Annual Meeting of the Association for Computational Linguistics (Volume 1: Long Papers)},
	pages        = {328--339}
}

@inproceedings{peters-etal-2018-deep,
	title        = {Deep Contextualized Word Representations},
	author       = {Peters, Matthew E.  and Neumann, Mark  and Iyyer, Mohit  and Gardner, Matt  and Clark, Christopher  and Lee, Kenton  and Zettlemoyer, Luke},
	year         = 2018,
	month        = jun,
	booktitle    = {Proceedings of the 2018 Conference of the North {A}merican Chapter of the Association for Computational Linguistics: Human Language Technologies, Volume 1 (Long Papers)},
	publisher    = {Association for Computational Linguistics},
	address      = {New Orleans, Louisiana},
	pages        = {2227--2237},
	doi          = {10.18653/v1/N18-1202},
	url          = {https://aclanthology.org/N18-1202}
}

@inproceedings{dai2019transformer,
	title        = {Transformer-XL: Attentive Language Models beyond a Fixed-Length Context},
	author       = {Dai, Zihang and Yang, Zhilin and Yang, Yiming and Carbonell, Jaime and Le, Quoc and Salakhutdinov, Ruslan},
	year         = 2019,
	booktitle    = {Proceedings of the 57th Annual Meeting of the Association for Computational Linguistics},
	organization = {Association for Computational Linguistics}
}

@inproceedings{lan2019albert,
	title        = {ALBERT: A Lite BERT for Self-supervised Learning of Language Representations},
	author       = {Lan, Zhenzhong and Chen, Mingda and Goodman, Sebastian and Gimpel, Kevin and Sharma, Piyush and Soricut, Radu},
	year         = 2019,
	booktitle    = {International Conference on Learning Representations}
}

@inproceedings{kitaev2019reformer,
	title        = {Reformer: The Efficient Transformer},
	author       = {Kitaev, Nikita and Kaiser, Lukasz and Levskaya, Anselm},
	year         = 2019,
	booktitle    = {International Conference on Learning Representations}
}

@article{wang2020linformer,
	title        = {Linformer: Self-attention with linear complexity},
	author       = {Wang, Sinong and Li, Belinda Z and Khabsa, Madian and Fang, Han and Ma, Hao},
	year         = 2020,
	journal      = {arXiv preprint arXiv:2006.04768}
}

@article{roy2021efficient,
	title        = {Efficient content-based sparse attention with routing transformers},
	author       = {Roy, Aurko and Saffar, Mohammad and Vaswani, Ashish and Grangier, David},
	year         = 2021,
	journal      = {Transactions of the Association for Computational Linguistics},
	publisher    = {MIT Press One Rogers Street, Cambridge, MA 02142-1209, USA journals-info~…},
	volume       = 9,
	pages        = {53--68}
}

@inproceedings{jaegle2021perceiver,
	title        = {Perceiver: General perception with iterative attention},
	author       = {Jaegle, Andrew and Gimeno, Felix and Brock, Andy and Vinyals, Oriol and Zisserman, Andrew and Carreira, Joao},
	year         = 2021,
	booktitle    = {International conference on machine learning},
	pages        = {4651--4664},
	organization = {PMLR}
}

@article{zhang2022opt,
	title        = {Opt: Open pre-trained transformer language models},
	author       = {Zhang, Susan and Roller, Stephen and Goyal, Naman and Artetxe, Mikel and Chen, Moya and Chen, Shuohui and Dewan, Christopher and Diab, Mona and Li, Xian and Lin, Xi Victoria and others},
	year         = 2022,
	journal      = {arXiv preprint arXiv:2205.01068}
}

@article{sajjad2023effect,
	title        = {On the effect of dropping layers of pre-trained transformer models},
	author       = {Sajjad, Hassan and Dalvi, Fahim and Durrani, Nadir and Nakov, Preslav},
	year         = 2023,
	journal      = {Computer Speech \& Language},
	publisher    = {Elsevier},
	volume       = 77,
	pages        = 101429
}

@article{ansar2021efficient,
	title        = {An efficient methodology for aspect-based sentiment analysis using BERT through refined aspect extraction},
	author       = {Ansar, Wazib and Goswami, Saptarsi and Chakrabarti, Amlan and Chakraborty, Basabi},
	year         = 2021,
	journal      = {Journal of Intelligent \& Fuzzy Systems},
	publisher    = {IOS Press},
	volume       = 40,
	number       = 5,
	pages        = {9627--9644}
}

@article{jelodar2020deep,
	title        = {Deep sentiment classification and topic discovery on novel coronavirus or COVID-19 online discussions: NLP using LSTM recurrent neural network approach},
	author       = {Jelodar, Hamed and Wang, Yongli and Orji, Rita and Huang, Shucheng},
	year         = 2020,
	journal      = {IEEE Journal of Biomedical and Health Informatics},
	publisher    = {IEEE},
	volume       = 24,
	number       = 10,
	pages        = {2733--2742}
}

@article{hoffmann2022empirical,
	title        = {An empirical analysis of compute-optimal large language model training},
	author       = {Hoffmann, Jordan and Borgeaud, Sebastian and Mensch, Arthur and Buchatskaya, Elena and Cai, Trevor and Rutherford, Eliza and de Las Casas, Diego and Hendricks, Lisa Anne and Welbl, Johannes and Clark, Aidan and others},
	year         = 2022,
	journal      = {Advances in Neural Information Processing Systems},
	volume       = 35,
	pages        = {30016--30030}
}

@inproceedings{lee2022deduplicating,
	title        = {Deduplicating Training Data Makes Language Models Better},
	author       = {Lee, Katherine and Ippolito, Daphne and Nystrom, Andrew and Zhang, Chiyuan and Eck, Douglas and Callison-Burch, Chris and Carlini, Nicholas},
	year         = 2022,
	booktitle    = {Proceedings of the 60th Annual Meeting of the Association for Computational Linguistics (Volume 1: Long Papers)},
	pages        = {8424--8445}
}

@inproceedings{mishra2020we,
	title        = {Do we need to create big datasets to learn a task?},
	author       = {Mishra, Swaroop and Sachdeva, Bhavdeep Singh},
	year         = 2020,
	booktitle    = {Proceedings of SustaiNLP: Workshop on Simple and Efficient Natural Language Processing},
	pages        = {169--173}
}

@inproceedings{le2020adversarial,
	title        = {Adversarial filters of dataset biases},
	author       = {Le Bras, Ronan and Swayamdipta, Swabha and Bhagavatula, Chandra and Zellers, Rowan and Peters, Matthew and Sabharwal, Ashish and Choi, Yejin},
	year         = 2020,
	booktitle    = {International conference on machine learning},
	pages        = {1078--1088},
	organization = {PMLR}
}

@article{ren2021survey,
	title        = {A survey of deep active learning},
	author       = {Ren, Pengzhen and Xiao, Yun and Chang, Xiaojun and Huang, Po-Yao and Li, Zhihui and Gupta, Brij B and Chen, Xiaojiang and Wang, Xin},
	year         = 2021,
	journal      = {ACM computing surveys (CSUR)},
	publisher    = {ACM New York, NY},
	volume       = 54,
	number       = 9,
	pages        = {1--40}
}

@inproceedings{yuan2020cold,
	title        = {Cold-start Active Learning through Self-supervised Language Modeling},
	author       = {Yuan, Michelle and Lin, Hsuan-Tien and Boyd-Graber, Jordan},
	year         = 2020,
	booktitle    = {Proceedings of the 2020 Conference on Empirical Methods in Natural Language Processing (EMNLP)},
	pages        = {7935--7948}
}

@inproceedings{margatina2021active,
	title        = {Active Learning by Acquiring Contrastive Examples},
	author       = {Margatina, Katerina and Vernikos, Giorgos and Barrault, Lo{\"\i}c and Aletras, Nikolaos},
	year         = 2021,
	booktitle    = {Proceedings of the 2021 Conference on Empirical Methods in Natural Language Processing},
	pages        = {650--663}
}

@inproceedings{margatina2023active,
	title        = {Active Learning Principles for In-Context Learning with Large Language Models},
	author       = {Margatina, Katerina and Schick, Timo and Aletras, Nikolaos and Dwivedi-Yu, Jane},
	year         = 2023,
	booktitle    = {Findings of the Association for Computational Linguistics: EMNLP 2023},
	pages        = {5011--5034}
}

@inproceedings{rouzegar2024enhancing,
	title        = {Enhancing Text Classification through LLM-Driven Active Learning and Human Annotation},
	author       = {Rouzegar, Hamidreza and Makrehchi, Masoud},
	year         = 2024,
	booktitle    = {Proceedings of the 18th Linguistic Annotation Workshop (LAW-XVIII)},
	pages        = {98--111}
}

@inproceedings{yuan2024hide,
	title        = {Hide and Seek in Noise Labels: Noise-Robust Collaborative Active Learning with LLMs-Powered Assistance},
	author       = {Yuan, Bo and Chen, Yulin and Zhang, Yin and Jiang, Wei},
	year         = 2024,
	booktitle    = {Proceedings of the 62nd Annual Meeting of the Association for Computational Linguistics (Volume 1: Long Papers)},
	pages        = {10977--11011}
}

@inproceedings{lesci2024anchoral,
	title        = {AnchorAL: Computationally Efficient Active Learning for Large and Imbalanced Datasets},
	author       = {Lesci, Pietro and Vlachos, Andreas},
	year         = 2024,
	booktitle    = {Proceedings of the 2024 Conference of the North American Chapter of the Association for Computational Linguistics: Human Language Technologies (Volume 1: Long Papers)},
	pages        = {8438--8457}
}

@inproceedings{press2021shortformer,
	title        = {Shortformer: Better Language Modeling using Shorter Inputs},
	author       = {Press, Ofir and Smith, Noah A and Lewis, Mike},
	year         = 2021,
	booktitle    = {Proceedings of the 59th Annual Meeting of the Association for Computational Linguistics and the 11th International Joint Conference on Natural Language Processing (Volume 1: Long Papers)},
	pages        = {5493--5505}
}

@incollection{ansar2023texim,
	title        = {TexIm: A Novel Text-to-Image Encoding Technique Using BERT},
	author       = {Ansar, Wazib and Goswami, Saptarsi and Chakrabarti, Amlan and Chakraborty, Basabi},
	year         = 2023,
	booktitle    = {Computer Vision and Machine Intelligence: Proceedings of CVMI 2022},
	publisher    = {Springer},
	pages        = {123--139}
}

@inproceedings{qiu2020blockwise,
	title        = {Blockwise Self-Attention for Long Document Understanding},
	author       = {Qiu, Jiezhong and Ma, Hao and Levy, Omer and Yih, Wen-tau and Wang, Sinong and Tang, Jie},
	year         = 2020,
	booktitle    = {Findings of the Association for Computational Linguistics: EMNLP 2020},
	pages        = {2555--2565}
}

@article{beltagy2020longformer,
	title        = {Longformer: The long-document transformer},
	author       = {Beltagy, Iz and Peters, Matthew E and Cohan, Arman},
	year         = 2020,
	journal      = {arXiv preprint arXiv:2004.05150}
}

@inproceedings{liu2018generating,
	title        = {Generating Wikipedia by Summarizing Long Sequences},
	author       = {Liu, Peter J and Saleh, Mohammad and Pot, Etienne and Goodrich, Ben and Sepassi, Ryan and Kaiser, Lukasz and Shazeer, Noam},
	year         = 2018,
	booktitle    = {International Conference on Learning Representations}
}

@article{child2019generating,
	title        = {Generating long sequences with sparse transformers},
	author       = {Child, Rewon and Gray, Scott and Radford, Alec and Sutskever, Ilya},
	year         = 2019,
	journal      = {arXiv preprint arXiv:1904.10509}
}

@inproceedings{du2022glam,
	title        = {Glam: Efficient scaling of language models with mixture-of-experts},
	author       = {Du, Nan and Huang, Yanping and Dai, Andrew M and Tong, Simon and Lepikhin, Dmitry and Xu, Yuanzhong and Krikun, Maxim and Zhou, Yanqi and Yu, Adams Wei and Firat, Orhan and others},
	year         = 2022,
	booktitle    = {International Conference on Machine Learning},
	pages        = {5547--5569},
	organization = {PMLR}
}

@article{choromanski2020masked,
	title        = {Masked language modeling for proteins via linearly scalable long-context transformers},
	author       = {Choromanski, Krzysztof and Likhosherstov, Valerii and Dohan, David and Song, Xingyou and Gane, Andreea and Sarlos, Tamas and Hawkins, Peter and Davis, Jared and Belanger, David and Colwell, Lucy and others},
	year         = 2020,
	journal      = {arXiv preprint arXiv:2006.03555}
}

@article{liang2021pruning,
	title        = {Pruning and quantization for deep neural network acceleration: A survey},
	author       = {Liang, Tailin and Glossner, John and Wang, Lei and Shi, Shaobo and Zhang, Xiaotong},
	year         = 2021,
	journal      = {Neurocomputing},
	publisher    = {Elsevier},
	volume       = 461,
	pages        = {370--403}
}

@inproceedings{louizos2018learning,
	title        = {Learning Sparse Neural Networks through L\_0 Regularization},
	author       = {Louizos, Christos and Welling, Max and Kingma, Diederik P},
	year         = 2018,
	booktitle    = {International Conference on Learning Representations}
}

@article{sanh2020movement,
	title        = {Movement pruning: Adaptive sparsity by fine-tuning},
	author       = {Sanh, Victor and Wolf, Thomas and Rush, Alexander},
	year         = 2020,
	journal      = {Advances in Neural Information Processing Systems},
	volume       = 33,
	pages        = {20378--20389}
}

@inproceedings{fan2019reducing,
	title        = {Reducing Transformer Depth on Demand with Structured Dropout},
	author       = {Fan, Angela and Grave, Edouard and Joulin, Armand},
	year         = 2019,
	booktitle    = {International Conference on Learning Representations}
}

@article{stanton2021does,
	title        = {Does knowledge distillation really work?},
	author       = {Stanton, Samuel and Izmailov, Pavel and Kirichenko, Polina and Alemi, Alexander A and Wilson, Andrew G},
	year         = 2021,
	journal      = {Advances in Neural Information Processing Systems},
	volume       = 34,
	pages        = {6906--6919}
}

@inproceedings{he2022fastermoe,
	title        = {FasterMoE: modeling and optimizing training of large-scale dynamic pre-trained models},
	author       = {He, Jiaao and Zhai, Jidong and Antunes, Tiago and Wang, Haojie and Luo, Fuwen and Shi, Shangfeng and Li, Qin},
	year         = 2022,
	booktitle    = {Proceedings of the 27th ACM SIGPLAN Symposium on Principles and Practice of Parallel Programming},
	pages        = {120--134}
}

@inproceedings{wolff2020carbontracker,
	title        = {Carbontracker: Tracking and predicting the carbon footprint of training deep learning models},
	author       = {Anthony, Lasse F Wolff and Kanding, Benjamin and Selvan, Raghavendra},
	year         = 2020,
	booktitle    = {ICML Workshop 2020 on Challenges in Deploying and monitoring Machine Learning Systems}
}

@inproceedings{durlich2023concept,
	title        = {On the Concept of Resource-Efficiency in NLP},
	author       = {D{\"u}rlich, Luise and Gogoulou, Evangelia and Nivre, Joakim},
	year         = 2023,
	booktitle    = {Proceedings of the 24th Nordic Conference on Computational Linguistics (NoDaLiDa)},
	pages        = {135--145}
}

@inproceedings{gupta2021chasing,
	title        = {Chasing carbon: The elusive environmental footprint of computing},
	author       = {Gupta, Udit and Kim, Young Geun and Lee, Sylvia and Tse, Jordan and Lee, Hsien-Hsin S and Wei, Gu-Yeon and Brooks, David and Wu, Carole-Jean},
	year         = 2021,
	booktitle    = {2021 IEEE International Symposium on High-Performance Computer Architecture (HPCA)},
	pages        = {854--867},
	organization = {IEEE}
}

@article{sutskever2014sequence,
	title        = {Sequence to sequence learning with neural networks},
	author       = {Sutskever, Ilya and Vinyals, Oriol and Le, Quoc V},
	year         = 2014,
	journal      = {Advances in neural information processing systems},
	volume       = 27
}

@inproceedings{lin2017structured,
	title        = {A structured self-attentive sentence embedding},
	author       = {Lin, Zhouhan and Feng, Minwei and dos Santos, Cicero and Yu, Mo and Xiang, Bing and Zhou, Bowen and Bengio, Yoshua},
	year         = 2017,
	booktitle    = {International Conference on Learning Representations},
	organization = {International Conference on Learning Representations, ICLR}
}

@inproceedings{kim2016structured,
	title        = {Structured Attention Networks},
	author       = {Kim, Yoon and Denton, Carl and Hoang, Luong and Rush, Alexander M},
	year         = 2016,
	booktitle    = {International Conference on Learning Representations}
}

@article{graves2014neural,
	title        = {Neural turing machines},
	author       = {Graves, Alex and Wayne, Greg and Danihelka, Ivo},
	year         = 2014,
	journal      = {arXiv preprint arXiv:1410.5401}
}

@inproceedings{luong2015effective,
	title        = {Effective Approaches to Attention-based Neural Machine Translation},
	author       = {Luong, Minh-Thang and Pham, Hieu and Manning, Christopher D},
	year         = 2015,
	booktitle    = {Proceedings of the 2015 Conference on Empirical Methods in Natural Language Processing},
	pages        = {1412--1421}
}

@inproceedings{hoffmann2022training,
	title        = {Training compute-optimal large language models},
	author       = {Hoffmann, Jordan and Borgeaud, Sebastian and Mensch, Arthur and Buchatskaya, Elena and Cai, Trevor and Rutherford, Eliza and de Las Casas, Diego and Hendricks, Lisa Anne and Welbl, Johannes and Clark, Aidan and others},
	year         = 2022,
	booktitle    = {Proceedings of the 36th International Conference on Neural Information Processing Systems},
	pages        = {30016--30030}
}

@article{collobert2011natural,
	title        = {Natural language processing (almost) from scratch},
	author       = {Collobert, Ronan and Weston, Jason and Bottou, L{\'e}on and Karlen, Michael and Kavukcuoglu, Koray and Kuksa, Pavel},
	year         = 2011,
	journal      = {Journal of machine learning research},
	volume       = 12,
	number       = {ARTICLE},
	pages        = {2493--2537}
}

@article{dai2015semi,
	title        = {Semi-supervised sequence learning},
	author       = {Dai, Andrew M and Le, Quoc V},
	year         = 2015,
	journal      = {Advances in neural information processing systems},
	volume       = 28
}

@article{raffel2020exploring,
	title        = {Exploring the limits of transfer learning with a unified text-to-text transformer},
	author       = {Raffel, Colin and Shazeer, Noam and Roberts, Adam and Lee, Katherine and Narang, Sharan and Matena, Michael and Zhou, Yanqi and Li, Wei and Liu, Peter J},
	year         = 2020,
	journal      = {The Journal of Machine Learning Research},
	publisher    = {JMLRORG},
	volume       = 21,
	number       = 1,
	pages        = {5485--5551}
}

@article{liu2019roberta,
	title        = {Roberta: A robustly optimized bert pretraining approach},
	author       = {Liu, Yinhan and Ott, Myle and Goyal, Naman and Du, Jingfei and Joshi, Mandar and Chen, Danqi and Levy, Omer and Lewis, Mike and Zettlemoyer, Luke and Stoyanov, Veselin},
	year         = 2019,
	journal      = {arXiv preprint arXiv:1907.11692}
}

@article{rogers2021primer,
	title        = {A primer in BERTology: What we know about how BERT works},
	author       = {Rogers, Anna and Kovaleva, Olga and Rumshisky, Anna},
	year         = 2021,
	journal      = {Transactions of the Association for Computational Linguistics},
	publisher    = {MIT Press One Rogers Street, Cambridge, MA 02142-1209, USA journals-info~…},
	volume       = 8,
	pages        = {842--866}
}

@inproceedings{houlsby2019parameter,
	title        = {Parameter-efficient transfer learning for NLP},
	author       = {Houlsby, Neil and Giurgiu, Andrei and Jastrzebski, Stanislaw and Morrone, Bruna and De Laroussilhe, Quentin and Gesmundo, Andrea and Attariyan, Mona and Gelly, Sylvain},
	year         = 2019,
	booktitle    = {International Conference on Machine Learning},
	pages        = {2790--2799},
	organization = {PMLR}
}

@article{karimi2021compacter,
	title        = {Compacter: Efficient low-rank hypercomplex adapter layers},
	author       = {Karimi Mahabadi, Rabeeh and Henderson, James and Ruder, Sebastian},
	year         = 2021,
	journal      = {Advances in Neural Information Processing Systems},
	volume       = 34,
	pages        = {1022--1035}
}

@inproceedings{aghajanyan2021intrinsic,
	title        = {Intrinsic Dimensionality Explains the Effectiveness of Language Model Fine-Tuning},
	author       = {Aghajanyan, Armen and Gupta, Sonal and Zettlemoyer, Luke},
	year         = 2021,
	booktitle    = {Proceedings of the 59th Annual Meeting of the Association for Computational Linguistics and the 11th International Joint Conference on Natural Language Processing (Volume 1: Long Papers)},
	pages        = {7319--7328}
}

@inproceedings{wang2022adamix,
	title        = {AdaMix: Mixture-of-Adaptations for Parameter-efficient Model Tuning},
	author       = {Wang, Yaqing and Agarwal, Sahaj and Mukherjee, Subhabrata and Liu, Xiaodong and Gao, Jing and Hassan, Ahmed and Gao, Jianfeng},
	year         = 2022,
	booktitle    = {Proceedings of the 2022 Conference on Empirical Methods in Natural Language Processing},
	pages        = {5744--5760}
}

@inproceedings{schick2021generating,
	title        = {Generating Datasets with Pretrained Language Models},
	author       = {Schick, Timo and Sch{\"u}tze, Hinrich},
	year         = 2021,
	booktitle    = {Proceedings of the 2021 Conference on Empirical Methods in Natural Language Processing},
	pages        = {6943--6951}
}

@inproceedings{wei2021finetuned,
	title        = {Finetuned Language Models are Zero-Shot Learners},
	author       = {Wei, Jason and Bosma, Maarten and Zhao, Vincent and Guu, Kelvin and Yu, Adams Wei and Lester, Brian and Du, Nan and Dai, Andrew M and Le, Quoc V},
	year         = 2021,
	booktitle    = {International Conference on Learning Representations}
}

@inproceedings{schick2021few,
	title        = {Few-shot text generation with natural language instructions},
	author       = {Schick, Timo and Sch{\"u}tze, Hinrich},
	year         = 2021,
	booktitle    = {Proceedings of the 2021 Conference on Empirical Methods in Natural Language Processing},
	pages        = {390--402}
}

@inproceedings{schick2021exploiting,
	title        = {Exploiting Cloze-Questions for Few-Shot Text Classification and Natural Language Inference},
	author       = {Schick, Timo and Sch{\"u}tze, Hinrich},
	year         = 2021,
	booktitle    = {Proceedings of the 16th Conference of the European Chapter of the Association for Computational Linguistics: Main Volume},
	pages        = {255--269}
}

@article{trinh2018simple,
	title        = {A simple method for commonsense reasoning},
	author       = {Trinh, Trieu H and Le, Quoc V},
	year         = 2018,
	journal      = {arXiv preprint arXiv:1806.02847}
}

@inproceedings{yin2019benchmarking,
	title        = {Benchmarking Zero-shot Text Classification: Datasets, Evaluation and Entailment Approach},
	author       = {Yin, Wenpeng and Hay, Jamaal and Roth, Dan},
	year         = 2019,
	booktitle    = {Proceedings of the 2019 Conference on Empirical Methods in Natural Language Processing and the 9th International Joint Conference on Natural Language Processing (EMNLP-IJCNLP)},
	pages        = {3914--3923}
}

@inproceedings{wu2020corefqa,
	title        = {CorefQA: Coreference resolution as query-based span prediction},
	author       = {Wu, Wei and Wang, Fei and Yuan, Arianna and Wu, Fei and Li, Jiwei},
	year         = 2020,
	booktitle    = {Proceedings of the 58th Annual Meeting of the Association for Computational Linguistics},
	pages        = {6953--6963}
}

@article{touvron2023llama,
	title        = {Llama: Open and efficient foundation language models},
	author       = {Touvron, Hugo and Lavril, Thibaut and Izacard, Gautier and Martinet, Xavier and Lachaux, Marie-Anne and Lacroix, Timoth{\'e}e and Rozi{\`e}re, Baptiste and Goyal, Naman and Hambro, Eric and Azhar, Faisal and others},
	year         = 2023,
	journal      = {arXiv preprint arXiv:2302.13971}
}

@article{touvron2023llama2,
	title        = {Llama 2: Open foundation and fine-tuned chat models},
	author       = {Touvron, Hugo and Martin, Louis and Stone, Kevin and Albert, Peter and Almahairi, Amjad and Babaei, Yasmine and Bashlykov, Nikolay and Batra, Soumya and Bhargava, Prajjwal and Bhosale, Shruti and others},
	year         = 2023,
	journal      = {arXiv preprint arXiv:2307.09288}
}

@article{thoppilan2022lamda,
	title        = {Lamda: Language models for dialog applications},
	author       = {Thoppilan, Romal and De Freitas, Daniel and Hall, Jamie and Shazeer, Noam and Kulshreshtha, Apoorv and Cheng, Heng-Tze and Jin, Alicia and Bos, Taylor and Baker, Leslie and Du, Yu and others},
	year         = 2022,
	journal      = {arXiv preprint arXiv:2201.08239}
}

@inproceedings{faiz2024llmcarbon,
	title        = {LLMCARBON: MODELING THE END-TO-END CARBON FOOTPRINT OF LARGE LANGUAGE MODELS},
	author       = {Faiz, Ahmad and Kaneda, Sotaro and Wang, Ruhan and Osi, Rita and Sharma, Prateek and Chen, Fan and Jiang, Lei},
	year         = 2024,
	booktitle    = {The Twelfth International Conference on Learning Representations},
	organization = {ICLR}
}

@article{chowdhery2023palm,
	title        = {Palm: Scaling language modeling with pathways},
	author       = {Chowdhery, Aakanksha and Narang, Sharan and Devlin, Jacob and Bosma, Maarten and Mishra, Gaurav and Roberts, Adam and Barham, Paul and Chung, Hyung Won and Sutton, Charles and Gehrmann, Sebastian and others},
	year         = 2023,
	journal      = {Journal of Machine Learning Research},
	volume       = 24,
	number       = 240,
	pages        = {1--113}
}

@inproceedings{xu2023survey,
	title        = {A survey on model compression and acceleration for pretrained language models},
	author       = {Xu, Canwen and McAuley, Julian},
	year         = 2023,
	booktitle    = {Proceedings of the AAAI Conference on Artificial Intelligence},
	volume       = 37,
	number       = 9,
	pages        = {10566--10575}
}

@inproceedings{zhang2020pegasus,
	title        = {Pegasus: Pre-training with extracted gap-sentences for abstractive summarization},
	author       = {Zhang, Jingqing and Zhao, Yao and Saleh, Mohammad and Liu, Peter},
	year         = 2020,
	booktitle    = {International Conference on Machine Learning},
	pages        = {11328--11339},
	organization = {PMLR}
}

@article{khadivi2023bibliometric,
	title        = {A Bibliometric Study of Natural Language Processing Using Dimensions Database: Development, Research Trend, and Future Research Directions},
	author       = {Khadivi, Nasim and Sato, Sho},
	year         = 2023,
	journal      = {Journal of Data Science, Informetrics, and Citation Studies},
	volume       = 2,
	number       = 2,
	pages        = {77--89}
}

@inproceedings{bannour2021evaluating,
	title        = {Evaluating the carbon footprint of NLP methods: a survey and analysis of existing tools},
	author       = {Bannour, Nesrine and Ghannay, Sahar and N{\'e}v{\'e}ol, Aur{\'e}lie and Ligozat, Anne-Laure},
	year         = 2021,
	booktitle    = {Proceedings of the Second Workshop on Simple and Efficient Natural Language Processing},
	pages        = {11--21}
}

@article{moradi2025critical,
	title        = {A Critical Review of Methods and Challenges in Large Language Models.},
	author       = {Moradi, Milad and Yan, Ke and Colwell, David and Samwald, Matthias and Asgari, Rhona},
	year         = 2025,
	journal      = {Computers, Materials \& Continua},
	volume       = 82,
	number       = 2
}

@article{catania2023conversational,
	title        = {Conversational agents in therapeutic interventions for neurodevelopmental disorders: a survey},
	author       = {Catania, Fabio and Spitale, Micol and Garzotto, Franca},
	year         = 2023,
	journal      = {ACM Computing Surveys},
	publisher    = {ACM New York, NY},
	volume       = 55,
	number       = 10,
	pages        = {1--34}
}

@inproceedings{ainslie2020etc,
	title        = {ETC: Encoding Long and Structured Inputs in Transformers},
	author       = {Ainslie, Joshua and Ontanon, Santiago and Alberti, Chris and Cvicek, Vaclav and Fisher, Zachary and Pham, Philip and Ravula, Anirudh and Sanghai, Sumit and Wang, Qifan and Yang, Li},
	year         = 2020,
	booktitle    = {Proceedings of the 2020 Conference on Empirical Methods in Natural Language Processing (EMNLP)},
	pages        = {268--284}
}

@article{zaheer2020big,
	title        = {Big bird: Transformers for longer sequences},
	author       = {Zaheer, Manzil and Guruganesh, Guru and Dubey, Kumar Avinava and Ainslie, Joshua and Alberti, Chris and Ontanon, Santiago and Pham, Philip and Ravula, Anirudh and Wang, Qifan and Yang, Li and others},
	year         = 2020,
	journal      = {Advances in neural information processing systems},
	volume       = 33,
	pages        = {17283--17297}
}

@inproceedings{moosavi2022adaptable,
	title        = {Adaptable Adapters},
	author       = {Moosavi, Nafise Sadat and Delfosse, Quentin and Kersting, Kristian and Gurevych, Iryna},
	year         = 2022,
	booktitle    = {Proceedings of the 2022 Conference of the North American Chapter of the Association for Computational Linguistics: Human Language Technologies},
	pages        = {3742--3753}
}

@inproceedings{li2021prefix,
	title        = {Prefix-Tuning: Optimizing Continuous Prompts for Generation},
	author       = {Li, Xiang Lisa and Liang, Percy},
	year         = 2021,
	booktitle    = {Proceedings of the 59th Annual Meeting of the Association for Computational Linguistics and the 11th International Joint Conference on Natural Language Processing (Volume 1: Long Papers)},
	organization = {Association for Computational Linguistics}
}

@inproceedings{ben2022bitfit,
	title        = {BitFit: Simple Parameter-efficient Fine-tuning for Transformer-based Masked Language-models},
	author       = {Ben-Zaken, Elad and Ravfogel, Shauli and Goldberg, Yoav},
	year         = 2022,
	booktitle    = {60th Annual Meeting of the Association for Computational Linguistics, ACL 2022},
	pages        = {1--9},
	organization = {Association for Computational Linguistics (ACL)}
}

@inproceedings{xu2021raise,
	title        = {Raise a Child in Large Language Model: Towards Effective and Generalizable Fine-tuning},
	author       = {Xu, Runxin and Luo, Fuli and Zhang, Zhiyuan and Tan, Chuanqi and Chang, Baobao and Huang, Songfang and Huang, Fei},
	year         = 2021,
	booktitle    = {Proceedings of the 2021 Conference on Empirical Methods in Natural Language Processing},
	pages        = {9514--9528}
}

@article{hu2022lora,
	title        = {Lora: Low-rank adaptation of large language models.},
	author       = {Hu, Edward J and Shen, Yelong and Wallis, Phillip and Allen-Zhu, Zeyuan and Li, Yuanzhi and Wang, Shean and Wang, Lu and Chen, Weizhu and others},
	year         = 2022,
	journal      = {ICLR},
	volume       = 1,
	number       = 2,
	pages        = 3
}

@inproceedings{zhang2023adaptive,
	title        = {Adaptive Budget Allocation for Parameter-Efficient Fine-Tuning},
	author       = {Zhang, Qingru and Chen, Minshuo and Bukharin, Alexander and He, Pengcheng and Cheng, Yu and Chen, Weizhu and Zhao, Tuo},
	year         = 2023,
	booktitle    = {International Conference on Learning Representations},
	organization = {Openreview}
}

@inproceedings{ding2023sparse,
	title        = {Sparse Low-rank Adaptation of Pre-trained Language Models},
	author       = {Ding, Ning and Lv, Xingtai and Wang, Qiaosen and Chen, Yulin and Zhou, Bowen and Liu, Zhiyuan and Sun, Maosong},
	year         = 2023,
	booktitle    = {Proceedings of the 2023 Conference on Empirical Methods in Natural Language Processing},
	pages        = {4133--4145}
}

@article{dettmers2023qlora,
	title        = {Qlora: Efficient finetuning of quantized llms, 2023},
	author       = {Dettmers, Tim and Pagnoni, Artidoro and Holtzman, Ari and Zettlemoyer, Luke},
	year         = 2023,
	journal      = {URL https://arxiv. org/abs/2305.14314},
	volume       = 2
}

@inproceedings{valipour2023dylora,
	title        = {DyLoRA: Parameter-Efficient Tuning of Pre-trained Models using Dynamic Search-Free Low-Rank Adaptation},
	author       = {Valipour, Mojtaba and Rezagholizadeh, Mehdi and Kobyzev, Ivan and Ghodsi, Ali},
	year         = 2023,
	booktitle    = {Proceedings of the 17th Conference of the European Chapter of the Association for Computational Linguistics},
	pages        = {3274--3287}
}

@article{wei2022chain,
	title        = {Chain-of-thought prompting elicits reasoning in large language models},
	author       = {Wei, Jason and Wang, Xuezhi and Schuurmans, Dale and Bosma, Maarten and Xia, Fei and Chi, Ed and Le, Quoc V and Zhou, Denny and others},
	year         = 2022,
	journal      = {Advances in neural information processing systems},
	volume       = 35,
	pages        = {24824--24837}
}

@article{xu2025chain,
	title        = {Chain of draft: Thinking faster by writing less},
	author       = {Xu, Silei and Xie, Wenhao and Zhao, Lingxiao and He, Pengcheng},
	year         = 2025,
	journal      = {arXiv preprint arXiv:2502.18600}
}

@article{hao2024training,
	title        = {Training large language models to reason in a continuous latent space},
	author       = {Hao, Shibo and Sukhbaatar, Sainbayar and Su, DiJia and Li, Xian and Hu, Zhiting and Weston, Jason and Tian, Yuandong},
	year         = 2024,
	journal      = {arXiv preprint arXiv:2412.06769}
}

@article{ning2023skeleton,
	title        = {Skeleton-of-thought: Large language models can do parallel decoding},
	author       = {Ning, Xuefei and Lin, Zinan and Zhou, Zixuan and Wang, Zifu and Yang, Huazhong and Wang, Yu},
	year         = 2023,
	journal      = {Proceedings ENLSP-III}
}

@article{nayab2024concise,
	title        = {Concise thoughts: Impact of output length on llm reasoning and cost},
	author       = {Nayab, Sania and Rossolini, Giulio and Simoni, Marco and Saracino, Andrea and Buttazzo, Giorgio and Manes, Nicolamaria and Giacomelli, Fabrizio},
	year         = 2024,
	journal      = {arXiv preprint arXiv:2407.19825}
}

@inproceedings{han2025token,
	title        = {Token-budget-aware llm reasoning},
	author       = {Han, Tingxu and Wang, Zhenting and Fang, Chunrong and Zhao, Shiyu and Ma, Shiqing and Chen, Zhenyu},
	year         = 2025,
	booktitle    = {Findings of the Association for Computational Linguistics: ACL 2025},
	pages        = {24842--24855}
}

@article{lewis2020retrieval,
	title        = {Retrieval-augmented generation for knowledge-intensive nlp tasks},
	author       = {Lewis, Patrick and Perez, Ethan and Piktus, Aleksandra and Petroni, Fabio and Karpukhin, Vladimir and Goyal, Naman and K{\"u}ttler, Heinrich and Lewis, Mike and Yih, Wen-tau and Rockt{\"a}schel, Tim and others},
	year         = 2020,
	journal      = {Advances in neural information processing systems},
	volume       = 33,
	pages        = {9459--9474}
}

@inproceedings{li2024refiner,
	title        = {Refiner: Restructure Retrieved Content Efficiently to Advance Question-Answering Capabilities},
	author       = {Li, Zhonghao and Hu, Xuming and Liu, Aiwei and Zheng, Kening and Huang, Sirui and Xiong, Hui},
	year         = 2024,
	booktitle    = {Findings of the Association for Computational Linguistics: EMNLP 2024},
	pages        = {8548--8572}
}

@inproceedings{chan2024rqrag,
	title        = {{RQ}-{RAG}: Learning to Refine Queries for Retrieval Augmented Generation},
	author       = {Chi-Min Chan and Chunpu Xu and Ruibin Yuan and Hongyin Luo and Wei Xue and Yike Guo and Jie Fu},
	year         = 2024,
	booktitle    = {First Conference on Language Modeling},
	url          = {https://openreview.net/forum?id=tzE7VqsaJ4}
}

@article{grootendorst2022bertopic,
	title        = {BERTopic: Neural topic modeling with a class-based TF-IDF procedure},
	author       = {Grootendorst, Maarten},
	year         = 2022,
	journal      = {arXiv preprint arXiv:2203.05794}
}

@article{rezaei2024rag,
	title        = {At-rag: An adaptive rag model enhancing query efficiency with topic filtering and iterative reasoning},
	author       = {Rezaei, Mohammad Reza and Hafezi, Maziar and Satpathy, Amit and Hodge, Lovell and Pourjafari, Ebrahim},
	year         = 2024,
	journal      = {arXiv preprint arXiv:2410.12886}
}

@incollection{shi2024enhancing,
	title        = {Enhancing Retrieval and Managing Retrieval: A Four-Module Synergy for Improved Quality and Efficiency in RAG Systems},
	author       = {Shi, Yunxiao and Zi, Xing and Shi, Zijing and Zhang, Haimin and Wu, Qiang and Xu, Min},
	year         = 2024,
	booktitle    = {ECAI 2024},
	publisher    = {IOS Press},
	pages        = {2258--2265}
}

@inproceedings{jeong2024adaptive,
	title        = {Adaptive-RAG: Learning to Adapt Retrieval-Augmented Large Language Models through Question Complexity},
	author       = {Jeong, Soyeong and Baek, Jinheon and Cho, Sukmin and Hwang, Sung Ju and Park, Jong C},
	year         = 2024,
	booktitle    = {Proceedings of the 2024 Conference of the North American Chapter of the Association for Computational Linguistics: Human Language Technologies (Volume 1: Long Papers)},
	pages        = {7029--7043}
}

@inproceedings{asai2023self,
	title        = {Self-rag: Self-reflective retrieval augmented generation},
	author       = {Asai, Akari and Wu, Zeqiu and Wang, Yizhong and Sil, Avirup and Hajishirzi, Hannaneh},
	year         = 2023,
	booktitle    = {NeurIPS 2023 workshop on instruction tuning and instruction following}
}

@article{zhang2024mapreduce,
	title        = {A mapreduce approach to effectively utilize long context information in retrieval augmented language models},
	author       = {Zhang, Gongbo and Xu, Zihan and Jin, Qiao and Chen, Fangyi and Fang, Yilu and Liu, Yi and Rousseau, Justin F and Xu, Ziyang and Lu, Zhiyong and Weng, Chunhua and others},
	year         = 2024,
	journal      = {arXiv preprint arXiv:2412.15271}
}

@article{dong2024don,
	title        = {Don't forget to connect! improving rag with graph-based reranking},
	author       = {Dong, Jialin and Fatemi, Bahare and Perozzi, Bryan and Yang, Lin F and Tsitsulin, Anton},
	year         = 2024,
	journal      = {arXiv preprint arXiv:2405.18414}
}

@article{yan2024corrective,
	title        = {Corrective Retrieval Augmented Generation},
	author       = {Yan, Shi-Qi and Gu, Jia-Chen and Zhu, Yun and Ling, Zhen-Hua},
	year         = 2024,
	journal      = {arXiv preprint arXiv:2401.15884}
}

@inproceedings{lawton2024quailora,
	title        = {QuAILoRA: Quantization-Aware Initialization for LoRA},
	author       = {Lawton, Neal G and Padmakumar, Aishwarya and Gaspers, Judith and FitzGerald, Jack and Kumar, Anoop and Ver Steeg, Greg and Galstyan, Aram},
	year         = 2024,
	booktitle    = {NeurIPS Efficient Natural Language and Speech Processing Workshop},
	pages        = {22--33},
	organization = {PMLR}
}

@article{muennighoff2023scaling,
	title        = {Scaling data-constrained language models},
	author       = {Muennighoff, Niklas and Rush, Alexander and Barak, Boaz and Le Scao, Teven and Tazi, Nouamane and Piktus, Aleksandra and Pyysalo, Sampo and Wolf, Thomas and Raffel, Colin A},
	year         = 2023,
	journal      = {Advances in Neural Information Processing Systems},
	volume       = 36,
	pages        = {50358--50376}
}

@inproceedings{diaz2024scaling,
	title        = {Scaling laws do not scale},
	author       = {Diaz, Fernando and Madaio, Michael},
	year         = 2024,
	booktitle    = {Proceedings of the AAAI/ACM Conference on AI, Ethics, and Society},
	volume       = 7,
	pages        = {341--357}
}

@inproceedings{sengupta2025position,
	title        = {Position: Enough of Scaling LLMs! Lets Focus on Downscaling},
	author       = {Ayan Sengupta and Yash Goel and Tanmoy Chakraborty},
	year         = 2025,
	booktitle    = {Proceedings of the $\mathit{42}^{nd}$ International Conference on Machine Learning},
	url          = {https://openreview.net/forum?id=CYJlJgEzZs}
}

@article{kaplan2020scaling,
	title        = {Scaling laws for neural language models},
	author       = {Kaplan, Jared and McCandlish, Sam and Henighan, Tom and Brown, Tom B and Chess, Benjamin and Child, Rewon and Gray, Scott and Radford, Alec and Wu, Jeffrey and Amodei, Dario},
	year         = 2020,
	journal      = {arXiv preprint arXiv:2001.08361}
}

@inproceedings{villalobos2024position,
	title        = {Position: Will we run out of data? Limits of LLM scaling based on human-generated data},
	author       = {Villalobos, Pablo and Ho, Anson and Sevilla, Jaime and Besiroglu, Tamay and Heim, Lennart and Hobbhahn, Marius},
	year         = 2024,
	booktitle    = {Forty-first International Conference on Machine Learning}
}

@article{jiang2024preventing,
	title        = {Preventing the immense increase in the life-cycle energy and carbon footprints of LLM-powered intelligent chatbots},
	author       = {Jiang, Peng and Sonne, Christian and Li, Wangliang and You, Fengqi and You, Siming},
	year         = 2024,
	journal      = {Engineering},
	publisher    = {Elsevier},
	volume       = 40,
	pages        = {202--210}
}

@article{fu2025llmco2,
	title        = {Llmco2: Advancing accurate carbon footprint prediction for llm inferences},
	author       = {Fu, Zhenxiao and Chen, Fan and Zhou, Shan and Li, Haitong and Jiang, Lei},
	year         = 2025,
	journal      = {ACM SIGENERGY Energy Informatics Review},
	publisher    = {ACM New York, NY, USA},
	volume       = 5,
	number       = 2,
	pages        = {63--68}
}

@article{jegham2025hungry,
	title        = {How hungry is ai? benchmarking energy, water, and carbon footprint of llm inference},
	author       = {Jegham, Nidhal and Abdelatti, Marwan and Elmoubarki, Lassad and Hendawi, Abdeltawab},
	year         = 2025,
	journal      = {arXiv preprint arXiv:2505.09598}
}

@article{clark2020electra,
	title        = {Electra: Pre-training text encoders as discriminators rather than generators},
	author       = {Clark, Kevin and Luong, Minh-Thang and Le, Quoc V and Manning, Christopher D},
	year         = 2020,
	journal      = {arXiv preprint arXiv:2003.10555}
}

@article{yu2024natural,
	title        = {Natural language reasoning, a survey},
	author       = {Yu, Fei and Zhang, Hongbo and Tiwari, Prayag and Wang, Benyou},
	year         = 2024,
	journal      = {ACM Computing Surveys},
	publisher    = {ACM New York, NY},
	volume       = 56,
	number       = 12,
	pages        = {1--39}
}

@article{wang2021kepler,
	title        = {KEPLER: A unified model for knowledge embedding and pre-trained language representation},
	author       = {Wang, Xiaozhi and Gao, Tianyu and Zhu, Zhaocheng and Zhang, Zhengyan and Liu, Zhiyuan and Li, Juanzi and Tang, Jian},
	year         = 2021,
	journal      = {Transactions of the Association for Computational Linguistics},
	publisher    = {MIT Press One Rogers Street, Cambridge, MA 02142-1209, USA journals-info~…},
	volume       = 9,
	pages        = {176--194}
}

@inproceedings{xiong2020pretrained,
	title        = {Pretrained Encyclopedia: Weakly Supervised Knowledge-Pretrained Language Model},
	author       = {Xiong, Wenhan and Du, Jingfei and Wang, William Yang and Stoyanov, Veselin},
	year         = 2020,
	booktitle    = {International Conference on Learning Representations}
}

@inproceedings{sun2020colake,
	title        = {CoLAKE: Contextualized Language and Knowledge Embedding},
	author       = {Sun, Tianxiang and Shao, Yunfan and Qiu, Xipeng and Guo, Qipeng and Hu, Yaru and Huang, Xuan-Jing and Zhang, Zheng},
	year         = 2020,
	booktitle    = {Proceedings of the 28th International Conference on Computational Linguistics},
	pages        = {3660--3670}
}

@article{sun2021ernie,
	title        = {Ernie 3.0: Large-scale knowledge enhanced pre-training for language understanding and generation},
	author       = {Sun, Yu and Wang, Shuohuan and Feng, Shikun and Ding, Siyu and Pang, Chao and Shang, Junyuan and Liu, Jiaxiang and Chen, Xuyi and Zhao, Yanbin and Lu, Yuxiang and others},
	year         = 2021,
	journal      = {arXiv preprint arXiv:2107.02137}
}

@inproceedings{guu2020retrieval,
	title        = {Retrieval augmented language model pre-training},
	author       = {Guu, Kelvin and Lee, Kenton and Tung, Zora and Pasupat, Panupong and Chang, Mingwei},
	year         = 2020,
	booktitle    = {International conference on machine learning},
	pages        = {3929--3938},
	organization = {PMLR}
}

@article{taylor2022galactica,
	title        = {Galactica: A large language model for science},
	author       = {Taylor, Ross and Kardas, Marcin and Cucurull, Guillem and Scialom, Thomas and Hartshorn, Anthony and Saravia, Elvis and Poulton, Andrew and Kerkez, Viktor and Stojnic, Robert},
	year         = 2022,
	journal      = {arXiv preprint arXiv:2211.09085}
}

@article{zhou2023efficient,
	title        = {Efficient prompting via dynamic in-context learning},
	author       = {Zhou, Wangchunshu and Jiang, Yuchen Eleanor and Cotterell, Ryan and Sachan, Mrinmaya},
	year         = 2023,
	journal      = {arXiv preprint arXiv:2305.11170}
}

@inproceedings{yin2023did,
	title        = {Did You Read the Instructions? Rethinking the Effectiveness of Task Definitions in Instruction Learning},
	author       = {Yin, Fan and Vig, Jesse and Laban, Philippe and Joty, Shafiq and Xiong, Caiming and Wu, Chien-Sheng},
	year         = 2023,
	booktitle    = {Proceedings of the 61st Annual Meeting of the Association for Computational Linguistics (Volume 1: Long Papers)},
	pages        = {3063--3079}
}

@article{jung2024discrete,
	title        = {Discrete prompt compression with reinforcement learning},
	author       = {Jung, Hoyoun and Kim, Kyung-Joong},
	year         = 2024,
	journal      = {IEEE Access},
	publisher    = {IEEE},
	volume       = 12,
	pages        = {72578--72587}
}

@inproceedings{jiang2023llmlingua,
	title        = {LLMLingua: Compressing Prompts for Accelerated Inference of Large Language Models},
	author       = {Jiang, Huiqiang and Wu, Qianhui and Lin, Chin-Yew and Yang, Yuqing and Qiu, Lili},
	booktitle    = {The 2023 Conference on Empirical Methods in Natural Language Processing}
}

@article{huang2023fewer,
	title        = {Fewer is more: Boosting llm reasoning with reinforced context pruning},
	author       = {Huang, Xijie and Zhang, Li Lyna and Cheng, Kwang-Ting and Yang, Fan and Yang, Mao},
	year         = 2023,
	journal      = {arXiv preprint arXiv:2312.08901}
}

@inproceedings{li2023compressing,
	title        = {Compressing Context to Enhance Inference Efficiency of Large Language Models},
	author       = {Li, Yucheng and Dong, Bo and Guerin, Frank and Lin, Chenghua},
	year         = 2023,
	booktitle    = {Proceedings of the 2023 Conference on Empirical Methods in Natural Language Processing},
	pages        = {6342--6353}
}

@inproceedings{xu2024recomp,
	title        = {RECOMP: IMPROVING RETRIEVAL-AUGMENTED LMS WITH COMPRESSION AND SELECTIVE AUGMENTATION},
	author       = {Xu, Fangyuan and Shi, Weijia and Choi, Eunsol},
	year         = 2024,
	booktitle    = {12th International Conference on Learning Representations, ICLR 2024}
}

@inproceedings{fei2024extending,
	title        = {Extending Context Window of Large Language Models via Semantic Compression},
	author       = {Fei, Weizhi and Niu, Xueyan and Zhou, Pingyi and Hou, Lu and Bai, Bo and Deng, Lei and Han, Wei},
	year         = 2024,
	booktitle    = {Findings of the Association for Computational Linguistics ACL 2024},
	pages        = {5169--5181}
}

@inproceedings{wingate2022prompt,
	title        = {Prompt Compression and Contrastive Conditioning for Controllability and Toxicity Reduction in Language Models},
	author       = {Wingate, David and Shoeybi, Mohammad and Sorensen, Taylor},
	year         = 2022,
	booktitle    = {Findings of the Association for Computational Linguistics: EMNLP 2022},
	pages        = {5621--5634}
}

@inproceedings{chevalier2023adapting,
	title        = {Adapting Language Models to Compress Contexts},
	author       = {Chevalier, Alexis and Wettig, Alexander and Ajith, Anirudh and Chen, Danqi},
	year         = 2023,
	booktitle    = {Proceedings of the 2023 Conference on Empirical Methods in Natural Language Processing},
	pages        = {3829--3846}
}

@inproceedings{ge2024incontext,
	title        = {In-context Autoencoder for Context Compression in a Large Language Model},
	author       = {Tao Ge and Hu Jing and Lei Wang and Xun Wang and Si-Qing Chen and Furu Wei},
	year         = 2024,
	booktitle    = {The Twelfth International Conference on Learning Representations},
	url          = {https://openreview.net/forum?id=uREj4ZuGJE}
}

@article{shazeer2019fast,
	title        = {Fast transformer decoding: One write-head is all you need},
	author       = {Shazeer, Noam},
	year         = 2019,
	journal      = {arXiv preprint arXiv:1911.02150}
}

@inproceedings{ainslie2023gqa,
	title        = {GQA: Training Generalized Multi-Query Transformer Models from Multi-Head Checkpoints},
	author       = {Ainslie, Joshua and Lee-Thorp, James and de Jong, Michiel and Zemlyanskiy, Yury and Lebron, Federico and Sanghai, Sumit},
	year         = 2023,
	booktitle    = {Proceedings of the 2023 Conference on Empirical Methods in Natural Language Processing},
	pages        = {4895--4901}
}

@misc{jiang2023mistral7b,
	title        = {Mistral 7B},
	author       = {Albert Q. Jiang and Alexandre Sablayrolles and Arthur Mensch and Chris Bamford and Devendra Singh Chaplot and Diego de las Casas and Florian Bressand and Gianna Lengyel and Guillaume Lample and Lucile Saulnier and Lélio Renard Lavaud and Marie-Anne Lachaux and Pierre Stock and Teven Le Scao and Thibaut Lavril and Thomas Wang and Timothée Lacroix and William El Sayed},
	year         = 2023,
	url          = {https://arxiv.org/abs/2310.06825},
	eprint       = {2310.06825},
	archiveprefix = {arXiv},
	primaryclass = {cs.CL}
}

@article{dubey2024llama,
	title        = {The llama 3 herd of models},
	author       = {Dubey, Abhimanyu and Jauhri, Abhinav and Pandey, Abhinav and Kadian, Abhishek and Al-Dahle, Ahmad and Letman, Aiesha and Mathur, Akhil and Schelten, Alan and Yang, Amy and Fan, Angela and others},
	year         = 2024,
	journal      = {arXiv e-prints},
	pages        = {arXiv--2407}
}

@article{ansar2025texim,
  title={TexIm FAST: Text-to-image encoding for semantic similarity evaluation of disproportionate sequences},
  author={Ansar, Wazib and Goswami, Saptarsi and Chakrabarti, Amlan and Chakraborty, Basabi},
  journal={ACM Transactions on Multimedia Computing, Communications and Applications},
  volume={21},
  number={6},
  pages={1--23},
  year={2025},
  publisher={ACM New York, NY}
}

@inproceedings{dettmers2024spqr,
	title        = {SpQR: A Sparse-Quantized Representation for Near-Lossless LLM Weight Compression},
	author       = {Dettmers, Tim and Svirschevski, Ruslan A and Egiazarian, Vage and Kuznedelev, Denis and Frantar, Elias and Ashkboos, Saleh and Borzunov, Alexander and Hoefler, Torsten and Alistarh, Dan},
	year         = 2024,
	booktitle    = {The Twelfth International Conference on Learning Representations}
}

@article{rokh2023comprehensive,
	title        = {A comprehensive survey on model quantization for deep neural networks in image classification},
	author       = {Rokh, Babak and Azarpeyvand, Ali and Khanteymoori, Alireza},
	year         = 2023,
	journal      = {ACM Transactions on Intelligent Systems and Technology},
	publisher    = {ACM New York, NY},
	volume       = 14,
	number       = 6,
	pages        = {1--50}
}

@inproceedings{rahman2023quantized,
	title        = {Quantized transformer language model implementations on edge devices},
	author       = {Rahman, Mohammad Wali Ur and Abrar, Murad Mehrab and Copening, Hunter Gibbons and Hariri, Salim and Shao, Sicong and Satam, Pratik and Salehi, Soheil},
	year         = 2023,
	booktitle    = {2023 International Conference on Machine Learning and Applications (ICMLA)},
	pages        = {709--716},
	organization = {IEEE}
}

@article{shen2024efficient,
	title        = {Efficient post-training quantization with fp8 formats},
	author       = {Shen, Haihao and Mellempudi, Naveen and He, Xin and Gao, Qun and Wang, Chang and Wang, Mengni},
	year         = 2024,
	journal      = {Proceedings of Machine Learning and Systems},
	volume       = 6,
	pages        = {483--498}
}

@article{shomron2021post,
	title        = {Post-training sparsity-aware quantization},
	author       = {Shomron, Gil and Gabbay, Freddy and Kurzum, Samer and Weiser, Uri},
	year         = 2021,
	journal      = {Advances in Neural Information Processing Systems},
	volume       = 34,
	pages        = {17737--17748}
}

@article{yao2022zeroquant,
	title        = {Zeroquant: Efficient and affordable post-training quantization for large-scale transformers},
	author       = {Yao, Zhewei and Yazdani Aminabadi, Reza and Zhang, Minjia and Wu, Xiaoxia and Li, Conglong and He, Yuxiong},
	year         = 2022,
	journal      = {Advances in neural information processing systems},
	volume       = 35,
	pages        = {27168--27183}
}

@inproceedings{chen2025efficientqat,
	title        = {Efficientqat: Efficient quantization-aware training for large language models},
	author       = {Chen, Mengzhao and Shao, Wenqi and Xu, Peng and Wang, Jiahao and Gao, Peng and Zhang, Kaipeng and Luo, Ping},
	year         = 2025,
	booktitle    = {Proceedings of the 63rd Annual Meeting of the Association for Computational Linguistics (Volume 1: Long Papers)},
	pages        = {10081--10100}
}

@article{liu2020bi,
	title        = {Bi-real net: Binarizing deep network towards real-network performance},
	author       = {Liu, Zechun and Luo, Wenhan and Wu, Baoyuan and Yang, Xin and Liu, Wei and Cheng, Kwang-Ting},
	year         = 2020,
	journal      = {International Journal of Computer Vision},
	publisher    = {Springer},
	volume       = 128,
	number       = 1,
	pages        = {202--219}
}

@article{hubara2016binarized,
	title        = {Binarized neural networks},
	author       = {Hubara, Itay and Courbariaux, Matthieu and Soudry, Daniel and El-Yaniv, Ran and Bengio, Yoshua},
	year         = 2016,
	journal      = {Advances in neural information processing systems},
	volume       = 29
}

@article{frantar2022gptq,
	title        = {Gptq: Accurate post-training quantization for generative pre-trained transformers},
	author       = {Frantar, Elias and Ashkboos, Saleh and Hoefler, Torsten and Alistarh, Dan},
	year         = 2022,
	journal      = {arXiv preprint arXiv:2210.17323}
}

@article{frantar2022optimal,
	title        = {Optimal brain compression: A framework for accurate post-training quantization and pruning},
	author       = {Frantar, Elias and Alistarh, Dan},
	year         = 2022,
	journal      = {Advances in Neural Information Processing Systems},
	volume       = 35,
	pages        = {4475--4488}
}

@article{lin2024awq,
	title        = {Awq: Activation-aware weight quantization for on-device llm compression and acceleration},
	author       = {Lin, Ji and Tang, Jiaming and Tang, Haotian and Yang, Shang and Chen, Wei-Ming and Wang, Wei-Chen and Xiao, Guangxuan and Dang, Xingyu and Gan, Chuang and Han, Song},
	year         = 2024,
	journal      = {Proceedings of machine learning and systems},
	volume       = 6,
	pages        = {87--100}
}

@inproceedings{lee2024owq,
	title        = {Owq: Outlier-aware weight quantization for efficient fine-tuning and inference of large language models},
	author       = {Lee, Changhun and Jin, Jungyu and Kim, Taesu and Kim, Hyungjun and Park, Eunhyeok},
	year         = 2024,
	booktitle    = {Proceedings of the AAAI Conference on Artificial Intelligence},
	volume       = 38,
	number       = 12,
	pages        = {13355--13364}
}

@inproceedings{kim2024squeezellm,
	title        = {SqueezeLLM: dense-and-sparse quantization},
	author       = {Kim, Sehoon and Hooper, Coleman and Gholami, Amir and Dong, Zhen and Li, Xiuyu and Shen, Sheng and Mahoney, Michael W and Keutzer, Kurt},
	year         = 2024,
	booktitle    = {Proceedings of the 41st International Conference on Machine Learning},
	pages        = {23901--23923}
}

@article{kim2023finequant,
	title        = {Finequant: Unlocking efficiency with fine-grained weight-only quantization for llms},
	author       = {Kim, Young Jin and Henry, Rawn and Fahim, Raffy and Awadalla, Hany Hassan},
	year         = 2023,
	journal      = {arXiv preprint arXiv:2308.09723}
}

@article{behdin2023quantease,
	title        = {QuantEase: Optimization-based quantization for language models},
	author       = {Behdin, Kayhan and Acharya, Ayan and Gupta, Aman and Song, Qingquan and Zhu, Siyu and Keerthi, Sathiya and Mazumder, Rahul},
	year         = 2023,
	journal      = {arXiv preprint arXiv:2309.01885}
}

@inproceedings{ma2024affinequant,
	title        = {AffineQuant: Affine Transformation Quantization for Large Language Models},
	author       = {Ma, Yuexiao and Li, Huixia and Zheng, Xiawu and Ling, Feng and Xiao, Xuefeng and Wang, Rui and Wen, Shilei and Chao, Fei and Ji, Rongrong},
	year         = 2024,
	booktitle    = {The Twelfth International Conference on Learning Representations}
}

@inproceedings{sheng2023flexgen,
	title        = {Flexgen: High-throughput generative inference of large language models with a single gpu},
	author       = {Sheng, Ying and Zheng, Lianmin and Yuan, Binhang and Li, Zhuohan and Ryabinin, Max and Chen, Beidi and Liang, Percy and R{\'e}, Christopher and Stoica, Ion and Zhang, Ce},
	year         = 2023,
	booktitle    = {International Conference on Machine Learning},
	pages        = {31094--31116},
	organization = {PMLR}
}

@inproceedings{xiao2023smoothquant,
	title        = {Smoothquant: Accurate and efficient post-training quantization for large language models},
	author       = {Xiao, Guangxuan and Lin, Ji and Seznec, Mickael and Wu, Hao and Demouth, Julien and Han, Song},
	year         = 2023,
	booktitle    = {International conference on machine learning},
	pages        = {38087--38099},
	organization = {PMLR}
}

@article{yuan2023rptq,
	title        = {Rptq: Reorder-based post-training quantization for large language models},
	author       = {Yuan, Zhihang and Niu, Lin and Liu, Jiawei and Liu, Wenyu and Wang, Xinggang and Shang, Yuzhang and Sun, Guangyu and Wu, Qiang and Wu, Jiaxiang and Wu, Bingzhe},
	year         = 2023,
	journal      = {arXiv preprint arXiv:2304.01089}
}

@inproceedings{guo2023olive,
	title        = {Olive: Accelerating large language models via hardware-friendly outlier-victim pair quantization},
	author       = {Guo, Cong and Tang, Jiaming and Hu, Weiming and Leng, Jingwen and Zhang, Chen and Yang, Fan and Liu, Yunxin and Guo, Minyi and Zhu, Yuhao},
	year         = 2023,
	booktitle    = {Proceedings of the 50th Annual International Symposium on Computer Architecture},
	pages        = {1--15}
}

@article{wu2023zeroquant,
	title        = {Zeroquant-fp: A leap forward in llms post-training w4a8 quantization using floating-point formats},
	author       = {Wu, Xiaoxia and Yao, Zhewei and He, Yuxiong},
	year         = 2023,
	journal      = {arXiv preprint arXiv:2307.09782}
}

@inproceedings{shao2024omniquant,
	title        = {OmniQuant: Omnidirectionally Calibrated Quantization for Large Language Models},
	author       = {Shao, Wenqi and Chen, Mengzhao and Zhang, Zhaoyang and Xu, Peng and Zhao, Lirui and Li, Zhiqian and Zhang, Kaipeng and Gao, Peng and Qiao, Yu and Luo, Ping},
	year         = 2024,
	booktitle    = {The Twelfth International Conference on Learning Representations}
}

@inproceedings{liu2024qllm,
	title        = {QLLM: Accurate and Efficient Low-Bitwidth Quantization for Large Language Models},
	author       = {Liu, Jing and Gong, Ruihao and Wei, Xiuying and Dong, Zhiwei and Cai, Jianfei and Zhuang, Bohan},
	year         = 2024,
	booktitle    = {International Conference on Learning Representations 2024},
	organization = {International Conference on Learning Representations (ICLR)}
}

@article{zhao2024atom,
	title        = {Atom: Low-bit quantization for efficient and accurate llm serving},
	author       = {Zhao, Yilong and Lin, Chien-Yu and Zhu, Kan and Ye, Zihao and Chen, Lequn and Zheng, Size and Ceze, Luis and Krishnamurthy, Arvind and Chen, Tianqi and Kasikci, Baris},
	year         = 2024,
	journal      = {Proceedings of Machine Learning and Systems},
	volume       = 6,
	pages        = {196--209}
}

@inproceedings{tao2022compression,
	title        = {Compression of Generative Pre-trained Language Models via Quantization},
	author       = {Tao, Chaofan and Hou, Lu and Zhang, Wei and Shang, Lifeng and Jiang, Xin and Liu, Qun and Luo, Ping and Wong, Ngai},
	year         = 2022,
	booktitle    = {Proceedings of the 60th Annual Meeting of the Association for Computational Linguistics (Volume 1: Long Papers)},
	pages        = {4821--4836}
}

@inproceedings{liu2024llm,
	title        = {LLM-QAT: Data-Free Quantization Aware Training for Large Language Models},
	author       = {Liu, Zechun and Oguz, Barlas and Zhao, Changsheng and Chang, Ernie and Stock, Pierre and Mehdad, Yashar and Shi, Yangyang and Krishnamoorthi, Raghuraman and Chandra, Vikas},
	year         = 2024,
	booktitle    = {Findings of the Association for Computational Linguistics ACL 2024},
	pages        = {467--484}
}

@article{wang2023bitnet,
	title        = {Bitnet: Scaling 1-bit transformers for large language models},
	author       = {Wang, Hongyu and Ma, Shuming and Dong, Li and Huang, Shaohan and Wang, Huaijie and Ma, Lingxiao and Yang, Fan and Wang, Ruiping and Wu, Yi and Wei, Furu},
	year         = 2023,
	journal      = {arXiv preprint arXiv:2310.11453}
}

@inproceedings{frantar2023sparsegpt,
	title        = {Sparsegpt: Massive language models can be accurately pruned in one-shot},
	author       = {Frantar, Elias and Alistarh, Dan},
	year         = 2023,
	booktitle    = {International conference on machine learning},
	pages        = {10323--10337},
	organization = {PMLR}
}

@inproceedings{sun2023a,
	title        = {A Simple and Effective Pruning Approach for Large Language Models},
	author       = {Mingjie Sun and Zhuang Liu and Anna Bair and J Zico Kolter},
	year         = 2023,
	booktitle    = {Workshop on Efficient Systems for Foundation Models @ ICML2023}
}

@inproceedings{shao2024one,
	title        = {One-shot sensitivity-aware mixed sparsity pruning for large language models},
	author       = {Shao, Hang and Liu, Bei and Qian, Yanmin},
	year         = 2024,
	booktitle    = {ICASSP 2024-2024 IEEE International Conference on Acoustics, Speech and Signal Processing (ICASSP)},
	pages        = {11296--11300},
	organization = {IEEE}
}

@article{tuli2023acceltran,
	title        = {AccelTran: A sparsity-aware accelerator for dynamic inference with transformers},
	author       = {Tuli, Shikhar and Jha, Niraj K},
	year         = 2023,
	journal      = {IEEE Transactions on Computer-Aided Design of Integrated Circuits and Systems},
	publisher    = {IEEE},
	volume       = 42,
	number       = 11,
	pages        = {4038--4051}
}

@article{bai2024gradient,
	title        = {Gradient-free adaptive global pruning for pre-trained language models},
	author       = {Bai, Guangji and Li, Yijiang and Ling, Chen and Kim, Kibaek and Zhao, Liang},
	year         = 2024,
    journal      = {arXiv preprint arXiv:2402.17946}
}

@article{ma2023llm,
	title        = {Llm-pruner: On the structural pruning of large language models},
	author       = {Ma, Xinyin and Fang, Gongfan and Wang, Xinchao},
	year         = 2023,
	journal      = {Advances in neural information processing systems},
	volume       = 36,
	pages        = {21702--21720}
}

@article{kurtic2023ziplm,
	title        = {Ziplm: Inference-aware structured pruning of language models},
	author       = {Kurti{\'c}, Eldar and Frantar, Elias and Alistarh, Dan},
	year         = 2023,
	journal      = {Advances in Neural Information Processing Systems},
	volume       = 36,
	pages        = {65597--65617}
}

@inproceedings{xia2024sheared,
	title        = {SHEARED LLAMA: ACCELERATING LANGUAGE MODEL PRE-TRAINING VIA STRUCTURED PRUNING},
	author       = {Xia, Mengzhou and Gao, Tianyu and Zeng, Zhiyuan and Chen, Danqi},
	year         = 2024,
	booktitle    = {12th International Conference on Learning Representations, ICLR 2024}
}

@article{valicenti2023mini,
	title        = {Mini-gpts: Efficient large language models through contextual pruning},
	author       = {Valicenti, Tim and Vidal, Justice and Patnaik, Ritik},
	year         = 2023,
	journal      = {arXiv preprint arXiv:2312.12682}
}

@inproceedings{liu2023deja,
	title        = {Deja vu: Contextual sparsity for efficient llms at inference time},
	author       = {Liu, Zichang and Wang, Jue and Dao, Tri and Zhou, Tianyi and Yuan, Binhang and Song, Zhao and Shrivastava, Anshumali and Zhang, Ce and Tian, Yuandong and Re, Christopher and others},
	year         = 2023,
	booktitle    = {International Conference on Machine Learning},
	pages        = {22137--22176},
	organization = {PMLR}
}

@inproceedings{timiryasov2023baby,
	title        = {Baby Llama: knowledge distillation from an ensemble of teachers trained on a small dataset with no performance penalty},
	author       = {Timiryasov, Inar and Tastet, Jean-Loup},
	year         = 2023,
	booktitle    = {Proceedings of the BabyLM Challenge at the 27th Conference on Computational Natural Language Learning},
	pages        = {279--289}
}

@inproceedings{gu2024minillm,
	title        = {MiniLLM: Knowledge Distillation of Large Language Models},
	author       = {Gu, Yuxian and Dong, Li and Wei, Furu and Huang, Minlie},
	year         = 2024,
	booktitle    = {The Twelfth International Conference on Learning Representations (ICLR}
}

@article{padmanabhan2023propagating,
	title        = {Propagating knowledge updates to lms through distillation},
	author       = {Padmanabhan, Shankar and Onoe, Yasumasa and Zhang, Michael and Durrett, Greg and Choi, Eunsol},
	year         = 2023,
	journal      = {Advances in Neural Information Processing Systems},
	volume       = 36,
	pages        = {47124--47142}
}

@inproceedings{liang2023less,
	title        = {Less is more: Task-aware layer-wise distillation for language model compression},
	author       = {Liang, Chen and Zuo, Simiao and Zhang, Qingru and He, Pengcheng and Chen, Weizhu and Zhao, Tuo},
	year         = 2023,
	booktitle    = {International Conference on Machine Learning},
	pages        = {20852--20867},
	organization = {PMLR}
}

@article{kim2023token,
	title        = {Token-scaled logit distillation for ternary weight generative language models},
	author       = {Kim, Minsoo and Lee, Sihwa and Lee, Janghwan and Hong, Sukjin and Chang, Du-Seong and Sung, Wonyong and Choi, Jungwook},
	year         = 2023,
	journal      = {Advances in Neural Information Processing Systems},
	volume       = 36,
	pages        = {42097--42118}
}

@inproceedings{zhang2023lifting,
	title        = {Lifting the Curse of Capacity Gap in Distilling Language Models},
	author       = {Zhang, Chen and Yang, Yang and Liu, Jiahao and Wang, Jingang and Xian, Yunsen and Wang, Benyou and Song, Dawei},
	year         = 2023,
	booktitle    = {Proceedings of the 61st Annual Meeting of the Association for Computational Linguistics (Volume 1: Long Papers)},
	pages        = {4535--4553}
}

@inproceedings{agarwal2024policy,
	title        = {On-policy distillation of language models: Learning from self-generated mistakes},
	author       = {Agarwal, Rishabh and Vieillard, Nino and Zhou, Yongchao and Stanczyk, Piotr and Garea, Sabela Ramos and Geist, Matthieu and Bachem, Olivier},
	year         = 2024,
	booktitle    = {The twelfth international conference on learning representations}
}

@article{achiam2023gpt,
	title        = {Gpt-4 technical report},
	author       = {Achiam, Josh and Adler, Steven and Agarwal, Sandhini and Ahmad, Lama and Akkaya, Ilge and Aleman, Florencia Leoni and Almeida, Diogo and Altenschmidt, Janko and Altman, Sam and Anadkat, Shyamal and others},
	year         = 2023,
	journal      = {arXiv preprint arXiv:2303.08774}
}

@article{huang2022context,
	title        = {In-context learning distillation: Transferring few-shot learning ability of pre-trained language models},
	author       = {Huang, Yukun and Chen, Yanda and Yu, Zhou and McKeown, Kathleen},
	year         = 2022,
	journal      = {arXiv preprint arXiv:2212.10670}
}

@inproceedings{zhao2024multistage,
	title        = {Multistage Collaborative Knowledge Distillation from a Large Language Model for Semi-Supervised Sequence Generation},
	author       = {Zhao, Jiachen and Zhao, Wenlong and Drozdov, Andrew and Rozonoyer, Benjamin and Sultan, Arafat and Lee, Jay-yoon and Iyyer, Mohit and McCallum, Andrew},
	year         = 2024,
	booktitle    = {Annual Meeting of the Association for Computational Linguistics}
}

@inproceedings{hsieh2023distilling,
	title        = {Distilling Step-by-Step! Outperforming Larger Language Models with Less Training Data and Smaller Model Sizes},
	author       = {Hsieh, Cheng-Yu and Li, Chun-Liang and YEH, CHIH-KUAN and Nakhost, Hootan and Fujii, Yasuhisa and Ratner, Alex Jason and Krishna, Ranjay and Lee, Chen-Yu and Pfister, Tomas},
	year         = 2023,
	booktitle    = {The 61st Annual Meeting Of The Association For Computational Linguistics}
}

@inproceedings{li2023symbolic,
	title        = {Symbolic Chain-of-Thought Distillation: Small Models Can Also" Think" Step-by-Step},
	author       = {Li, Liunian Harold and Hessel, Jack and Yu, Youngjae and Ren, Xiang and Chang, Kai-Wei and Choi, Yejin},
	year         = 2023,
	booktitle    = {The 61st Annual Meeting Of The Association For Computational Linguistics}
}

@inproceedings{ho2023large,
	title        = {Large Language Models Are Reasoning Teachers},
	author       = {Ho, Namgyu and Schmid, Laura and Yun, Se-Young},
	year         = 2023,
	booktitle    = {The 61st Annual Meeting Of The Association For Computational Linguistics}
}

@inproceedings{shridhar2023distilling,
	title        = {Distilling Reasoning Capabilities into Smaller Language Models},
	author       = {Shridhar, Kumar and Stolfo, Alessandro and Sachan, Mrinmaya},
	year         = 2023,
	booktitle    = {The 61st Annual Meeting Of The Association For Computational Linguistics}
}

@inproceedings{chen2023disco,
	title        = {DISCO: Distilling Counterfactuals with Large Language Models},
	author       = {Chen, Zeming and Gao, Qiyue and Bosselut, Antoine and Sabharwal, Ashish and Richardson, Kyle},
	year         = 2023,
	booktitle    = {Proceedings of the 61st Annual Meeting of the Association for Computational Linguistics (Volume 1: Long Papers)},
	pages        = {5514--5528}
}

@inproceedings{wu2024lamini,
	title        = {LaMini-LM: A Diverse Herd of Distilled Models from Large-Scale Instructions},
	author       = {Wu, Minghao and Waheed, Abdul and Zhang, Chiyu and Abdul-Mageed, Muhammad and Aji, Alham Fikri},
	year         = 2024,
	booktitle    = {Proceedings of the 18th Conference of the European Chapter of the Association for Computational Linguistics (Volume 1: Long Papers)},
	pages        = {944--964}
}

@inproceedings{jiang2023lion,
	title        = {Lion: Adversarial Distillation of Proprietary Large Language Models},
	author       = {Jiang, Yuxin and Chan, Chunkit and Chen, Mingyang and Wang, Wei},
	year         = 2023,
	booktitle    = {Proceedings of the 2023 Conference on Empirical Methods in Natural Language Processing},
	pages        = {3134--3154}
}

@inproceedings{xin2020deebert,
	title        = {DeeBERT: Dynamic Early Exiting for Accelerating BERT Inference},
	author       = {Xin, Ji and Tang, Raphael and Lee, Jaejun and Yu, Yaoliang and Lin, Jimmy},
	year         = 2020,
	booktitle    = {Proceedings of the 58th Annual Meeting of the Association for Computational Linguistics},
	pages        = {2246--2251}
}

@inproceedings{schwartz2020right,
	title        = {The Right Tool for the Job: Matching Model and Instance Complexities},
	author       = {Schwartz, Roy and Stanovsky, Gabriel and Swayamdipta, Swabha and Dodge, Jesse and Smith, Noah A},
	year         = 2020,
	booktitle    = {Proceedings of the 58th Annual Meeting of the Association for Computational Linguistics},
	pages        = {6640--6651}
}

@article{zhou2020bert,
	title        = {Bert loses patience: Fast and robust inference with early exit},
	author       = {Zhou, Wangchunshu and Xu, Canwen and Ge, Tao and McAuley, Julian and Xu, Ke and Wei, Furu},
	year         = 2020,
	journal      = {Advances in Neural Information Processing Systems},
	volume       = 33,
	pages        = {18330--18341}
}

@inproceedings{zhang2022pcee,
	title        = {PCEE-BERT: Accelerating BERT inference via patient and confident early exiting},
	author       = {Zhang, Zhen and Zhu, Wei and Zhang, Jinfan and Wang, Peng and Jin, Rize and Chung, Tae-Sun},
	year         = 2022,
	booktitle    = {Findings of the Association for Computational Linguistics: NAACL 2022},
	pages        = {327--338}
}

@inproceedings{wang2022skipbert,
	title        = {Skipbert: Efficient inference with shallow layer skipping},
	author       = {Wang, Jue and Chen, Ke and Chen, Gang and Shou, Lidan and McAuley, Julian},
	year         = 2022,
	booktitle    = {Proceedings of the 60th Annual Meeting of the Association for Computational Linguistics (Volume 1: Long Papers)},
	pages        = {7287--7301}
}

@inproceedings{din2024jump,
	title        = {Jump to Conclusions: Short-Cutting Transformers with Linear Transformations},
	author       = {Din, Alexander Yom and Karidi, Taelin and Choshen, Leshem and Geva, Mor},
	year         = 2024,
	booktitle    = {Proceedings of the 2024 Joint International Conference on Computational Linguistics, Language Resources and Evaluation (LREC-COLING 2024)},
	pages        = {9615--9625}
}

@article{schuster2022confident,
	title        = {Confident adaptive language modeling},
	author       = {Schuster, Tal and Fisch, Adam and Gupta, Jai and Dehghani, Mostafa and Bahri, Dara and Tran, Vinh and Tay, Yi and Metzler, Donald},
	year         = 2022,
	journal      = {Advances in Neural Information Processing Systems},
	volume       = 35,
	pages        = {17456--17472}
}

@inproceedings{tang2023you,
	title        = {You need multiple exiting: Dynamic early exiting for accelerating unified vision language model},
	author       = {Tang, Shengkun and Wang, Yaqing and Kong, Zhenglun and Zhang, Tianchi and Li, Yao and Ding, Caiwen and Wang, Yanzhi and Liang, Yi and Xu, Dongkuan},
	year         = 2023,
	booktitle    = {Proceedings of the IEEE/CVF conference on computer vision and pattern recognition},
	pages        = {10781--10791}
}

@inproceedings{liao2021global,
	title        = {A global past-future early exit method for accelerating inference of pre-trained language models},
	author       = {Liao, Kaiyuan and Zhang, Yi and Ren, Xuancheng and Su, Qi and Sun, Xu and He, Bin},
	year         = 2021,
	booktitle    = {Proceedings of the 2021 conference of the north american chapter of the association for computational linguistics: Human language technologies},
	pages        = {2013--2023}
}

@inproceedings{sun2022simple,
	title        = {A Simple Hash-Based Early Exiting Approach For Language Understanding and Generation},
	author       = {Sun, Tianxiang and Liu, Xiangyang and Zhu, Wei and Geng, Zhichao and Wu, Lingling and He, Yilong and Ni, Yuan and Xie, Guotong and Huang, Xuan-Jing and Qiu, Xipeng},
	year         = 2022,
	booktitle    = {Findings of the Association for Computational Linguistics: ACL 2022},
	pages        = {2409--2421}
}

@article{del2023skipdecode,
	title        = {Skipdecode: Autoregressive skip decoding with batching and caching for efficient llm inference},
	author       = {Del Corro, Luciano and Del Giorno, Allie and Agarwal, Sahaj and Yu, Bin and Awadallah, Ahmed and Mukherjee, Subhabrata},
	year         = 2023,
	journal      = {arXiv preprint arXiv:2307.02628}
}

@inproceedings{liu2022towards,
	title        = {Towards Efficient NLP: A Standard Evaluation and A Strong Baseline},
	author       = {Liu, Xiangyang and Sun, Tianxiang and He, Junliang and Wu, Jiawen and Wu, Lingling and Zhang, Xinyu and Jiang, Hao and Cao, Zhao and Huang, Xuan-Jing and Qiu, Xipeng},
	year         = 2022,
	booktitle    = {Proceedings of the 2022 Conference of the North American Chapter of the Association for Computational Linguistics: Human Language Technologies},
	pages        = {3288--3303}
}

@inproceedings{mangrulkar2022be3r,
	title        = {Be3r: Bert based early-exit using expert routing},
	author       = {Mangrulkar, Sourab and MS, Ankith and Sembium, Vivek},
	year         = 2022,
	booktitle    = {Proceedings of the 28th ACM SIGKDD Conference on Knowledge Discovery and Data Mining},
	pages        = {3504--3512}
}

@inproceedings{wang2021spatten,
	title        = {Spatten: Efficient sparse attention architecture with cascade token and head pruning},
	author       = {Wang, Hanrui and Zhang, Zhekai and Han, Song},
	year         = 2021,
	booktitle    = {2021 IEEE International Symposium on High-Performance Computer Architecture (HPCA)},
	pages        = {97--110},
	organization = {IEEE}
}

@inproceedings{kim2022learned,
	title        = {Learned token pruning for transformers},
	author       = {Kim, Sehoon and Shen, Sheng and Thorsley, David and Gholami, Amir and Kwon, Woosuk and Hassoun, Joseph and Keutzer, Kurt},
	year         = 2022,
	booktitle    = {Proceedings of the 28th ACM SIGKDD conference on knowledge discovery and data mining},
	pages        = {784--794}
}

@inproceedings{li2023constraint,
	title        = {Constraint-aware and ranking-distilled token pruning for efficient transformer inference},
	author       = {Li, Junyan and Zhang, Li Lyna and Xu, Jiahang and Wang, Yujing and Yan, Shaoguang and Xia, Yunqing and Yang, Yuqing and Cao, Ting and Sun, Hao and Deng, Weiwei and others},
	year         = 2023,
	booktitle    = {Proceedings of the 29th ACM SIGKDD Conference on Knowledge Discovery and Data Mining},
	pages        = {1280--1290}
}

@inproceedings{ye2021tr,
	title        = {TR-BERT: Dynamic Token Reduction for Accelerating BERT Inference},
	author       = {Ye, Deming and Lin, Yankai and Huang, Yufei and Sun, Maosong},
	year         = 2021,
	booktitle    = {Proceedings of the 2021 Conference of the North American Chapter of the Association for Computational Linguistics: Human Language Technologies},
	pages        = {5798--5809}
}

@inproceedings{guan2022transkimmer,
	title        = {Transkimmer: Transformer Learns to Layer-wise Skim},
	author       = {Guan, Yue and Li, Zhengyi and Leng, Jingwen and Lin, Zhouhan and Guo, Minyi},
	year         = 2022,
	booktitle    = {Proceedings of the 60th Annual Meeting of the Association for Computational Linguistics (Volume 1: Long Papers)},
	pages        = {7275--7286}
}

@inproceedings{cao2023pumer,
	title        = {PuMer: Pruning and Merging Tokens for Efficient Vision Language Models},
	author       = {Cao, Qingqing and Paranjape, Bhargavi and Hajishirzi, Hannaneh},
	year         = 2023,
	booktitle    = {Proceedings of the 61st Annual Meeting of the Association for Computational Linguistics (Volume 1: Long Papers)},
	pages        = {12890--12903}
}

@article{tan2023infor,
	title        = {Infor-Coef: Information Bottleneck-based Dynamic Token Downsampling for Compact and Efficient language model},
	author       = {Tan, Wenxi},
	year         = 2023,
	journal      = {arXiv preprint arXiv:2305.12458}
}

@inproceedings{wang2024smarttrim,
	title        = {SmartTrim: Adaptive Tokens and Attention Pruning for Efficient Vision-Language Models},
	author       = {Wang, Zekun and Chen, Jingchang and Zhou, Wangchunshu and Zhu, Haichao and Liang, Jiafeng and Shan, Liping and Liu, Ming and Xu, Dongliang and Yang, Qing and Qin, Bing},
	year         = 2024,
	booktitle    = {Proceedings of the 2024 Joint International Conference on Computational Linguistics, Language Resources and Evaluation (LREC-COLING 2024)},
	pages        = {14937--14953}
}

@inproceedings{pan2024llmlingua,
	title        = {LLMLingua-2: Data Distillation for Efficient and Faithful Task-Agnostic Prompt Compression},
	author       = {Pan, Zhuoshi and Wu, Qianhui and Jiang, Huiqiang and Xia, Menglin and Luo, Xufang and Zhang, Jue and Lin, Qingwei and R{\"u}hle, Victor and Yang, Yuqing and Lin, Chin-Yew and others},
	year         = 2024,
	booktitle    = {ACL (Findings)}
}

@inproceedings{kim2024compressed,
	title        = {COMPRESSED CONTEXT MEMORY FOR ONLINE LANGUAGE MODEL INTERACTION},
	author       = {Kim, Jang Hyun and Yeom, Junyoung and Yun, Sangdoo and Song, Hyun Oh},
	year         = 2024,
	booktitle    = {12th International Conference on Learning Representations, ICLR 2024}
}

@inproceedings{dongprompt,
	title        = {Prompt-prompted Adaptive Structured Pruning for Efficient LLM Generation},
	author       = {Dong, Harry and Chen, Beidi and Chi, Yuejie},
	year         = 2024,
	booktitle    = {Workshop on Efficient Systems for Foundation Models II@ ICML2024}
}

@inproceedings{fulazyllm,
	title        = {LazyLLM: Dynamic Token Pruning for Efficient Long Context LLM Inference},
	author       = {Fu, Qichen and Cho, Minsik and Merth, Thomas and Mehta, Sachin and Rastegari, Mohammad and Najibi, Mahyar},
	year         = 2024,
	booktitle    = {Workshop on Efficient Systems for Foundation Models II@ ICML2024}
}

@article{burton2012speculative,
	title        = {Speculative computation, parallelism, and functional programming},
	author       = {Burton, F Warren},
	year         = 2012,
	journal      = {IEEE Transactions on Computers},
	publisher    = {IEEE},
	volume       = 100,
	number       = 12,
	pages        = {1190--1193}
}

@inproceedings{leviathan2023fast,
	title        = {Fast inference from transformers via speculative decoding},
	author       = {Leviathan, Yaniv and Kalman, Matan and Matias, Yossi},
	year         = 2023,
	booktitle    = {International Conference on Machine Learning},
	pages        = {19274--19286},
	organization = {PMLR}
}

@article{chen2023accelerating,
	title        = {Accelerating large language model decoding with speculative sampling},
	author       = {Chen, Charlie and Borgeaud, Sebastian and Irving, Geoffrey and Lespiau, Jean-Baptiste and Sifre, Laurent and Jumper, John},
	year         = 2023,
	journal      = {arXiv preprint arXiv:2302.01318}
}

@inproceedings{spectoraccelerating,
	title        = {Accelerating LLM Inference with Staged Speculative Decoding},
	author       = {Spector, Benjamin Frederick and Re, Christopher},
	year         = 2023,
	booktitle    = {Workshop on Efficient Systems for Foundation Models@ ICML2023}
}

@ARTICLE{ansar2024beexformer,
  author={Ansar, Wazib and Goswami, Saptarsi and Chakrabarti, Amlan},
  journal={IEEE Transactions on Sustainable Computing}, 
  title={BEExformer: A Fast Inferencing Binarized Transformer With Early Exits}, 
  year={2026},
  volume={11},
  number={2},
  pages={98-110}
  }

@article{yuan2023asvd,
	title        = {Asvd: Activation-aware singular value decomposition for compressing large language models},
	author       = {Yuan, Zhihang and Shang, Yuzhang and Song, Yue and Wu, Qiang and Yan, Yan and Sun, Guangyu},
	year         = 2023,
	journal      = {arXiv preprint arXiv:2312.05821}
}

@inproceedings{hsulanguage,
	title        = {Language model compression with weighted low-rank factorization},
	author       = {Hsu, Yen-Chang and Hua, Ting and Chang, Sungen and Lou, Qian and Shen, Yilin and Jin, Hongxia},
	year         = 2022,
	booktitle    = {International Conference on Learning Representations}
}

@inproceedings{wangsvd,
	title        = {SVD-LLM: Truncation-aware Singular Value Decomposition for Large Language Model Compression},
	author       = {Wang, Xin and Zheng, Yu and Wan, Zhongwei and Zhang, Mi},
	year         = 2024,
	booktitle    = {The Thirteenth International Conference on Learning Representations}
}

@inproceedings{li2023losparse,
	title        = {Losparse: Structured compression of large language models based on low-rank and sparse approximation},
	author       = {Li, Yixiao and Yu, Yifan and Zhang, Qingru and Liang, Chen and He, Pengcheng and Chen, Weizhu and Zhao, Tuo},
	year         = 2023,
	booktitle    = {International Conference on Machine Learning},
	pages        = {20336--20350},
	organization = {PMLR}
}

@article{chand2023dsformer,
	title        = {Dsformer: Effective compression of text-transformers by dense-sparse weight factorization},
	author       = {Chand, Rahul and Prabhu, Yashoteja and Kumar, Pratyush},
	year         = 2023,
	journal      = {arXiv preprint arXiv:2312.13211}
}

@article{saha2023matrix,
	title        = {Matrix compression via randomized low rank and low precision factorization},
	author       = {Saha, Rajarshi and Srivastava, Varun and Pilanci, Mert},
	year         = 2023,
	journal      = {Advances in Neural Information Processing Systems},
	volume       = 36,
	pages        = {18828--18872}
}

@article{yao2023zeroquant,
	title        = {Zeroquant-v2: Exploring post-training quantization in llms from comprehensive study to low rank compensation},
	author       = {Yao, Zhewei and Wu, Xiaoxia and Li, Cheng and Youn, Stephen and He, Yuxiong},
	year         = 2023,
	journal      = {arXiv preprint arXiv:2303.08302}
}

@article{chen2021drone,
	title        = {Drone: Data-aware low-rank compression for large nlp models},
	author       = {Chen, Patrick and Yu, Hsiang-Fu and Dhillon, Inderjit and Hsieh, Cho-Jui},
	year         = 2021,
	journal      = {Advances in neural information processing systems},
	volume       = 34,
	pages        = {29321--29334}
}

@inproceedings{kaushal2024lord,
	title        = {LoRD: Low-Rank Decomposition of Monolingual Code LLMs for One-Shot Compression},
	author       = {Kaushal, Ayush and Vaidhya, Tejas and Rish, Irina},
	year         = 2024,
	booktitle    = {ICML 2024 Workshop on Foundation Models in the Wild}
}

@article{xu2023tensorgpt,
	title        = {TensorGPT: Efficient Compression of Large Language Models based on Tensor-Train Decomposition},
	author       = {Xu, Mingxue and Xu, Yao Lei and Mandic, Danilo P},
	year         = 2023,
	journal      = {arXiv preprint arXiv:2307.00526}
}

@inproceedings{reid2021subformer,
	title        = {Subformer: Exploring Weight Sharing for Parameter Efficiency in Generative Transformers},
	author       = {Reid, Machel and Marrese-Taylor, Edison and Matsuo, Yutaka},
	year         = 2021,
	booktitle    = {Findings of the Association for Computational Linguistics: EMNLP 2021},
	pages        = {4081--4090}
}

@inproceedings{lepikhingshard,
	title        = {GShard: Scaling Giant Models with Conditional Computation and Automatic Sharding},
	author       = {Lepikhin, Dmitry and Lee, HyoukJoong and Xu, Yuanzhong and Chen, Dehao and Firat, Orhan and Huang, Yanping and Krikun, Maxim and Shazeer, Noam and Chen, Zhifeng},
	year         = 2021,
	booktitle    = {International Conference on Learning Representations}
}

@article{fedus2022switch,
	title        = {Switch transformers: Scaling to trillion parameter models with simple and efficient sparsity},
	author       = {Fedus, William and Zoph, Barret and Shazeer, Noam},
	year         = 2022,
	journal      = {Journal of Machine Learning Research},
	volume       = 23,
	number       = 120,
	pages        = {1--39}
}

@inproceedings{lewis2021base,
	title        = {Base layers: Simplifying training of large, sparse models},
	author       = {Lewis, Mike and Bhosale, Shruti and Dettmers, Tim and Goyal, Naman and Zettlemoyer, Luke},
	year         = 2021,
	booktitle    = {International Conference on Machine Learning},
	pages        = {6265--6274},
	organization = {PMLR}
}

@article{zhou2022mixture,
	title        = {Mixture-of-experts with expert choice routing},
	author       = {Zhou, Yanqi and Lei, Tao and Liu, Hanxiao and Du, Nan and Huang, Yanping and Zhao, Vincent and Dai, Andrew M and Le, Quoc V and Laudon, James and others},
	year         = 2022,
	journal      = {Advances in Neural Information Processing Systems},
	volume       = 35,
	pages        = {7103--7114}
}

@article{ren2023pangu,
	title        = {Pangu-$\Sigma$: Towards trillion parameter language model with sparse heterogeneous computing},
	author       = {Ren, Xiaozhe and Zhou, Pingyi and Meng, Xinfan and Huang, Xinjing and Wang, Yadao and Wang, Weichao and Li, Pengfei and Zhang, Xiaoda and Podolskiy, Alexander and Arshinov, Grigory and others},
	year         = 2023,
	journal      = {arXiv preprint arXiv:2303.10845}
}

@article{wang2021ernie,
	title        = {Ernie 3.0 titan: Exploring larger-scale knowledge enhanced pre-training for language understanding and generation},
	author       = {Wang, Shuohuan and Sun, Yu and Xiang, Yang and Wu, Zhihua and Ding, Siyu and Gong, Weibao and Feng, Shikun and Shang, Junyuan and Zhao, Yanbin and Pang, Chao and others},
	year         = 2021,
	journal      = {arXiv preprint arXiv:2112.12731}
}

@article{jiang2024mixtral,
	title        = {Mixtral of experts},
	author       = {Jiang, Albert Q and Sablayrolles, Alexandre and Roux, Antoine and Mensch, Arthur and Savary, Blanche and Bamford, Chris and Chaplot, Devendra Singh and Casas, Diego de las and Hanna, Emma Bou and Bressand, Florian and others},
	year         = 2024,
	journal      = {arXiv preprint arXiv:2401.04088}
}

@inproceedings{shen2024mixture,
	title        = {Mixture-of-Experts Meets Instruction Tuning: A Winning Combination for Large Language Models},
	author       = {Shen, Sheng and Hou, Le and Zhou, Yanqi and Du, Nan and Longpre, Shayne and Wei, Jason and Chung, Hyung Won and Zoph, Barret and Fedus, William and Chen, Xinyun and others},
	year         = 2024,
	booktitle    = {ICLR}
}

@article{liu2024deepseek3,
	title        = {Deepseek-v3 technical report},
	author       = {Liu, Aixin and Feng, Bei and Xue, Bing and Wang, Bingxuan and Wu, Bochao and Lu, Chengda and Zhao, Chenggang and Deng, Chengqi and Zhang, Chenyu and Ruan, Chong and others},
	year         = 2024,
	journal      = {arXiv preprint arXiv:2412.19437}
}

@article{guo2025deepseek,
	title        = {Deepseek-r1: Incentivizing reasoning capability in llms via reinforcement learning},
	author       = {Guo, Daya and Yang, Dejian and Zhang, Haowei and Song, Junxiao and Zhang, Ruoyu and Xu, Runxin and Zhu, Qihao and Ma, Shirong and Wang, Peiyi and Bi, Xiao and others},
	year         = 2025,
	journal      = {arXiv preprint arXiv:2501.12948}
}

@article{liu2024deepseek,
	title        = {Deepseek-v2: A strong, economical, and efficient mixture-of-experts language model},
	author       = {Liu, Aixin and Feng, Bei and Wang, Bin and Wang, Bingxuan and Liu, Bo and Zhao, Chenggang and Dengr, Chengqi and Ruan, Chong and Dai, Damai and Guo, Daya and others},
	year         = 2024,
	journal      = {arXiv preprint arXiv:2405.04434}
}

@inproceedings{dai2024deepseekmoe,
	title        = {DeepSeekMoE: Towards Ultimate Expert Specialization in Mixture-of-Experts Language Models},
	author       = {Dai, Damai and Deng, Chengqi and Zhao, Chenggang and Xu, Rx and Gao, Huazuo and Chen, Deli and Li, Jiashi and Zeng, Wangding and Yu, Xingkai and Wu, Y and others},
	year         = 2024,
	booktitle    = {Proceedings of the 62nd Annual Meeting of the Association for Computational Linguistics (Volume 1: Long Papers)},
	pages        = {1280--1297}
}

@inproceedings{alberti2023sumformer,
	title        = {Sumformer: Universal Approximation for Efficient Transformers},
	author       = {Alberti, Silas and Dern, Niclas and Thesing, Laura and Kutyniok, Gitta},
	year         = 2023,
	booktitle    = {Topological, Algebraic and Geometric Learning Workshops 2023},
	pages        = {72--86},
	organization = {PMLR}
}

@inproceedings{wang2022clusterformer,
	title        = {ClusterFormer: Neural clustering attention for efficient and effective transformer},
	author       = {Wang, Ningning and Gan, Guobing and Zhang, Peng and Zhang, Shuai and Wei, Junqiu and Liu, Qun and Jiang, Xin},
	year         = 2022,
	booktitle    = {Proceedings of the 60th Annual Meeting of the Association for Computational Linguistics (Volume 1: Long Papers)},
	pages        = {2390--2402}
}

@inproceedings{han2024hyperattention,
	title        = {HyperAttention: Long-context Attention in Near-Linear Time},
	author       = {Han, Insu and Jayaram, R and Karbasi, A and Mirrokno, V and Woodruff, D and Zandieh, A},
	year         = 2024,
	booktitle    = {International Conference on Learning Representations},
	organization = {International Conference on Learning Representations}
}

@article{vyas2020fast,
	title        = {Fast transformers with clustered attention},
	author       = {Vyas, Apoorv and Katharopoulos, Angelos and Fleuret, Fran{\c{c}}ois},
	year         = 2020,
	journal      = {Advances in Neural Information Processing Systems},
	volume       = 33,
	pages        = {21665--21674}
}

@article{dao2022flashattention,
	title        = {Flashattention: Fast and memory-efficient exact attention with io-awareness},
	author       = {Dao, Tri and Fu, Dan and Ermon, Stefano and Rudra, Atri and R{\'e}, Christopher},
	year         = 2022,
	journal      = {Advances in neural information processing systems},
	volume       = 35,
	pages        = {16344--16359}
}

@article{pagliardini2023fast,
	title        = {Fast attention over long sequences with dynamic sparse flash attention},
	author       = {Pagliardini, Matteo and Paliotta, Daniele and Jaggi, Martin and Fleuret, Fran{\c{c}}ois},
	year         = 2023,
	journal      = {Advances in Neural Information Processing Systems},
	volume       = 36,
	pages        = {59808--59831}
}

@inproceedings{sarthi2024raptor,
	title        = {Raptor: Recursive abstractive processing for tree-organized retrieval},
	author       = {Sarthi, Parth and Abdullah, Salman and Tuli, Aditi and Khanna, Shubh and Goldie, Anna and Manning, Christopher D},
	year         = 2024,
	booktitle    = {The Twelfth International Conference on Learning Representations}
}

@article{chen2023walking,
	title        = {Walking down the memory maze: Beyond context limit through interactive reading},
	author       = {Chen, Howard and Pasunuru, Ramakanth and Weston, Jason and Celikyilmaz, Asli},
	year         = 2023,
	journal      = {arXiv preprint arXiv:2310.05029}
}

@inproceedings{xiaoefficient,
	title        = {Efficient Streaming Language Models with Attention Sinks},
	author       = {Xiao, Guangxuan and Tian, Yuandong and Chen, Beidi and Han, Song and Lewis, Mike},
	year         = 2024,
	booktitle    = {The Twelfth International Conference on Learning Representations}
}

@inproceedings{martins2022former,
	title        = {$\infty$-former: Infinite Memory Transformer},
	author       = {Martins, Pedro Henrique  and Marinho, Zita  and Martins, Andre},
	year         = 2022,
	booktitle    = {Proceedings of the 60th Annual Meeting of the Association for Computational Linguistics (Volume 1: Long Papers)},
	publisher    = {Association for Computational Linguistics},
	pages        = {5468--5485}
}

@inproceedings{xia2019tied,
	title        = {Tied transformers: Neural machine translation with shared encoder and decoder},
	author       = {Xia, Yingce and He, Tianyu and Tan, Xu and Tian, Fei and He, Di and Qin, Tao},
	year         = 2019,
	booktitle    = {Proceedings of the AAAI conference on artificial intelligence},
	volume       = 33,
	number       = {01},
	pages        = {5466--5473}
}

@article{rothe2020leveraging,
	title        = {Leveraging pre-trained checkpoints for sequence generation tasks},
	author       = {Rothe, Sascha and Narayan, Shashi and Severyn, Aliaksei},
	year         = 2020,
	journal      = {Transactions of the Association for Computational Linguistics},
	publisher    = {MIT Press One Rogers Street, Cambridge, MA 02142-1209, USA journals-info~…},
	volume       = 8,
	pages        = {264--280}
}

@inproceedings{dabre2019recurrent,
	title        = {Recurrent stacking of layers for compact neural machine translation models},
	author       = {Dabre, Raj and Fujita, Atsushi},
	year         = 2019,
	booktitle    = {Proceedings of the AAAI Conference on Artificial Intelligence},
	volume       = 33,
	number       = {01},
	pages        = {6292--6299}
}

@inproceedings{dehghaniuniversal,
	title        = {Universal Transformers},
	author       = {Dehghani, Mostafa and Gouws, Stephan and Vinyals, Oriol and Uszkoreit, Jakob and Kaiser, Lukasz},
	year         = 2019,
	booktitle    = {International Conference on Learning Representations}
}

@inproceedings{takase2023lessons,
	title        = {Lessons on Parameter Sharing across Layers in Transformers},
	author       = {Takase, Sho and Kiyono, Shun},
	year         = 2023,
	booktitle    = {The Fourth Workshop on Simple and Efficient Natural Language Processing},
	pages        = 78
}

@inproceedings{wang2025basis,
	title        = {Basis Sharing: Cross-Layer Parameter Sharing for Large Language Model Compression},
	author       = {Wang, Jingcun and Chen, Yu-Guang and Lin, Ing-Chao and Li, Bing and Zhang, Grace Li},
	year         = 2025,
	booktitle    = {Proceedings of the International Conference on Learning Representations (ICLR)}
}

@article{ansar2026lesum,
title = {LeSum: Cost-effective LLM-driven hybrid summarization of Indian legal judgments},
journal = {Expert Systems with Applications},
volume = {332},
pages = {133683},
year = {2026},
issn = {0957-4174},
author = {Wazib Ansar and Saptarsi Goswami and Amlan Chakrabarti}
}

@inproceedings{wenzek2020ccnet,
	title        = {CCNet: Extracting High Quality Monolingual Datasets from Web Crawl Data},
	author       = {Wenzek, Guillaume and Lachaux, Marie-Anne and Conneau, Alexis and Chaudhary, Vishrav and Guzm{\'a}n, Francisco and Joulin, Armand and Grave, {\'E}douard},
	year         = 2020,
	booktitle    = {Proceedings of the Twelfth Language Resources and Evaluation Conference},
	pages        = {4003--4012}
}

@article{tirumala2023d4,
	title        = {D4: Improving llm pretraining via document de-duplication and diversification},
	author       = {Tirumala, Kushal and Simig, Daniel and Aghajanyan, Armen and Morcos, Ari},
	year         = 2023,
	journal      = {Advances in Neural Information Processing Systems},
	volume       = 36,
	pages        = {53983--53995}
}

@article{cunha2025noise,
	title        = {A noise-oriented and redundancy-aware instance selection framework},
	author       = {Cunha, Washington and Moreo Fern{\'a}ndez, Alejandro and Esuli, Andrea and Sebastiani, Fabrizio and Rocha, Leonardo and Gon{\c{c}}alves, Marcos Andr{\'e}},
	year         = 2025,
	journal      = {ACM Transactions on Information Systems},
	publisher    = {ACM New York, NY},
	volume       = 43,
	number       = 2,
	pages        = {1--33}
}

@inproceedings{guang2024data,
	title        = {Data Redundancy Elimination and Noise Processing via Large Language Model Prompt Engineering},
	author       = {Guang, Jin and Chao, Jing and Tianqi, Zong and Siya, Chen and Chaoyuan, Cui and Jun, Fan},
	year         = 2024,
	booktitle    = {2024 10th International Conference on Big Data and Information Analytics (BigDIA)},
	pages        = {818--825},
	organization = {IEEE}
}

@article{yin2024entropy,
	title        = {Entropy law: The story behind data compression and llm performance},
	author       = {Yin, Mingjia and Wu, Chuhan and Wang, Yufei and Wang, Hao and Guo, Wei and Wang, Yasheng and Liu, Yong and Tang, Ruiming and Lian, Defu and Chen, Enhong},
	year         = 2024,
	journal      = {arXiv preprint arXiv:2407.06645}
}

@article{mitchell2022measuring,
	title        = {Measuring data},
	author       = {Mitchell, Margaret and Luccioni, Alexandra Sasha and Lambert, Nathan and Gerchick, Marissa and McMillan-Major, Angelina and Ozoani, Ezinwanne and Rajani, Nazneen and Thrush, Tristan and Jernite, Yacine and Kiela, Douwe},
	year         = 2022,
	journal      = {arXiv preprint arXiv:2212.05129}
}

@inproceedings{pattnaik2024enhancing,
	title        = {Enhancing alignment using curriculum learning \& ranked preferences},
	author       = {Pattnaik, Pulkit and Maheshwary, Rishabh and Ogueji, Kelechi and Yadav, Vikas and Madhusudhan, Sathwik Tejaswi},
	year         = 2024,
	booktitle    = {Findings of the Association for Computational Linguistics: EMNLP 2024},
	pages        = {12891--12907}
}

@article{abdin2024phi,
	title        = {Phi-4 technical report},
	author       = {Abdin, Marah and Aneja, Jyoti and Behl, Harkirat and Bubeck, S{\'e}bastien and Eldan, Ronen and Gunasekar, Suriya and Harrison, Michael and Hewett, Russell J and Javaheripi, Mojan and Kauffmann, Piero and others},
	year         = 2024,
	journal      = {arXiv preprint arXiv:2412.08905}
}

@article{yang2025qwen3,
	title        = {Qwen3 technical report},
	author       = {Yang, An and Li, Anfeng and Yang, Baosong and Zhang, Beichen and Hui, Binyuan and Zheng, Bo and Yu, Bowen and Gao, Chang and Huang, Chengen and Lv, Chenxu and others},
	year         = 2025,
	journal      = {arXiv preprint arXiv:2505.09388}
}

@inproceedings{xu2024context,
	title        = {In-Context Example Ordering Guided by Label Distributions},
	author       = {Xu, Zhichao and Cohen, Daniel and Wang, Bei and Srikumar, Vivek},
	year         = 2024,
	booktitle    = {Findings of the Association for Computational Linguistics: NAACL 2024},
	pages        = {2623--2640}
}

@article{wu2024prompt,
	title        = {Prompt optimization with EASE? efficient ordering-aware automated selection of exemplars},
	author       = {Wu, Zhaoxuan and Lin, Xiaoqiang and Dai, Zhongxiang and Hu, Wenyang and Shu, Yao and Ng, See-Kiong and Jaillet, Patrick and Low, Bryan Kian Hsiang},
	year         = 2024,
	journal      = {Advances in Neural Information Processing Systems},
	volume       = 37,
	pages        = {122706--122740}
}

@inproceedings{guo2024makes,
	title        = {What makes a good order of examples in in-context learning},
	author       = {Guo, Qi and Wang, Leiyu and Wang, Yidong and Ye, Wei and Zhang, Shikun},
	year         = 2024,
	booktitle    = {Findings of the Association for Computational Linguistics: ACL 2024},
	pages        = {14892--14904}
}

@article{sachdeva2024train,
	title        = {How to train data-efficient llms},
	author       = {Sachdeva, Noveen and Coleman, Benjamin and Kang, Wang-Cheng and Ni, Jianmo and Hong, Lichan and Chi, Ed H and Caverlee, James and McAuley, Julian and Cheng, Derek Zhiyuan},
	year         = 2024,
	journal      = {arXiv preprint arXiv:2402.09668}
}

@article{parkar2024selectllm,
	title        = {SelectLLM: Can LLMs Select Important Instructions to Annotate?},
	author       = {Parkar, Ritik Sachin and Kim, Jaehyung and Park, Jong Inn and Kang, Dongyeop},
	year         = 2024,
	journal      = {arXiv preprint arXiv:2401.16553}
}

@inproceedings{saranathan2025sublime,
	title        = {SubLIME: Subset Selection via Rank Correlation Prediction for Data-Efficient LLM Evaluation},
	author       = {Saranathan, Gayathri and Xu, Cong and Alam, Mahammad Parwez and Kumar, Tarun and Foltin, Martin and Wong, Soon Yee and Bhattacharya, Suparna},
	year         = 2025,
	booktitle    = {Proceedings of the 63rd Annual Meeting of the Association for Computational Linguistics (Volume 1: Long Papers)},
	pages        = {30572--30593}
}

@article{wu2025lamdas,
	title        = {LAMDAS: LLM as an Implicit Classifier for Domain-specific Data Selection},
	author       = {Wu, Jian and Yu, Hang and Liu, Bingchang and Yang, Wenjie and Di, Peng and Li, Jianguo and Zhang, Yue},
	year         = 2025,
	journal      = {arXiv preprint arXiv:2509.06524}
}

\end{document}